\documentclass{iopjournal}
\RequirePackage{fix-cm}
\usepackage[utf8]{inputenc}
\usepackage[T1]{fontenc}
\usepackage{url}
\usepackage{booktabs}
\usepackage{amsfonts}
\usepackage{amsmath}
\usepackage{nicefrac}
\usepackage{microtype}
\usepackage{xcolor}
\definecolor{mypurple}{RGB}{230,220,255}
\usepackage{wrapfig}
\usepackage{makecell}
\usepackage{pifont}
\usepackage{multirow}
\usepackage[numbers,sort&compress]{natbib}

\usepackage{multirow}
\usepackage{pifont}

\usepackage{hyperref}                  
\usepackage{bm}
\usepackage{colortbl}
\usepackage{xcolor}
\usepackage{etoolbox}
\usepackage{orcidlink}

\begin{document}

\articletype{Paper}

\title{REALM: Retrospective Encoder Alignment for LFP Modeling}

\author{Peicheng Wu$^1$\orcidlink{0009-0002-9779-4427}, Zhenyu Bu$^1$\orcidlink{0000-0001-8360-0535}, Runze Ma$^2$\orcidlink{0009-0001-2620-0575} and Lin Du$^{1,3,*}$\orcidlink{0000-0002-6111-4947}}\\
\affil{$^1$Department of Biomedical Engineering, The Ohio State University, Columbus, OH, United States}\\
\affil{$^2$Department of Information Technology, Monash University Malaysia, Subang Jaya, Malaysia}\\
\affil{$^3$Department of Neurological Surgery, The Ohio State University, Columbus, OH, United States}\\
\affil{$^*$Author to whom any correspondence should be addressed.}\\
\email{du.1093@osu.edu}

\keywords{local field potentials, brain--computer interfaces, state space models, knowledge distillation, neural decoding}

\begin{abstract}

\textit{Objective}. Spike activity has been the dominant neural signal for behavior decoding because its high spatial and temporal resolution supports accurate decoding. However, as intracortical brain--computer interfaces (iBCIs) move toward higher channel counts and wireless operation, the high sampling rates required to record spike signals create substantial power and data-bandwidth demands. Local field potentials (LFPs) capture brain activity at a different spatiotemporal scale and offer complementary advantages, including greater long-term stability, reduced energy consumption, and lower bandwidth requirements. Despite these benefits, LFP-based decoding models often achieve lower accuracy and rely on non-causal architectures that cannot be used directly for real-time deployment. \textit{Approach}. We propose REALM: a retrospective knowledge distillation (RKD) framework for high-performance causal LFP behavior decoding. Inspired by offline-to-online distillation strategies in speech recognition, REALM transfers non-causal representational knowledge learned by a pretrained, multi-session bidirectional LFP teacher to a causal student model for real-time deployment. Specifically, we first pretrain a bidirectional Mamba-2 teacher model across multiple recording sessions using continuous masked autoencoding (CMAE). We then distill its representation into a compact causal student model using a combined objective of representation alignment and autoencoding. \textit{Main results}. REALM achieves the highest mean accuracy among the decoders compared, under both linear-probe evaluation and per-session fine-tuning. Its improvements over each baseline, including the state-of-the-art CrossModalDistill, are statistically significant. REALM achieves this performance using a single modality---local field potentials---throughout pretraining, distillation, and decoding, while using less than half as many parameters as CrossModalDistill's published student and requiring one-tenth of the pretraining time. \textit{Significance}. These results demonstrate that REALM effectively bridges the gap between offline and real-time neural decoding. Importantly, REALM shows that a causal LFP-only model can achieve competitive decoding accuracy compared with a non-causal multi-modal model, offering a practical and scalable alternative for next-generation wireless and implantable iBCIs.
\end{abstract}

\section{Introduction}

Intracortical brain--computer interfaces (iBCIs) are emerging as a promising strategy for restoring motor function and communication for people with conditions such as amyotrophic lateral sclerosis, spinal cord injury, and stroke~\cite{R25_shenoy2014combining, R26_shanechi2019brain, R27_oganesian2024brain}. By translating neural activity into control signals, iBCIs enable users to interact with external devices such as robotic arms, computer pointers, or neuroprosthetic systems. Central challenges in iBCI systems are signal stability, a wireless low-bandwidth budget, and real-time inference. To date, most neural decoding approaches have relied on extracellular spike recordings. These include single-session models based on linear dynamical systems~\cite{R1_gao2016linear, R28_sani2021modeling}, sequential autoencoders~\cite{R2_pandarinath2018inferring, R29_hurwitz2021targeted}, and Transformers~\cite{R4_ye2021representation, R31_le2022stndt}, as well as multi-session models such as CEBRA~\cite{R6_schneider2023learnable}, NDT2~\cite{R7_ye2023neural}, NDT3~\cite{R110_ye2025ndt3}, and their scaled-up counterparts~\cite{R8_azabou2023unified, R9_zhang2024universal, R10_azabou2024multi}. However, the consistent separation of single units and multi-units becomes increasingly difficult over time~\cite{R7_ye2023neural, R8_azabou2023unified} due to electrode migration, encapsulation, and loss of neighboring neurons. Another important consideration is that all these neural decoding models rely on extracellular spikes recorded at rates around $30$\,kHz~\cite{R2_pandarinath2018inferring, R110_ye2025ndt3}, creating substantial demands on power consumption and data bandwidth for wireless iBCI systems. These requirements translate into a sustained power budget of tens of milliwatts, which exceeds what an inductively powered, transcutaneously driven implant can safely dissipate. Most of these models use bidirectional encoders~\cite{R2_pandarinath2018inferring, R29_hurwitz2021targeted, R8_azabou2023unified}, often trained with masked autoencoding~\cite{R4_ye2021representation, R31_le2022stndt, R9_zhang2024universal}, allowing each output token to depend on both past and future contexts. How to transfer such models from offline analysis to online decoding therefore remains an open question. Consequently, both the high resource requirements and the long-term instability of spike signals pose fundamental barriers to continuous, fully implantable, and wireless iBCI devices.

Local field potentials (LFPs) are a promising signal source for long-term, wireless, real-time iBCIs. They measure the aggregated synaptic currents from local neural populations~\cite{R18_buzsaki2012origin, R11_pesaran2018investigating}, providing a complementary view of neural dynamics at a mesoscopic scale. LFPs offer several practical advantages that address key limitations of spike-based systems. First, LFP signals can remain stable over months to years because they reflect the collective activity of neural populations rather than individual neurons. They may therefore be less sensitive to electrode micromotion, tissue encapsulation, and gradual neuronal loss~\cite{R13_wang2014long, R16_sharma2015time, R17_flint2016long}. In practice, chronically implanted silicon arrays have yielded motor-cortical LFP recordings that remained stable over intervals from three days to more than three years~\cite{R17_flint2016long}. LFPs also retained kinematic information in an animal from which no spikes could be recorded after $671$ days~\cite{R13_wang2014long}. Second, LFPs occupy a much lower frequency range than extracellular spikes: LFP activity is typically below $300$\,Hz, so a $1$\,kHz sampling rate suffices, which lowers the demands on analog-to-digital conversion, data transmission, and on-chip processing, enabling a dramatic reduction in power consumption from tens of milliwatts to the sub-milliwatt regime that is compatible with on-skull energy harvesting, inductive recharging, or thermoelectric scavenging~\cite{R14_stavisky2015high, R44_jackson2017decoding}. This work takes the bandwidth constraint further and uses only the $0.05$--$50$\,Hz band at $100$\,Hz (section~\ref{subsubsec:preprocessing}), a deliberate trade-off between the information retained and the data rate a wireless implant must sustain. Finally, the lower data rates of LFP signals facilitate real-time decoding using microcontrollers, small field-programmable gate arrays, or low-power systems-on-a-chip such as the NVIDIA Jetson Orin Nano or Raspberry Pi 5. This allows decoding to run on-device in a wearable unit carried by the patient~\cite{R15_slutzky2017physiological}. Together, these characteristics make LFP recordings a practical and scalable signal modality for next-generation iBCIs, particularly in settings that require chronic implantation, wireless operation, and energy-efficient real-time decoding.

\begin{figure}[t]
  \centering
  \includegraphics[width=\textwidth]{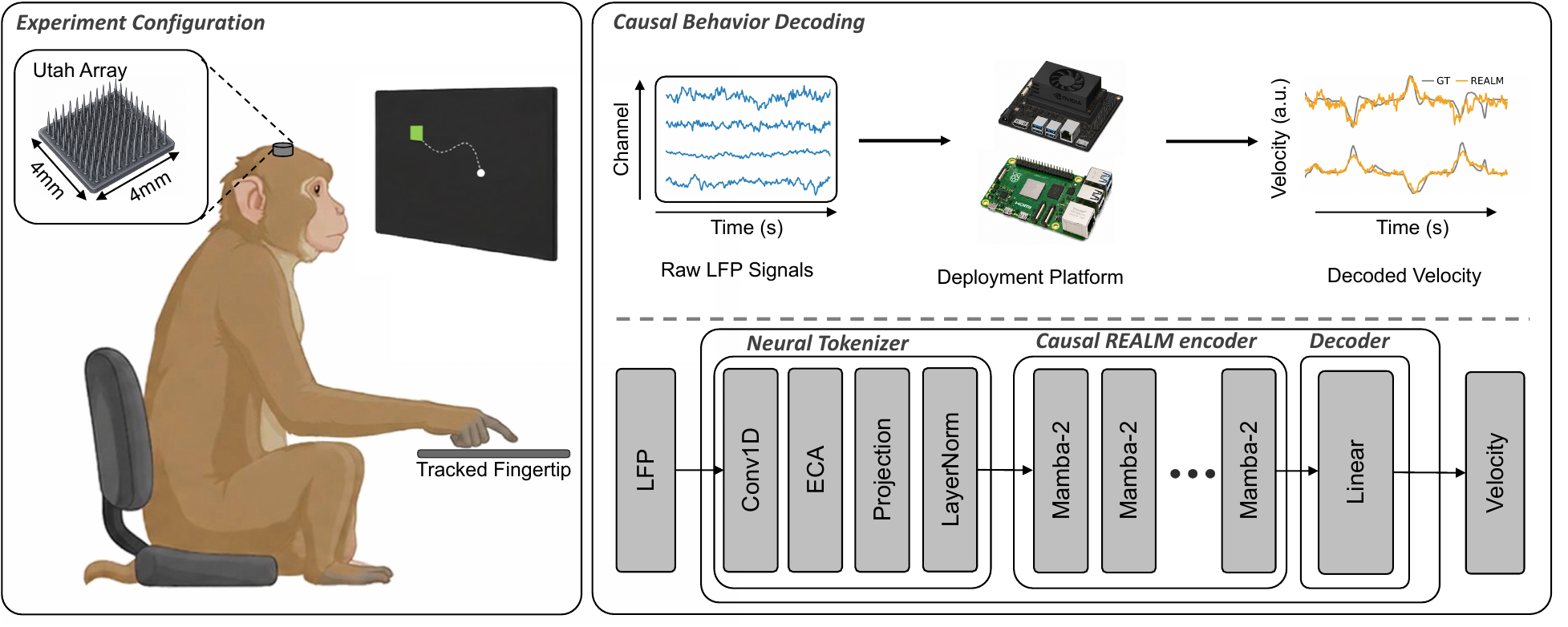}
  \caption{Overview of the REALM framework for causal behavior decoding from local field potentials (LFPs).   Left: Experimental configuration, illustrated for Makin (Monkey~I). A rhesus macaque performs a 2D motor task:   continuous random-target reaching for Makin or center-out reaching for Flint,   driving a cursor toward visual targets while LFPs are recorded from primary motor cortex (M1) via a chronically   implanted 96-channel Utah array ($4\,\text{mm} \times 4\,\text{mm}$). In Makin, the cursor follows the tracked fingertip; in Flint (Monkey~C), the monkey moves the cursor with the handle of a two-link planar manipulandum. Right:   Causal behavior-decoding pipeline. Raw multi-channel LFP signals are converted into   neural tokens by the neural tokenizer, encoded into temporally causal latent   representations by the REALM encoder, which was distilled from a non-causal teacher, and mapped to 2D velocity by a linear decoder. The full causal student is designed for real-time deployment   on edge hardware, illustrated here on the NVIDIA Jetson Orin Nano or Raspberry Pi 5.}
  \label{fig:entire_configuration}
\end{figure}

Despite these advantages, LFP-based decoding has historically been less accurate than spike-based decoding, since spatial averaging results in a lower signal-to-noise ratio than that of the localized activity captured by isolated spikes~\cite{R18_buzsaki2012origin, R20_belitski2008low, R12_abbaspourazad2021multiscale, R22_gallego2022local, R19_bansal2011relationships, R23_ahmadipour2024multimodal}. Prior efforts to decode behavior from LFPs have employed a range of approaches, including linear Kalman filters~\cite{R14_stavisky2015high}, recurrent architectures such as long short-term memory (LSTM) networks~\cite{R45_ahmadi2019decoding}, multiscale joint spike--LFP models~\cite{R12_abbaspourazad2021multiscale, R43_hsieh2018spikefield}, and engineered feature extractors such as Dual-VCT~\cite{R111_li2026dualvct}. However, these models are typically trained and evaluated within single sessions, rely on spike supervision during training, or lack a causal formulation. More recently, CrossModalDistill~\cite{R24_erturk2025cross} distills the representations learned by a Transformer teacher pretrained on spiking activity into an LFP student. However, in its published pipeline it still requires sorted spikes from the test sessions to fine-tune the model and fits different sorted units in different sessions. Such dependencies substantially limit its practicality, particularly in clinical scenarios where spike signals may degrade over time, intracortical contact is not feasible, or spike data are simply unavailable. Moreover, the strongest spike decoders~\cite{R7_ye2023neural, R110_ye2025ndt3, R8_azabou2023unified, R9_zhang2024universal, R10_azabou2024multi} have already measured what causality costs: NDT2 is pretrained with a causal attention mask, as a real-time iBCI would require, and reports that bidirectional attention nonetheless improves its modeling. The accuracy cost of causal decoding has not yet been systematically evaluated for LFP-based models.  Dual-VCT's variational mode decomposition solves a fixed point over the whole segment, and its attention is unmasked~\cite{R111_li2026dualvct}; CrossModalDistill pretrains its encoder by masked autoencoding, reconstructing dropped tokens from the unmasked ones across space and time~\cite{R24_erturk2025cross}. Both approaches can condition their outputs on future samples that are unavailable to a streaming decoder. The offline-to-online deployment gap that the speech-recognition community has grappled with for over a decade~\cite{R102_doutre2021improving, R103_moritz2021dual, R104_tang2023reducing} remains entirely open for intracortical signals. To our knowledge, no LFP decoder has been demonstrated to achieve causal, real-time decoding on a portable, low-power computing platform.

In this work, we investigate whether \emph{LFPs, used as the sole neural input modality, can support high-fidelity motor decoding in real-time, resource-constrained iBCIs, particularly fully implantable, battery-free wireless systems.} We introduce REALM (Retrospective Encoder Alignment for LFP Modeling), a retrospective knowledge distillation (RKD) framework that transfers representations from a bidirectional, non-causal teacher to a strictly causal student using LFP data alone. Figure~\ref{fig:entire_configuration} gives an overview of the experimental setup and the decoding pipeline. This framework's practical and technical contributions are summarized as follows:

\begin{enumerate}
\renewcommand{\labelenumi}{(\roman{enumi})}
\item Transfer across recording sessions. We develop a session-agnostic neural tokenizer consisting of a shared per-channel temporal convolution, efficient channel attention (ECA), and a value embedding. It enables direct transfer of a frozen encoder to sessions never seen during pretraining, with no session-specific parameters: it needs none of the per-session read-in layers or session embeddings that multi-session models normally require.
\item Causal representation learning. We establish an RKD framework that aligns a strictly causal student with the contextual representations of a non-causal teacher. This approach addresses the trade-off existing LFP decoders face between using the full temporal context and remaining deployable, while narrowing the accuracy cost of causality in behavior decoding.
\item Real-time edge deployment. We deploy the causal REALM end-to-end on low-power edge accelerators, which addresses the deployment limitation that non-causal decoders leave unresolved, and realizes streaming inference at the full acquisition rate: one kinematic estimate per incoming sample at $100$\,Hz on an NVIDIA Jetson Orin Nano and, in its compact form, on a Raspberry Pi~5.
\end{enumerate}

\section{Method}\label{sec:method} 

REALM is built on three stages, beginning with self-supervised pretraining of a bidirectional Mamba-2 (BiMamba-2) teacher using continuous masked autoencoding (CMAE) on $5.9$\,h of M1 LFP recordings from $34$ non-held-out Makin and Flint sessions, followed by RKD into a compact causal Mamba-2~\cite{R60_dao2024transformers} student. The third stage, described in section~\ref{subsec:protocol}, evaluates the models on held-out sessions using either linear-probe evaluation, with the encoder frozen, or the per-session supervised fine-tuning pipeline. The complete set of hyperparameters for every stage, together with its hardware and wall-clock time, is provided in Supplementary Section~S2.

\subsection{Datasets and LFP preprocessing}\label{subsubsec:preprocessing}

REALM uses two publicly available intracortical datasets: Makin~\cite{R94_makin2018superior, R93_odoherty2017nonhuman}, recorded from Monkey~I, and Flint~\cite{R95_flint2012accurate}, recorded from Monkey~C. Both datasets were recorded with chronic Utah arrays in the primary motor cortex (M1) and include continuous kinematics: cursor position for Makin and hand velocity for Flint. In Makin, the monkey uses its tracked fingertip to drive a cursor to targets on an $8\times 8$ grid in most sessions ($17\times 8$ in one session). Targets are presented consecutively, with no inter-trial interval. The monkey acquires each target by holding its fingertip on it for $450$\,ms, resulting in continuous, self-paced reaches. Flint records center-out reaching in a trial-structured format, in which the monkey moves the cursor by grasping the handle of a two-link planar manipulandum. Supplementary Section~S3 and Supplementary Table~S18 report subject counts, channel counts, session counts, and recording durations. 

The two corpora are released at different sampling rates. Makin provides broadband neural recordings at $f_0\approx24.4$\,kHz and kinematics sampled at $250$\,Hz, whereas Flint provides neural recordings at $2$\,kHz and kinematics sampled at $100$\,Hz. We place the neural signal and the kinematics from both datasets on a common $100$\,Hz grid, matching the lower of the two kinematic sampling rates without upsampling either measured kinematic signal. The $0.05$--$50$\,Hz band and the $100$\,Hz grid are design choices for fully implantable wireless iBCI systems. (i) A $100$\,Hz LFP signal has a lower bandwidth requirement, so a fully implantable iBCI can transmit neural signals to an external inference platform or process them by itself. (ii) The grid lets the model output at time index $t$ be compared directly with the velocity at the same index without further resampling or an imposed lag. This is a convenience rather than a necessity, since a higher-rate input could also be mapped to the kinematic rate by temporal aggregation or learned downsampling. (iii) Most LFP power lies below $50$\,Hz, the Nyquist frequency of the grid, but this does not establish that most of the behaviorally informative content does, and activity above $50$\,Hz, including the high-gamma band, is excluded by this choice. Retaining it would require either a higher analysis rate or band-power features on the $100$\,Hz grid, and band-power features would reintroduce the hand-designed feature extraction that this work avoids. 

After preprocessing, we divide each recording into $5$\,s segments. The segment length fixes the context of every model in the comparison: all states are reset at the start of each segment, so every architecture decodes from the same $5$\,s of history, whereas longer segments would give the recurrent Mamba-2 encoder access to more history than the Transformer baseline attends to within its training window, and would raise the quadratic training cost of self-attention. Segments do not overlap during RKD or evaluation. For masked pretraining, we also sample segments at a $2.5$\,s stride, producing $50\%$ overlap as data augmentation (Supplementary Table~S11). Each segment is $\mathbf{X}\in\mathbb{R}^{C\times T}$, with $T=500$ samples at $f_s=100$\,Hz, and a target $\mathbf{v}_t\in\mathbb{R}^2$ at each index $t$. The arrays differ in channel count, $C=96$ on Makin against $C=95$ on Flint, so a Flint segment is zero-padded to $96$ rows and every segment carries a binary channel mask that is $1$ on recorded channels and $0$ on the padding. The two corpora differ only in how they reach this form:

\emph{Monkey I: LFP extraction.} The Makin recording is released only as a $24.4$\,kHz broadband signal, so the LFP has to be extracted first. Each channel is decimated to $763$\,Hz in three stages by factors of $4$, $4$ and $2$, each stage applying an eighth-order Chebyshev type-I anti-alias low-pass filter at $0.8$ of that stage's output Nyquist frequency before samples are discarded. Decimating in stages leaves each filter with a wide transition band relative to its input rate, whereas a single sharp filter would ring. This leaves Makin as $96$ channels sampled at $763$\,Hz.

\emph{Monkey C: trial assembly.} Flint is released as a $2$\,kHz neural signal stored trial by trial, so no LFP extraction from broadband recordings is needed. Consecutive trials are separated by at most one sample of the $100$\,Hz kinematic clock. However, the median trial lasts only $3.9$\,s, and only $5\%$ of trials reach $5$\,s. Restricting $5$\,s segments to individual trials would therefore retain less than one-tenth of the recording. For each trial, we retain the hand velocity and all LFP channels containing data, truncating the channels to a common length. Trials with no populated LFP channel are excluded. We concatenate the retained trials in recording order \emph{before} filtering, so that the notch, band-pass and anti-alias filters operate on a continuous record instead of introducing a filter transient at every trial boundary. Because velocity timestamps are irregular at the concatenation points, we interpolate the concatenated velocity onto a uniform $100$\,Hz grid spanning the session. The assembled Flint recording contains $95$ channels sampled at $2$\,kHz.

From here, the two corpora are treated identically, each at its own intermediate rate: (i) Each channel has its mean removed and is then notched at $60$\,Hz and at every harmonic of it below Nyquist by a second-order IIR section of quality factor $30$. (ii) The channel is band-limited to $[0.05,50]$\,Hz by second-order Butterworth high-pass and low-pass sections. The lower edge keeps the local motor potential, the slowest movement-related component of the field potential and a decoding feature in its own right~\cite{R95_flint2012accurate}, while removing drift with periods longer than $20$\,s. (iii) The array mean is subtracted at each time point (common average reference). (iv) The channels are resampled to $100$\,Hz and the velocity is placed on the same time base. The resampling is done in the Fourier domain over the whole record, which discards every component above the $50$\,Hz Nyquist frequency of the target grid; this, rather than the low-pass of step (ii), is what makes the step alias-free, since a second-order section with its edge at $50$\,Hz only attenuates the band toward that frequency. (v) Each channel is standardized to $\tilde{x}_c[t]=(x_c[t]-\mu_c)/\sigma_c$, where $\mu_c$ and $\sigma_c$ are that channel's mean and standard deviation over the session and $\sigma_c$ is set to $1$ on a flat channel. Zero-padding to $96$ rows follows step (v) on Flint, so the padded row is identically zero and enters no statistics.

For evaluation of the causal student, the data preprocessing pipeline should also be strictly causal, and the three steps of this chain that read later samples are replaced as follows. All zero-phase filters of steps (i) and (ii) are run forward only, with the high-pass started from the steady state of the first sample and the $50$\,Hz low-pass replaced by a minimum-phase FIR filter. The whole-record resampling of step (iv) becomes staged decimation with causal anti-aliasing filters. The session statistics of step (v) become a running estimate: each sample is standardized with the mean and standard deviation of all earlier samples of the session, and the first $5$\,s, over which this estimate builds up, are set to zero. Supplementary Section~S4.3 gives the details and compares causal deployment with zero-phase evaluation.

The behavioral target is brought onto the same grid. Makin provides cursor position at $250$\,Hz, and velocity is taken as its central finite difference and resampled to $100$\,Hz; Flint supplies velocity directly at $100$\,Hz, assembled trial by trial as described under \emph{Monkey C: trial assembly} above. The two corpora are aligned in opposite directions: in Makin the velocity's time vector defines the grid and the LFP is interpolated onto it, trimmed to the interval the two recordings share; in Flint the grid spans the neural record and the velocity is interpolated onto it. In both, input sample $t$ is paired with velocity sample $t$, without an imposed lead or lag between them. 

\subsection{Neural tokenizer module}

The first component of REALM is an LFP-specific neural tokenizer that extracts temporal and spatial features. A shared temporal convolution processes each channel, ECA weights the resulting features, and a value embedding maps them into tokens. The tokenizer has no session-specific parameters, so its pretrained weights can be applied directly to LFP recordings from a new session.

\begin{figure}[t]
  \centering
  \includegraphics[width=\textwidth]{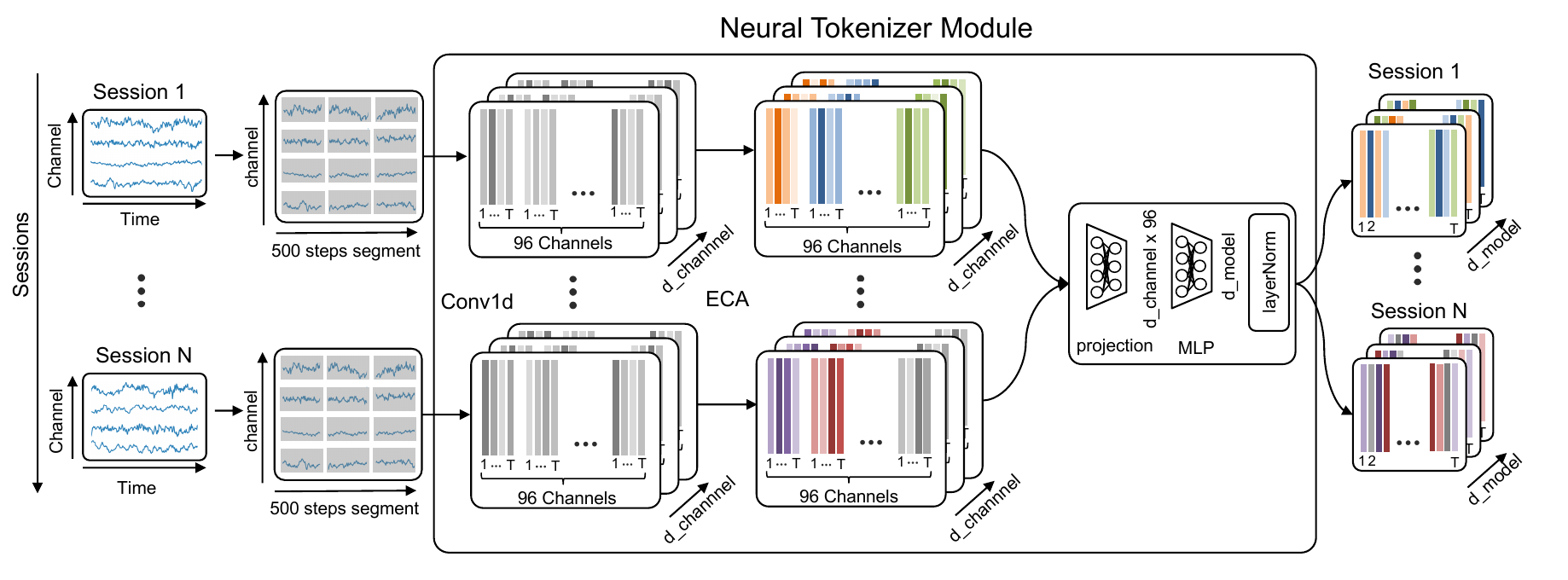}
  \caption{Tokenization pipeline for multi-session LFP signals. Raw signals are first segmented into 500-timestep windows. A per-channel 1D convolution followed by an efficient channel attention (ECA) module produces channel-attended features of shape $(B, 96, d_{\text{ch}}, T)$, where $B$ is the batch size, $96$ is the number of channels, $d_{\text{ch}} = 8$ is the number of features per channel, and $T = 500$ is the number of timesteps. These features are then mapped through a value embedding, which yields the final token representations. No session identifier enters the tokenizer, so the same weights apply to every session, including ones never seen during pretraining.}
  \label{fig:neuro_tokenizer_module}
\end{figure}

\subsubsection{Per-channel temporal convolution}

As shown in figure~\ref{fig:neuro_tokenizer_module}, the neural tokenizer converts raw LFP signals into a sequence of token embeddings in three stages. First, the raw LFP input $\mathbf{X} \in \mathbb{R}^{B \times C \times 1 \times T}$ ($C{=}96$ channels, $T{=}500$ timesteps sampled at $100$\,Hz) is processed by a shared per-channel Conv1d layer that expands the one-dimensional raw-voltage feature to $d_{\text{ch}}{=}8$ features per channel. Concretely, with kernel size $K{=}3$, stride $1$ and zero-padding $1$, the per-channel temporal embedding is computed as
\begin{equation}
    \mathbf{H}_{b,c,d,t}
    = \mathrm{GELU}\!\left(
        \sum_{k=0}^{K-1} \mathbf{W}_{d,k}\,
        \mathbf{X}_{b,c,1,\,t+k-\lfloor K/2 \rfloor}
        + \mathbf{b}_{d}
      \right),
\label{eq:per_channel_tcn}
\end{equation}
for $d \in \{1,\dots,d_{\text{ch}}\}$ and $t \in \{1,\dots,T\}$, where $\mathbf{W} \in \mathbb{R}^{d_{\text{ch}} \times K}$ and $\mathbf{b} \in \mathbb{R}^{d_{\text{ch}}}$ are \emph{shared across all $C$ channels}, yielding $\mathbf{H} \in \mathbb{R}^{B \times C \times d_{\text{ch}} \times T}$. This lightweight convolution captures local temporal patterns, including oscillatory bursts and transient waveforms, while keeping the model compact through parameter sharing. In the causal student, the same kernel is instead applied with $K-1$ zeros of left padding, so that $\mathbf{H}_{b,c,d,t}$ depends on $\mathbf{X}_{b,c,1,t-K+1},\dots,\mathbf{X}_{b,c,1,t}$ only and no future sample enters the tokenizer.

\subsubsection{Efficient channel attention}
Next, to weight each channel adaptively, we apply an ECA module~\cite{R84_wang2020eca} along the channel dimension. The ECA block first computes a descriptor for each channel by averaging $\mathbf{H}$ over the temporal ($T$) and feature ($d_{\text{ch}}$) dimensions, which gives a channel-energy vector $\mathbf{e} \in \mathbb{R}^{B \times C}$:
\begin{equation}
    \mathbf{e}_{b,c} = \frac{1}{d_{\text{ch}}\,T} \sum_{d=1}^{d_{\text{ch}}} \sum_{t=1}^{T} \mathbf{H}_{b,c,d,t}.
\label{eq:eca_energy}
\end{equation}
In the causal variant, the average over the full window $[1, T]$ is replaced by a running mean over $[1, t]$, yielding a time-indexed descriptor $\mathbf{e}_t \in \mathbb{R}^{B \times C}$ that depends only on past observations and is updated incrementally as new samples arrive:
\begin{equation}
    \mathbf{e}_{b,c,t} = \frac{1}{d_{\text{ch}}\,t} \sum_{d=1}^{d_{\text{ch}}} \sum_{\tau=1}^{t} \mathbf{H}_{b,c,d,\tau}.
\label{eq:eca_energy_causal}
\end{equation}
A 1D convolution over $\mathbf{e}$ followed by a sigmoid then gives channel-level attention coefficients $\mathbf{a} \in [0,1]^{B \times C}$:
\begin{equation}
    \mathbf{a}_{b,c} = \sigma\!\left( \sum_{j=0}^{k-1} \boldsymbol{\omega}_{j}\, \mathbf{e}_{b,\,c+j-\lfloor k/2 \rfloor} \right),
\label{eq:eca_attn}
\end{equation}
where $\boldsymbol{\omega} \in \mathbb{R}^{k}$ are the shared 1D-convolution weights with kernel size $k{=}5$ (zero-padded at the channel boundaries), and $\sigma(\cdot)$ denotes the sigmoid function. The input features are then reweighted as $\tilde{\mathbf{H}}_{b,c,d,t} = \mathbf{a}_{b,c}\,\mathbf{H}_{b,c,d,t}$, i.e., $\tilde{\mathbf{H}} = \mathbf{a} \odot \mathbf{H}$ with broadcasting along the $d_{\text{ch}}$ and $T$ dimensions. In the causal variant, equation~(\ref{eq:eca_attn}) is instead evaluated at every step $t$ on the causal descriptor of equation~(\ref{eq:eca_energy_causal}), giving time-indexed coefficients $\mathbf{a}_{b,c,t} = \sigma\big( \sum_{j=0}^{k-1} \boldsymbol{\omega}_{j}\, \mathbf{e}_{b,\,c+j-\lfloor k/2 \rfloor,\,t} \big)$, and the features are reweighted as $\tilde{\mathbf{H}}_{b,c,d,t} = \mathbf{a}_{b,c,t}\,\mathbf{H}_{b,c,d,t}$, with broadcasting along $d_{\text{ch}}$ only, so the coefficient applied at step $t$ uses no later sample. This allows the model to emphasize informative channels and suppress irrelevant ones.

\subsubsection{Value embeddings}

Finally, a \emph{value embedding} maps the channel-attended features of each timestep to the model dimension. It flattens $\tilde{\mathbf{H}}$ across channels and features ($C \times d_{\text{ch}} = 768$) and projects the result with a fully connected linear layer to $d_{\text{model}}=256$ dimensions, followed by LayerNorm, giving the tokens $\mathbf{V} \in \mathbb{R}^{B \times T \times d_{\text{model}}}$. Concretely, let $\tilde{\mathbf{h}}_{b,t} \in \mathbb{R}^{C\,d_{\text{ch}}}$ denote the flattened per-timestep feature, with $\tilde{\mathbf{h}}_{b,t,\,(c-1)d_{\text{ch}}+d} = \tilde{\mathbf{H}}_{b,c,d,t}$; the value tokens are then
\begin{equation}
    \mathbf{V}_{b,t} = \mathrm{LayerNorm}\!\left( \mathbf{W}_{v}\,\tilde{\mathbf{h}}_{b,t} + \mathbf{b}_{v} \right), \quad \mathbf{W}_{v} \in \mathbb{R}^{d_{\text{model}} \times C\,d_{\text{ch}}},\ \mathbf{b}_{v} \in \mathbb{R}^{d_{\text{model}}}.
\label{eq:value_embed}
\end{equation}
The value tokens are the final input tokens, $\mathbf{Z} = \mathbf{V} \in \mathbb{R}^{B \times T \times d_{\text{model}}}$; no session-dependent term is added.

\subsection{Continuous masked LFP teacher model pretraining}\label{subsec:pretrain}

\begin{figure}[t]
  \centering
  \includegraphics[width=\textwidth]{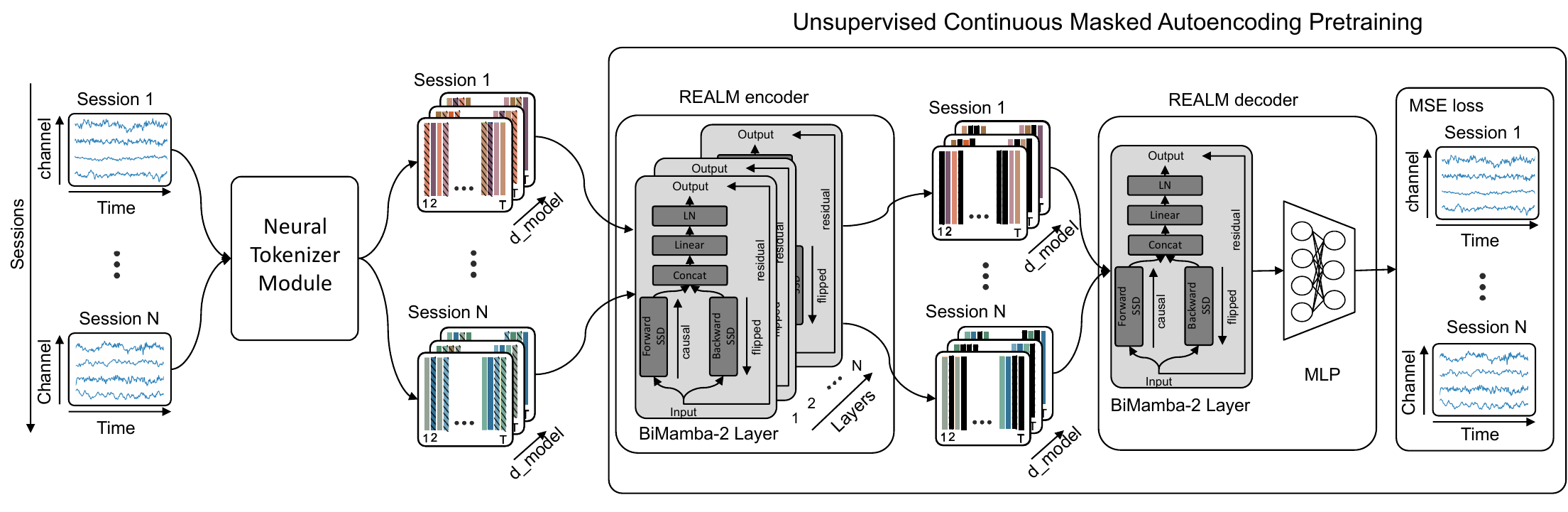}
  \caption{Architecture of the multi-session pretrained bidirectional Mamba-2 (BiMamba-2) teacher. For self-supervised pretraining, the model reconstructs the raw LFP signals from their latent representations by continuous masked autoencoding (CMAE). This allows the model to learn representations shared across the $34$ non-held-out Makin and Flint sessions, recorded from two monkeys during random-target and center-out reaching. The CMAE predictor, labeled REALM decoder in the figure (one BiMamba-2 layer followed by a two-layer fully connected network, drawn as MLP), is lightweight relative to the teacher encoder (10 layers). MSE: mean squared error.}
  \label{fig:pretrain}
\end{figure}

After tokenization, we pretrain the multi-session teacher with the self-supervised CMAE objective shown in figure~\ref{fig:pretrain}. Given the token sequence $\mathbf{Z} \in \mathbb{R}^{B \times T \times d_{\text{model}}}$ produced by the neural tokenizer, we generate a binary mask $\mathbf{m} = (m_1, \ldots, m_T)$ with $m_i \in \{0, 1\}$ using a continuous block masking strategy: blocks of $l \sim \text{Uniform}(10, 50)$ contiguous timesteps are placed iteratively until a proportion $r{=}0.6$ of tokens is masked ($m_i{=}1$). Compared with the random masking of the standard masked autoencoder~\cite{R80_he2022masked}, continuous block masking prevents naive interpolation from neighboring visible tokens and forces the encoder to learn long-range temporal dependencies, which is particularly important given the smooth, temporally correlated nature of LFP signals. Masked tokens are replaced with a learnable \texttt{[MASK]} embedding $\mathbf{e}_M \in \mathbb{R}^{d_{\text{model}}}$, forming the corrupted input $\mathbf{Z}^{\mathbf{m}} \in \mathbb{R}^{B \times T \times d_{\text{model}}}$.

To improve robustness across recording sessions, we also augment the pretraining inputs of the unmasked tokens with channel dropout ($p{=}0.15$), per-channel amplitude jitter with a scale factor drawn from $\text{Uniform}(0.85, 1.15)$, and additive Gaussian noise ($\sigma{=}0.05$). The reconstruction target, however, remains the \emph{original, unaugmented} signal. The augmentation increases the effective number of distinct training segments and discourages the model from memorizing segment-specific artifacts.

\subsubsection{BiMamba-2 encoder}
The teacher encoder is built from BiMamba-2 layers and the causal student encoders from unidirectional Mamba-2 layers. We choose Mamba-2 for three reasons. First, REALM must run in real time, and Mamba-2 is recurrent, so its per-step cost and memory do not grow with the context. Second, its training cost grows linearly with sequence length, whereas that of self-attention grows quadratically. Third, Mamba-2 carries an explicit hidden state $h_t$ that summarizes the signal history. Supplementary Section~S4.1 compares Mamba-2 with an LSTM and a Transformer distilled through the same pipeline at matched capacity (Supplementary Table~S19): the Transformer is not separated from Mamba-2 in accuracy at $n=8$ in either pipeline, so the choice rests on the reasons above rather than on accuracy.

A unidirectional Mamba-2 block $\mathcal{M}(\cdot)$ takes $X \in \mathbb{R}^{B \times T \times d_{\text{model}}}$, splits it into a content stream and a gating stream, mixes neighboring timesteps with a depthwise 1D convolution, and feeds the result into a state-space-duality (SSD) kernel. Inside the SSD kernel, each of the $H$ heads runs an input-dependent linear recurrence with a scalar decay. For head $h$ at step $t$, the discretized state transition is
\begin{align}
\bar{A}_{h,t} &= \exp\!\big(A_h \, \Delta_t\big), \\
A_h &= -\exp(A_h^{\log}) < 0, \\
\Delta_t &= \mathrm{softplus}(W_{\Delta} u_t)_h,
\end{align}
so that $\bar{A}_{h,t} \in (0,1)$ acts as a per-step forgetting factor whose strength is predicted from the input $u_t$. Two further input-dependent vectors $B_t, C_t \in \mathbb{R}^{N}$ play the role of write and read keys, and we apply rotary position embeddings (RoPE)~\cite{R86_su2024roformer} inside the SSD kernel to inject relative position information along the temporal axis,
\begin{align}
\tilde{B}_t &= \mathrm{RoPE}(B_t,\, t), \\
\tilde{C}_t &= \mathrm{RoPE}(C_t,\, t),
\end{align}
which we found important for transferring across sessions whose absolute time origins are arbitrary. The hidden state $h_{h,t} \in \mathbb{R}^{P \times N}$ and output $y_{h,t} \in \mathbb{R}^{P}$ then evolve as
\begin{align}
h_{h,t} &= \bar{A}_{h,t}\, h_{h,t-1} + u_{h,t}\, \tilde{B}_t^{\top}, \\
y_{h,t} &= h_{h,t}\, \tilde{C}_t + D_h\, u_{h,t},
\end{align}
where $P$ is the per-head channel width and $D_h$ is a learnable skip gain. Although written sequentially, this recurrence is computed during training by the chunked SSD algorithm of Mamba-2, whose time and memory are linear in $T$, so an entire 500-sample window is processed in parallel. Concatenating $y_{h,t}$ across heads yields $\mathrm{SSD}(U)_t$, which, after the output projection, gives the block output $\mathcal{M}(X)$.

Each BiMamba-2 layer wraps two independently parameterized Mamba-2 layers $\mathcal{M}_{\rightarrow}, \mathcal{M}_{\leftarrow}$ that process the sequence in opposite temporal directions, and fuses the two streams through a linear projection followed by LayerNorm:
\begin{equation}
\left\{
\begin{aligned}
\overrightarrow{H} &= \mathcal{M}_{\rightarrow}(X), \\
\overleftarrow{H}  &= \mathrm{Flip}_{T}\!\big(\mathcal{M}_{\leftarrow}(\mathrm{Flip}_{T}(X))\big),
\end{aligned}
\right.
\end{equation}
\begin{equation}
\mathrm{BiMamba2}(X) = \mathrm{LayerNorm}\!\big(W_{m}\,[\overrightarrow{H} \,\Vert\, \overleftarrow{H}]\big),
\end{equation}
where $\mathrm{Flip}_{T}$ reverses the time axis, $[\cdot\Vert\cdot]$ concatenates along channels, and $W_{m} \in \mathbb{R}^{d_{\text{model}} \times 2d_{\text{model}}}$ fuses the two directions back to $d_{\text{model}}$. Bidirectionality is essential for masked LFP reconstruction, which is intrinsically non-causal: predicting a masked block at time $t$ should draw on both the past \emph{and} future context surrounding it.

\subsubsection{Reconstruction head and objective}
The corrupted sequence $\mathbf{Z}^{\mathbf{m}}$ is first processed by the BiMamba-2 encoder (10 layers, expand${=}2$). After the encoder, a lightweight asymmetric predictor comprising a single BiMamba-2 layer followed by a two-layer fully connected network maps the encoded tokens back toward the signal space. A final linear head projects each token to the per-channel LFP values:
\begin{equation}
  \hat{\mathbf{x}}_i = \mathrm{Linear}\big(\mathrm{Predictor}\big(\mathrm{Encoder}(\mathbf{Z}^{\mathbf{m}})\big)\big)_i, \quad \hat{\mathbf{x}}_i \in \mathbb{R}^{C}.
\end{equation}
The training objective is the mean squared error at masked positions only:
\begin{equation}
  \mathcal{L}_{\text{CMAE}} = \frac{1}{\sum_{i=1}^{T} m_i} \sum_{i=1}^{T} m_i \, \| \hat{\mathbf{x}}_i - \mathbf{x}_i \|_2^2,
\end{equation}
where $\mathbf{x}_i \in \mathbb{R}^{C}$ denotes the original, unaugmented LFP values at timestep $i$. Restricting the loss to masked positions prevents the encoder from taking a shortcut by copying visible tokens through the predictor and concentrates the gradient signal on positions that genuinely require inference from context.

\subsection{Retrospective knowledge distillation (RKD)}
\label{sec:distill}

\begin{figure}[t]
  \centering
  \includegraphics[height=0.3264\textwidth]{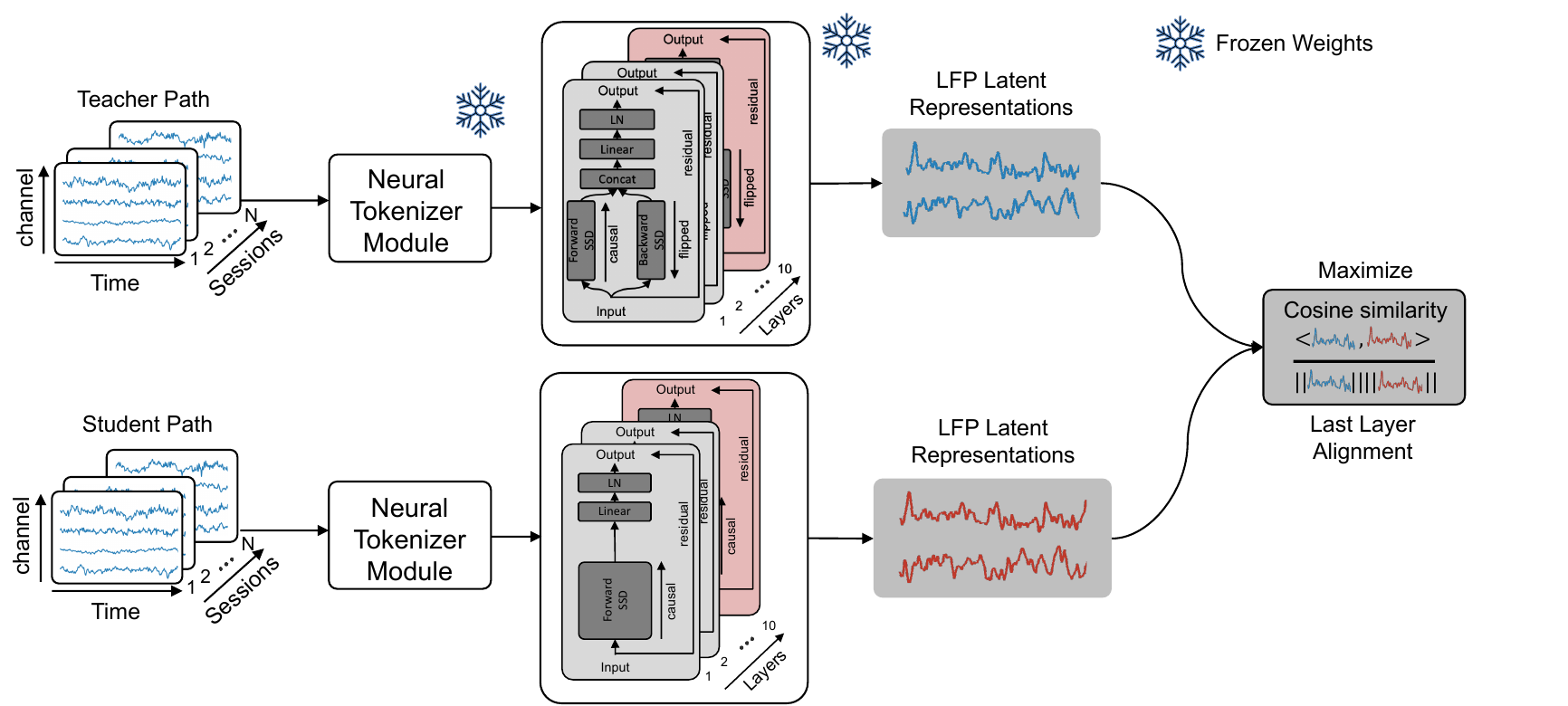}
  \caption{Cross-session distillation from a frozen 10-layer BiMamba-2 teacher to a 10-layer causal Mamba-2 student. Both paths use the neural tokenizer module (the student's copy is initialized from the teacher's and runs in causal form) and process multi-session LFP inputs, producing last-layer latent representations that are aligned by maximizing cosine similarity. The causal student thereby inherits the teacher's bidirectional context while remaining deployable in a real-time intracortical brain--computer interface (iBCI).}
  \label{fig:distill}
\end{figure}

After pretraining the BiMamba-2 teacher on multi-session LFP data, we transfer its knowledge to a compact causal Mamba-2 student by RKD in the latent space, as depicted in figure~\ref{fig:distill}. The distillation objective comprises two terms: (i) a cosine representation alignment loss that aligns the last encoder layer of the student with that of the teacher, and (ii) a reconstruction loss that recovers the LFP input from the student's causal representation.

The teacher is distilled separately on the non-held-out sessions of Makin and on those of Flint, giving one student per corpus (and per seed; section~\ref{subsec:protocol}), and each held-out session is decoded by the student of its own corpus. The student's neural tokenizer (the per-channel temporal convolution, ECA, and the linear projection and LayerNorm of the value embedding) is initialized as a copy of the teacher's and applied in its causal form, while the rest of the student, its Mamba-2 layers included, is initialized at random; all student parameters, the tokenizer included, are then trained.

The alignment acts only on the final encoder layer. Last-layer alignment distills enough knowledge while giving the student freedom to learn a causal representation (Supplementary Table~S20). Removing the term altogether costs $0.192$ ($[+0.087, +0.297]$, $d_z = 1.53$, $8/8$, $p = 0.003$) and returns the student to the level of the same backbone without RKD. The objective needs alignment, not a particular depth. The distillation loss is
\begin{equation}
\mathcal{L}_{\text{distill}} 
= \underbrace{\lambda_{\text{repr}} 
\left(1 - \frac{1}{T}\sum_{t=1}^{T}
\frac{\langle \mathbf{z}^{\mathrm{stu}}_t,\, \mathbf{z}^{\mathrm{tea}}_t \rangle}
{\|\mathbf{z}^{\mathrm{stu}}_t\|_2 \, \|\mathbf{z}^{\mathrm{tea}}_t\|_2}\right)}_{\text{Representation alignment}}
+ \underbrace{\lambda_{\text{ae}}\,\frac{1}{|\mathcal{A}|}\sum_{t\in\mathcal{A}}\|\mathbf{x}_t - f_\psi(\mathbf{z}^{\mathrm{stu}}_t)\|_2^2}_{\text{Reconstruction}},
\label{eq:distill}
\end{equation}
where $\mathbf{z}^{\mathrm{stu}}_t$ and $\mathbf{z}^{\mathrm{tea}}_t$ are the last-layer representations of the student and of the teacher at timestep $t$, $\mathbf{x}_t \in \mathbb{R}^{C}$ is the original LFP input, and $f_\psi(\cdot)$ is a reconstruction head of two linear layers with a GELU between them. The student's input is not masked: the reconstruction term is evaluated on a set $\mathcal{A}$ of $30\%$ of the $T$ timesteps, drawn at random for every segment at every training step (Supplementary Table~S12). The weights $\lambda_{\text{repr}}$ and $\lambda_{\text{ae}}$ set the contributions of the alignment and reconstruction terms. The teacher is not fine-tuned before distillation and stays frozen during it; only the student and $f_\psi$ are updated. We set $\lambda_{\text{repr}} = \lambda_{\text{ae}} = 1.0$ and $T = 500$ timesteps.

\section{Experimental setup}\label{sec:setup}

\subsection{Data splitting and evaluation protocol}\label{subsec:protocol}

We hold out 8 sessions for evaluation: 5 from Makin and 3 from Flint. These sessions are used in neither pretraining nor RKD. Within each held-out session, we divide the non-overlapping $5$\,s segments into training, validation, and test folds in a $72/8/20$ ratio. The partition is drawn once per random seed and shared by all models, including baselines (Supplementary Section~S3.1).  

Every REALM result in this paper comes from one of two evaluation pipelines. Both pipelines begin with the same CMAE pretraining and RKD stages and differ only in how the model is adapted and read out on a held-out session. We repeat each pipeline with three random seeds, $s \in \{42, 123, 456\}$, for REALM and every reproduced baseline.

(i) In linear-probe evaluation, the encoder is frozen and a closed-form ridge read-out (decoder) is fitted on each held-out session's training and validation folds, then scored once on the untouched test fold. The read-out regresses $\mathbf{v}_t$ on the encoder outputs at steps $t-10,\dots,t$, a past-only context of $11$ frames ($110$\,ms; Supplementary Table~S14). Velocity never reaches the encoder here; it enters only through the read-out. This pipeline evaluates the model's representation quality. Besides this fixed context, the read-out has one hyperparameter, the penalty $\alpha$. We select it per session and seed on the validation fold, from $\{1,10,10^2,10^3,10^4,10^5,10^6\}$, and refit the read-out with the chosen value before scoring; the test fold never enters the selection. The same rule applies to every model, so each baseline is read out at its own best $\alpha$ rather than at a single value fixed for all of them.

(ii) In the fine-tuned (supervised) pipeline, nothing is frozen except the neural tokenizer: the encoder is adapted on the training fold with velocity labels under a mean squared error loss. Early stopping and checkpoint selection use the validation fold, which is disjoint from the training and test folds, and the selected weights are scored once on the test fold. The read-out is a single linear layer from the encoder output at step $t$ to $\hat{\mathbf{v}}_t$. Supplementary Table~S15 lists the optimizer, schedule, and stopping criterion.

\subsection{Evaluation metric}\label{subsubsec:evaluation_metric}

We use $R^2$ as the evaluation metric. Unlike a correlation coefficient, which is invariant to gain and offset, $R^2$ penalizes predictions with the wrong amplitude or offset even when they follow the target's overall shape. A velocity decoder must get both the shape and the magnitude right. Root-mean-square error (RMSE) is in the units of the target, so it cannot be pooled across sessions and subjects with different velocity distributions. $R^2$ is computed as
\begin{equation}
R^2 \;=\; \frac{1}{2}\sum_{a\in\{x,y\}}
\left(
1-\frac{\sum_{t=1}^{N}\bigl(v_{t,a}-\hat{v}_{t,a}\bigr)^{2}}
       {\sum_{t=1}^{N}\bigl(v_{t,a}-\bar{v}_{a}\bigr)^{2}}
\right),
\qquad
\bar{v}_{a}=\frac{1}{N}\sum_{t=1}^{N} v_{t,a},
\label{eq:r2}
\end{equation}
where $\mathbf{v}_t=(v_{t,x},v_{t,y})\in\mathbb{R}^2$ denotes the measured cursor or hand velocity at step $t$ and $\hat{\mathbf{v}}_t$ the decoded velocity, with $t=1,\dots,N$ indexing the evaluation split of a single session. Velocity is $z$-scored within that session, so every quantity here is dimensionless. The denominator of equation~(\ref{eq:r2}) scores the decoder against the per-axis mean of the evaluation split, so $R^2=0$ means predicting that mean and $R^2<0$ means doing worse than it. Since $\bar{v}_a$ is computed on the scored samples themselves, this is the stricter of the two variants in common use. Like our primary baseline, CrossModalDistill~\cite{R24_erturk2025cross}, we average the two axes with equal weight rather than pooling them by variance.

\subsection{Statistical analysis}\label{subsubsec:stats}

Every statistical test in this paper is run at the session level rather than the segment level. Each held-out session is scored once per seed, yielding $24$ scores across eight sessions and three seeds. We first average the three seeds within each session. We then compare the two models session by session across those eight sessions ($n = 8$). All pairwise comparisons use a two-sided paired $t$-test on those eight differences, and the $p$-value, interval, and effect size all come from that one normal-theory model. Each comparison is reported with five numbers: the mean paired difference $\Delta R^2$, a $95\%$ confidence interval from the $t$ distribution on seven degrees of freedom, Cohen's $d_z$ (that mean difference divided by its standard deviation), the count of sessions on which the paired difference is positive, and $p$, printed to three decimals and as $p < 0.001$ below that. Differences and their statistics are computed from unrounded session means, so a printed difference can differ by $0.001$ from the difference of the printed means. Dispersion is the standard deviation: across the three seeds in per-session panels, and across sessions where a panel aggregates them (figures~\ref{fig:stacking}(f), \ref{fig:fewshot}(a, b) and \ref{fig:realtime}(a, b)); figure~\ref{fig:fewshot}(c) shows the $95\%$ confidence interval of the paired difference, and box plots span the interquartile range over the eight sessions $\times$ three seeds.

\subsection{Comparative experiment design}\label{subsec:comparisons}

We conduct two sets of comparisons. (i) We compare REALM with three members of its model family: REALM-PT, which is a pretrained non-causal teacher model; REALM-PT-c, which shares REALM's architecture and corpora but is trained by the teacher's masked pretraining; and REALM-RI, which has randomly initialized weights. (ii) We compare REALM with two temporal backbones (an LSTM and a Transformer) and four published methods (LFADS, DPAD, Dual-VCT and CrossModalDistill). All models are evaluated on the same held-out sessions using the $72/8/20$ data split and the same metric described in sections~\ref{subsec:protocol} and~\ref{subsubsec:evaluation_metric}. Within each evaluation pipeline, we use the same read-out protocol across models. We score every model on the full test fold, with one exception: DPAD's per-session fine-tuned fit is scored, and early-stopped, with the ten artifact segments of Supplementary Table~S17 dropped, because the transients they carry pull that fit down ($0.464$ on the full test fold against $0.494$). Apart from that exception, which favors DPAD, the protocol is the same for every model. How each baseline is fine-tuned is listed in Supplementary Table~S16: the LSTM, the Transformer and Dual-VCT with REALM's recipe (Supplementary Table~S15); LFADS as a whole model with a linear head; CrossModalDistill by its paper's procedure; and DPAD refitted on each session by its own method. The LSTM and Transformer backbones have a parameter capacity comparable to REALM's ($4.9$--$6.1$\,M) and go through the same pretraining and evaluation pipeline, without RKD. The published methods, like REALM, use no label of any session before the read-out, so Dual-VCT's first stage is trained by masked reconstruction of the LFP rather than by velocity regression; DPAD, whose latent states are defined by behavior, is the one exception. Supplementary Section~S2.1 compares this label-free first stage with Dual-VCT's original supervised one. CrossModalDistill is run in the multi-session form its paper also describes: one LFP student, distilled on the non-held-out sessions, so that, like every other model, it uses no spikes of the evaluated sessions. Every baseline is run in its causal form in figures~\ref{fig:causal_result} and~\ref{fig:finetuned}, on the causally processed held-out sessions of section~\ref{subsubsec:preprocessing}, and in its non-causal form in figure~\ref{fig:bi_result}. REALM-PT, the bidirectional models of figure~\ref{fig:bi_result} and Dual-VCT, whose front end decomposes each $5$\,s window as a whole and has no causal form, are offline by construction and are scored on the zero-phase chain. Supplementary Section~S2.1 lists the configuration of every baseline.

\section{Results}\label{sec:result}

\subsection{Retrospective knowledge distillation enhances the decodability of causal representations under linear-probe evaluation}\label{subsec:REALM}

\begin{figure}[t]
  \centering
  \includegraphics[width=\textwidth]{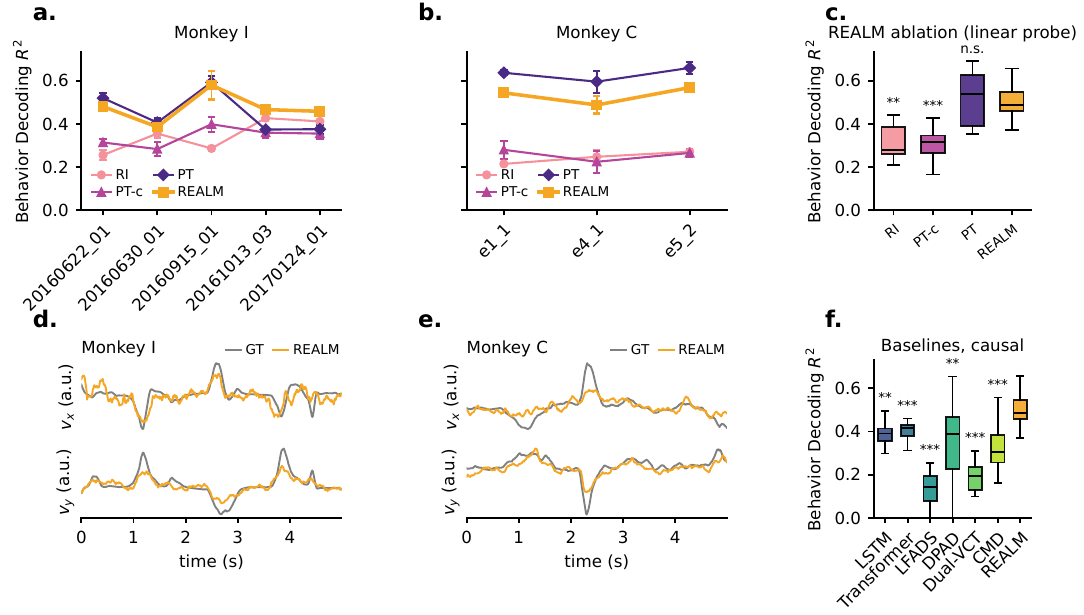}
    \caption{\textbf{Linear-probe evaluation of causal decoders.} All models are evaluated by the linear probe under the protocol of section~\ref{subsec:protocol}; no model uses labels before the read-out except DPAD, which is fitted with the velocity of the non-held-out sessions (section~\ref{subsec:comparisons}). Every causal model is scored on the causally processed held-out sessions (section~\ref{subsubsec:preprocessing}); PT and Dual-VCT, which have no causal form, are scored on the zero-phase chain. The metric is per-axis $R^2$ averaged over $v_x$ and $v_y$.
\textbf{(a, b)} Per-session $R^2$ on the held-out Makin (a, $N=5$) and Flint (b, $N=3$) sessions for the four models of the ablation chain defined in (c): RI, PT-c, PT and REALM. Markers are the mean over three seeds; error bars are the s.d.\ over the three seeds.
\textbf{(c)} The ablation chain, four configurations of the same architecture: RI, the architecture at initialization; PT-c, causal masked pretraining without retrospective knowledge distillation (RKD); PT, the bidirectional teacher from which REALM is distilled; REALM, that teacher transferred into a strictly causal student.
\textbf{(d, e)} Decoded velocity ($v_x$, top; $v_y$, bottom) from REALM (orange) against ground truth (GT, gray), for one Makin (d) and one Flint (e) session.
\textbf{(f)} The external comparison: two other temporal backbones, a long short-term memory network (LSTM) and a Transformer, pretrained and read out exactly as REALM is but without RKD, and four published decoders run under this protocol (latent factor analysis via dynamical systems (LFADS)~\cite{R2_pandarinath2018inferring}, dissociative prioritized analysis of dynamics (DPAD)~\cite{R34_sani2024dissociative}, Dual-VCT~\cite{R111_li2026dualvct} and CrossModalDistill (CMD)~\cite{R24_erturk2025cross}, all in their causal form). Every model shown, DPAD included, is scored on the full test fold; Supplementary Table~S17 reports how DPAD's scores change with the ten artifact segments dropped.
\textit{Statistics.} Boxes in (c, f) span the interquartile range (IQR) over the eight sessions $\times$ three seeds, the line inside each box marks the median, and the whiskers reach the most extreme value within $1.5\times$ IQR, with values beyond them not drawn; the stars above the boxes mark a two-sided paired $t$-test against REALM on the eight session means ($n = 8$, three seeds averaged within a session): $^{*}p < 0.05$, $^{**}p < 0.01$, $^{***}p < 0.001$. Margins quoted in the text use the same test; see section~\ref{subsubsec:stats}.}
\label{fig:causal_result}
\end{figure}

We first test whether the representation of a causal encoder trained without any labels decodes movement well (figure~\ref{fig:causal_result}), using the linear-probe evaluation of section~\ref{subsec:protocol}: pretraining and RKD see only LFP, and velocity enters only once, through the ridge read-out.

To test whether the boosted decodability comes from RKD, we ablate REALM (figure~\ref{fig:causal_result}(a), (b) and (c); per-session values in Supplementary Table~S1); the models are introduced in section~\ref{subsec:comparisons}. REALM reaches $R^2 = 0.496$, and REALM-PT-c, the same backbone and corpus without RKD, reaches $0.309$, so distillation adds $\Delta R^2 = +0.187$ ($95\%$ CI $[+0.119, +0.254]$, $d_z = 2.30$, $8/8$ sessions, $p < 0.001$). The architecture alone does not explain the gain: REALM-RI, the same network at initialization, reaches $0.308$ ($+0.188$, $[+0.081, +0.295]$, $d_z = 1.47$, $8/8$, $p = 0.004$). REALM-PT, the teacher, is the fully offline reference: it is read bidirectionally, with access to the entire segment, and on the zero-phase chain. It reaches $0.519$, and the causally deployed student is not significantly different from it ($-0.024$, $[-0.088, +0.041]$, $d_z = -0.31$, $2/8$, $p = 0.416$).

The six external models of figure~\ref{fig:causal_result}(f) are evaluated on the same held-out sessions and splits, with the same read-out and protocol as REALM (per-session values in Supplementary Table~S2). With CMAE but no RKD, an LSTM backbone reaches $0.385$ and a Transformer $0.399$; of the published decoders, each re-run in its causal form under this protocol, LFADS~\cite{R2_pandarinath2018inferring} reaches $0.111$, DPAD~\cite{R34_sani2024dissociative} $0.357$, Dual-VCT~\cite{R111_li2026dualvct} $0.170$ (on the zero-phase chain, see section~\ref{subsubsec:preprocessing}) and CrossModalDistill~\cite{R24_erturk2025cross} $0.328$.

REALM is above all six on average, and significantly so against each: by $+0.111$ ($[+0.060, +0.162]$, $d_z = 1.82$, $8/8$, $p = 0.001$) over the LSTM, $+0.096$ ($[+0.058, +0.134]$, $d_z = 2.13$, $8/8$, $p < 0.001$) over the Transformer, $+0.385$ ($[+0.256, +0.514]$, $d_z = 2.50$, $8/8$, $p < 0.001$) over LFADS, $+0.139$ ($[+0.054, +0.224]$, $d_z = 1.37$, $7/8$, $p = 0.006$) over DPAD, $+0.326$ ($[+0.241, +0.411]$, $d_z = 3.22$, $8/8$, $p < 0.001$) over Dual-VCT and $+0.168$ ($[+0.098, +0.239]$, $d_z = 1.99$, $8/8$, $p < 0.001$) over CrossModalDistill.

The traces in figure~\ref{fig:causal_result}(d), (e) show what this accuracy looks like. For both subjects, REALM follows the true velocity (gray) along both axes, capturing rapid reversals such as the $v_y$ peaks in the Makin session (d) and holding still through the pause between movements in the Flint session (e), without reading any future sample. That a decoder that is causal from preprocessing to read-out tracks these transients is the practical point: a decoder that jitters between movements injects noise into cursor control even when its $R^2$ looks acceptable.

\subsection{Retrospective knowledge distillation successfully aligns representations}
\label{subsec:alignment}

\begin{figure}[t]
  \centering
  \includegraphics[width=\textwidth]{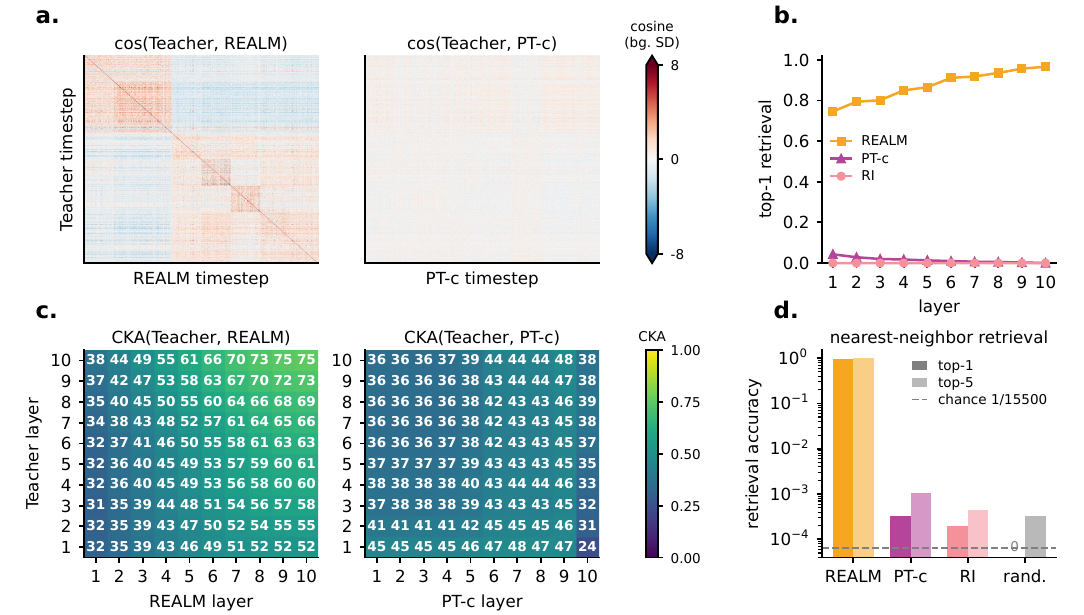}
\caption{ \textbf{RKD places the causal student in the teacher's frame.} All panels show the per-timestep representation after each layer of the frozen encoders of section~\ref{subsec:REALM}; at the last layer it is the representation the distillation loss of equation~(\ref{eq:distill}) aligns. Each held-out session is read by the REALM student distilled on its own corpus, on the zero-phase chain on which the distillation was trained, since this figure examines the distillation rather than the deployed decoder. The timesteps are sampled from whole sessions, not only from their test folds: each of the eight held-out sessions is cut into non-overlapping $5$\,s segments, $100$ segments are drawn at random per session (all $75$ of indy\_20160915\_01, which has no more), and every $25$th timestep of each segment is kept, giving $15{,}500$ timesteps. Panel (a) uses $800$ of these timesteps, $100$ timesteps per session drawn at random, because a matrix over all of them is not legible, and panels (b)--(d) use all $15{,}500$. \textbf{(a)} Every cell is the cosine similarity between the teacher's (REALM-PT's) representation of one timestep and a model's representation of another. The diagonal holds the matched pairs and everything off it is a mismatch. Left, the teacher against REALM; right, the teacher against REALM-PT-c, the same causal architecture after causal masked pretraining without RKD. Values are in units of the standard deviation of the between-session background. \textbf{(b)} Top-1 retrieval against encoder depth, for REALM, REALM-PT-c (PT-c in the legend) and REALM-RI (RI), that architecture at random initialization. \textbf{(c)} Layer-wise centered kernel alignment (CKA)~\cite{R109_kornblith2019cka} between the 10-layer teacher and, left, REALM and, right, REALM-PT-c. CKA is computed after normalizing each timestep's representation to zero mean and unit variance across its $256$ features, without a learned gain or bias. The number in each cell is the CKA multiplied by $100$. \textbf{(d)} Nearest-neighbor retrieval accuracy, top-1 and top-5, for REALM, REALM-PT-c, REALM-RI and random tensors (rand.), against the $1/15{,}500$ chance level; a top-1 accuracy of zero, which a log axis cannot show, is marked 0.}
    \label{fig:alignment}
\end{figure}

We then ask whether, through RKD, a causal student learns from its bidirectional teacher, REALM-PT. We compare REALM with REALM-RI and REALM-PT-c (Supplementary Table~S11).

We first compare the teacher and REALM in the left matrix of figure~\ref{fig:alignment}(a), whose diagonal holds the matched pairs: the two models' representations of the same held-out timestep. For comparison, the right matrix pairs the teacher with REALM-PT-c, and its diagonal does not show a similar structure. Cells that pair timesteps from different sessions carry no timestep-level correspondence, so we use them as the background: pooled over both matrices, their mean is $-0.007$ and their standard deviation $0.103$. This background is not uniform. Each corpus is read by its own student, so in REALM's matrix the cells that pair two sessions of the same corpus sit slightly above those that pair the two corpora ($+0.052$ against $-0.072$), which appears as a two-by-two block pattern in panel (a). On that scale, REALM's matched pair averages $+0.661$, or $6.5$ standard deviations above background, against $+0.097$ within a session and $-0.014$ between sessions. REALM-PT-c has no diagonal at all: its matched pair averages $+0.022$, or $0.3$ standard deviations, and nothing in its matrix stands out from the background.

We next ask how the three models compare with depth. Figure~\ref{fig:alignment}(b) profiles top-1 retrieval at matched depth; for each held-out timestep, top-1 retrieval asks whether a model's representation of that timestep is the nearest neighbor of the teacher's among all $15{,}500$ candidates. REALM is above both controls at every layer, and the separation widens with depth: at the first layer it retrieves at $0.746$ against $0.043$ for REALM-PT-c and below $0.001$ for REALM-RI, and by the tenth it is at $0.966$ while REALM-PT-c has fallen to $0.000$. As depth increases, the causal student's representation moves closer to that of a bidirectional teacher it can never imitate directly, since at every layer it still sees only the past.

We also compute centered kernel alignment (CKA)~\cite{R109_kornblith2019cka}, which measures representational similarity up to orthogonal transformation and isotropic scaling, so it asks whether two representations have the same shape without asking whether they share a frame. Because our objective aligns cosine similarities only at the final encoder layer and never optimizes CKA, CKA is a fair test of the alignment. REALM's CKA rises monotonically all the same, from $0.322$ at the first layer to $0.754$ at the tenth, while REALM-PT-c's stays between $0.375$ and $0.469$ and is $0.376$ at the tenth.

Finally, we ask the finer question directly: if each row of the similarity matrix over all $15{,}500$ timesteps is read as a ranking, does the matching timestep rank first (figure~\ref{fig:alignment}(d))? REALM puts it first $96.6\%$ of the time and within the top five $98.5\%$ of the time, a median rank of $1$ in $15{,}500$. REALM-PT-c reaches $0.03\%$ at a median rank of $5585$, REALM-RI $0.02\%$ at $7376$ and random tensors $0.00\%$ at $7746.5$, against a chance median of $7750$ (Supplementary Table~S3).

Together, the four analyses support the idea that the teacher--REALM alignment is not due to architectural similarity or a shared distribution of inputs but is genuinely the effect of the distillation process. The alignment appears in each of them: in the matched pairs of the similarity matrix, in retrieval of the teacher at every encoder depth, in layer-wise CKA and in the rank of the matching timestep. Thus, the retrospective distillation framework not only transfers the predictive capacity of the teacher to the student but also reproduces its geometric organization in the latent space.

\begin{figure}[t]
  \centering
  \includegraphics[width=\textwidth]{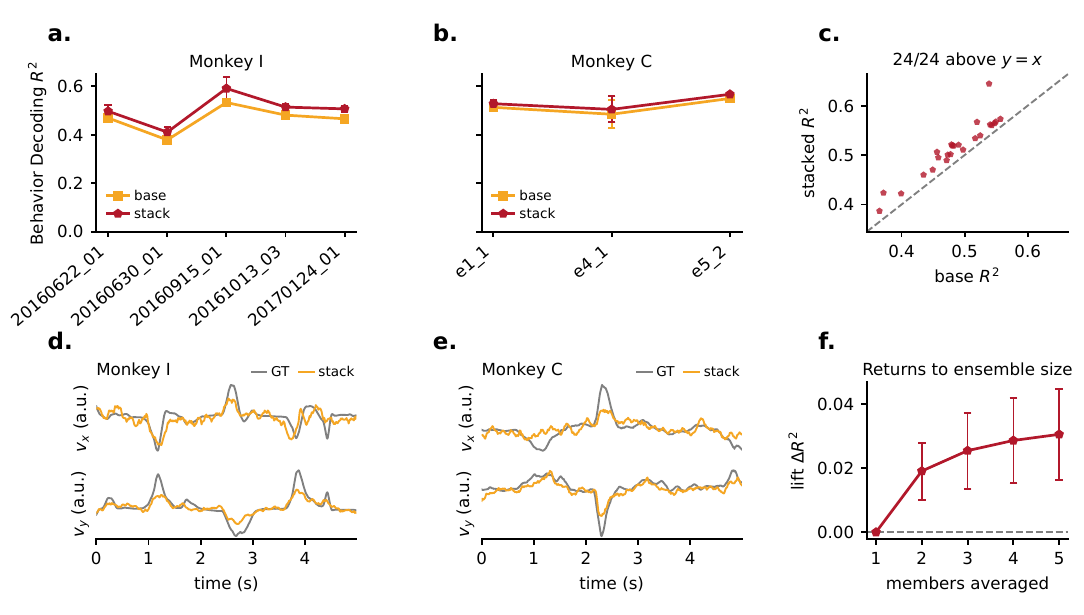}
    \caption{
    \textbf{Five-fold stacking provides a consistent improvement over single REALM base models.}
    All models are scored on the causally processed held-out sessions (section~\ref{subsubsec:preprocessing}).
    \textbf{(a, b)} Per-session decoding $R^2$ of the base models (base, the mean over the five) and of the 5-fold stacked ensemble (stack) on the held-out Makin (a) and Flint (b) sessions, for the $4.9$\,M causal REALM student.
    \textbf{(c)} Scatter of stacked $R^2$ vs.\ base-model $R^2$ across all (held-out session $\times$ evaluation split) combinations ($n = 24$); the dashed line denotes $y=x$.
    \textbf{(d, e)} Example decoded velocity traces ($v_x$, top; $v_y$, bottom) comparing GT (gray) with the 5-fold ensemble prediction (orange), shown for one Makin (d) and one Flint (e) session. A single base model is not drawn: its trace is not separable from the ensemble's at this scale.
    \textbf{(f)} Lift $\Delta R^2 = R^2_\text{stacked} - R^2_\text{base}$ as a function of ensemble size $k$, aggregated across the eight held-out sessions and three evaluation splits and averaged over all $\binom{5}{k}$ subsets of that size.
     \textit{Statistics.} Two-sided paired $t$-tests on the eight held-out session means ($n = 8$); markers in (a, b) are the mean and error bars the s.d.\ over the three seeds, and markers in (f) the mean and error bars the s.d.\ over the eight sessions. Full specification in section~\ref{subsubsec:stats}.}
    \label{fig:stacking}
\end{figure}

\subsection{Stacking 5-fold REALM models further improves decoding accuracy}
\label{sec:stacking}

In addition to single-model evaluation, we examine whether REALM benefits from a simple ensembling strategy on top of its existing pretraining-plus-distillation pipeline. Specifically, we partition the non-held-out training sessions into five disjoint folds and train five base models, each distilled on four of the five folds, with a different fold withheld each time. At inference time, we average the per-fold predictions to obtain a final stacked ensemble prediction. Each base model may capture different LFP features, and averaging their outputs can prevent prediction oscillations and boost accuracy. We refer to the per-fold mean of the five base models as ``base $R^2$'' and the ensemble mean as ``stacked $R^2$''; the lift $\Delta R^2 = R^2_\text{stacked} - R^2_\text{base}$ quantifies the gain attributable to ensembling alone.

Stacking produces a positive lift on every held-out session for both monkeys (figure~\ref{fig:stacking}(a), (b)). The scatter in figure~\ref{fig:stacking}(c) places all $24$ (session $\times$ evaluation split) points above the identity line, so the gain does not come from a few sessions offsetting the rest. On average over the eight held-out sessions and three evaluation splits, the five-member ensemble raises $R^2$ from $0.484$ to $0.515$ (Supplementary Table~S4), a lift of $\Delta R^2 = +0.031$ ($+6.3\%$ relative; $95\%$ CI $[+0.019, +0.042]$, $d_z = 2.15$, $8/8$ sessions, $p < 0.001$). The gain is larger on Makin ($0.465 \rightarrow 0.504$, $+0.038$) than on Flint ($0.516 \rightarrow 0.533$, $+0.017$), where the base models already sit higher. Figure~\ref{fig:stacking}(f) resolves the lift against ensemble size: it grows monotonically but with clearly diminishing returns, from $+0.019 \pm 0.009$ at two members to $+0.025 \pm 0.012$ at three, $+0.029 \pm 0.013$ at four and $+0.031 \pm 0.014$ at five (mean $\pm$ s.d.\ over the eight held-out sessions), so most of the available benefit is already captured by the first two or three folds. Figure~\ref{fig:stacking}(d), (e) show the ensemble's output against the ground truth on one Makin and one Flint session. It follows the slow envelope of velocity and the rapid reversals alike.

\begin{figure}[t]
  \centering
  \includegraphics[width=\textwidth]{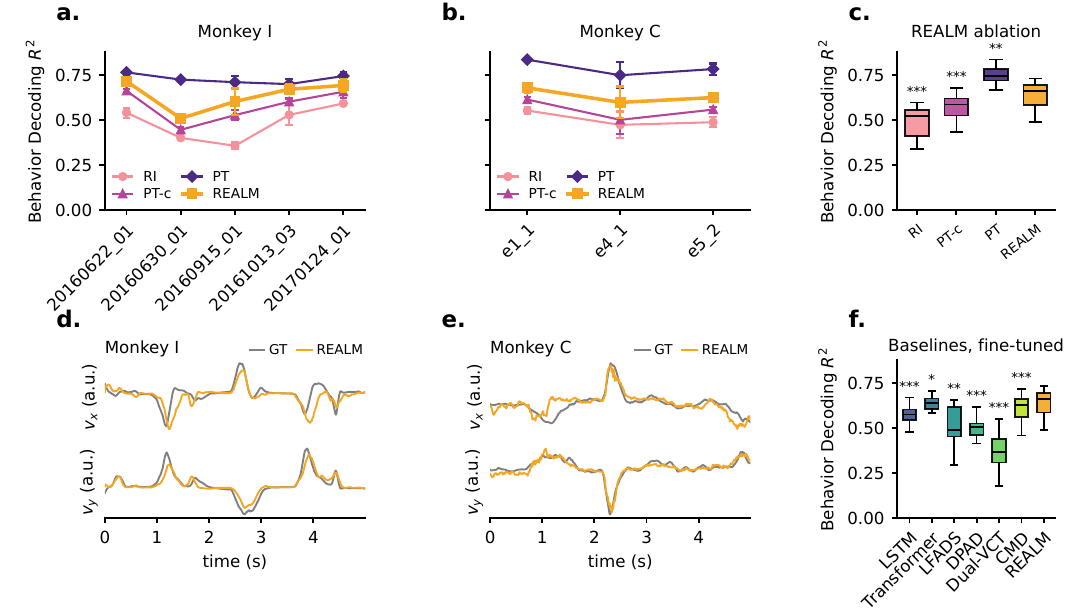}
    \caption{\textbf{Fine-tuned (supervised) causal decoding.} Every model is adapted, with the encoder unfrozen, on the $72\%$ training fold of the session it is scored on, early-stopped on that session's $8\%$ validation fold and scored once on the $20\%$ test fold; the distillation itself stays label-free. \textbf{(a, b)} Per-session $R^2$ on Makin (a, $N=5$) and Flint (b, $N=3$) for the models of (c). \textbf{(c)} The ablation chain of figure~\ref{fig:causal_result}(c), fine-tuned. \textbf{(d, e)} Decoded velocity ($v_x$, top; $v_y$, bottom) from REALM (orange) against GT (gray) for one Makin (d) and one Flint (e) session. \textbf{(f)} The baselines of figure~\ref{fig:causal_result}(f), each adapted on the held-out session under this protocol (Supplementary Table~S16), all in their causal form. DPAD alone is scored, and early-stopped, with the ten artifact segments of Supplementary Table~S17 dropped, because its per-session fit is pulled down by their transients (dropping them raises its $R^2$ by $+0.029$); every other model is scored on the full test fold. Every causal model is scored on the causally processed held-out sessions, PT and Dual-VCT on the zero-phase chain (section~\ref{subsubsec:preprocessing}). \textit{Statistics.} Per-axis $R^2$ averaged over $v_x$ and $v_y$, on the eight held-out sessions and three seeds; markers and error bars in (a, b) are the mean and s.d.\ over the three seeds, and boxes in (c, f) are drawn as in figure~\ref{fig:causal_result}; stars mark two-sided paired $t$-tests against REALM on the eight session means ($n = 8$; $^{*}p < 0.05$, $^{**}p < 0.01$, $^{***}p < 0.001$). Margins quoted in the text use the same test; see section~\ref{subsubsec:stats}.}
    \label{fig:finetuned}
\end{figure}

\subsection{Retrospective knowledge distillation also enhances the decodability of the causal student in the fine-tuned (supervised) pipeline}\label{subsec:finetuned}

The linear probe of section~\ref{subsec:REALM} measures the representation itself, since labels are used only to fit a linear read-out on the frozen encoder, but in some scenarios enough labeled data from the evaluation session are available to adapt the encoder as well. Whether the RKD advantage survives such adaptation does not follow from the linear-probe result. We therefore repeat the comparison with the encoder unfrozen except for the neural tokenizer, fine-tuning each model on the same held-out session with the same $72/8/20$ split ($72\%$ for fine-tuning, $8\%$ for early stopping and model selection, and $20\%$ for scoring). The two regimes differ in two respects: whether gradients reach the encoder after distillation, and which read-out is used. The fine-tuned read-out is a linear layer on the encoder output at each step, trained on the $72\%$ training fold (section~\ref{subsec:protocol}).

In the fine-tuned pipeline (per-session values in Supplementary Table~S5), REALM reaches $R^2 = 0.636$ against $0.571$ for the identical backbone without RKD (PT-c), a gain of $+0.066$ ($95\%$ CI $[+0.051, +0.080]$, $d_z = 3.71$, $8/8$ sessions, $p < 0.001$), and $+0.145$ over the same architecture at initialization (RI; $[+0.106, +0.184]$, $d_z = 3.10$, $8/8$, $p < 0.001$). The margin is smaller than the $+0.187$ shown under linear-probe evaluation, as expected when supervision lets the model repair a weaker representation, but fine-tuning cannot replace RKD. Compared with the bidirectional teacher (PT), REALM is lower once both are fine-tuned, $0.636$ against $0.751$ ($-0.115$, $[-0.170, -0.059]$, $d_z = -1.73$, $0/8$ sessions, $p = 0.002$). This is expected, as the teacher, which reads the future of each segment and the zero-phase chain, sets the offline ceiling.

Against the other baselines, all in their causal form, REALM is above all six on average (figure~\ref{fig:finetuned}(f); Supplementary Table~S6), and significantly so against each: $+0.080$ ($[+0.051, +0.109]$, $d_z = 2.29$, $8/8$, $p < 0.001$) over an LSTM, $+0.022$ ($[+0.001, +0.043]$, $d_z = 0.89$, $7/8$, $p = 0.040$) over a Transformer backbone, $+0.133$ ($[+0.063, +0.203]$, $d_z = 1.60$, $8/8$, $p = 0.003$) over LFADS~\cite{R2_pandarinath2018inferring}, $+0.143$ over DPAD~\cite{R34_sani2024dissociative} with the artifact segments dropped ($[+0.091, +0.194]$, $d_z = 2.31$, $8/8$, $p < 0.001$), $+0.270$ ($[+0.191, +0.348]$, $d_z = 2.88$, $8/8$, $p < 0.001$) over Dual-VCT~\cite{R111_li2026dualvct} and $+0.035$ ($[+0.021, +0.050]$, $d_z = 2.07$, $8/8$, $p < 0.001$) over CrossModalDistill~\cite{R24_erturk2025cross}. Fine-tuning on each session's labeled data closes most of the gap that separates REALM from CrossModalDistill under linear-probe evaluation, from $+0.168$ to $+0.035$; CrossModalDistill's student is distilled from a spike teacher, whereas REALM uses no spikes at any stage. 

\subsection{Label efficiency of the fine-tuned pipeline}\label{subsec:fewshot}

Section~\ref{subsec:finetuned} fine-tunes on the whole training fold of the held-out session, which is more labeled data than a session would realistically be calibrated with. Shrinking that budget, with the validation and test folds untouched so every point is scored on identical segments, turns the single number into a curve and addresses the question a deployment actually faces: how much supervision the decoder needs before it is usable. 

\begin{figure}[t]
  \centering
  \includegraphics[width=\textwidth]{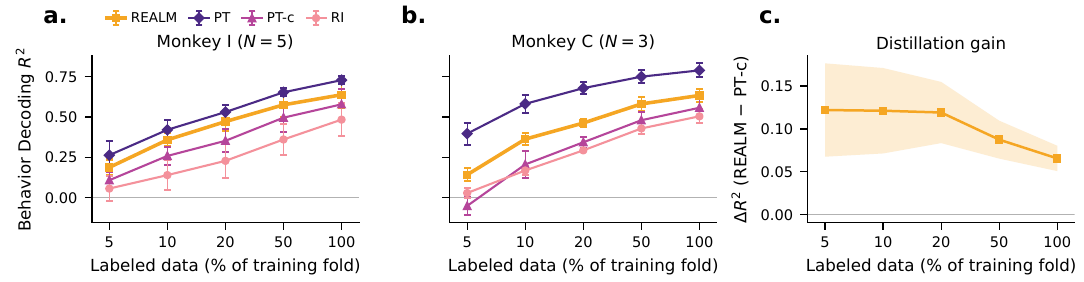}
    \caption{\textbf{Label efficiency of the fine-tuned decoder.} The fine-tuning of figure~\ref{fig:finetuned} repeated with the labeled budget set to $5\%$, $10\%$, $20\%$, $50\%$ and $100\%$ of the held-out session's training fold, taken as its earliest segments in time; the validation and test folds are untouched, so every point is scored on the same segments. Models, colors and markers are as in figure~\ref{fig:finetuned}, and so is the preprocessing: PT on the zero-phase chain, every causal model on the causally processed held-out sessions. \textbf{(a, b)} Per-axis $R^2$ on Makin (a, $N=5$) and Flint (b, $N=3$). \textbf{(c)} The distillation gain, REALM against PT-c. \textit{Statistics.} Points in (a, b) are the mean over the sessions of that dataset, with the three seeds averaged within a session first, and error bars the s.d.\ over those session means; the band in (c) is the $95\%$ confidence interval of the paired difference over the eight session means ($n = 8$, two-sided paired $t$-test). Full specification in section~\ref{subsubsec:stats}.}
    \label{fig:fewshot}
\end{figure}

The budget is a fraction of that fold rather than a fixed number of minutes, because the eight held-out sessions differ in duration. Every session can supply any fraction of its own fold, and the $100\%$ point is the fine-tuned pipeline of section~\ref{subsec:finetuned}, which uses the whole $72\%$ training fold. How the labeled segments are drawn matters as much as how many of them there are. Whole $5$\,s segments are assigned to folds, as in section~\ref{subsec:protocol}, so the folds do not overlap, and the budget consists of the earliest segments of the training fold in time order rather than a random draw from the whole fold. Because the folds are drawn at random over the session (Supplementary Section~S3.1), these segments do not form one uninterrupted block: a budget of a given fraction of the fold spans roughly the same fraction of the session from its start, with the validation and test segments that fall in that stretch interleaved among them. The $8\%$ validation fold is used whole for early stopping at every budget, so the data used for model selection do not shrink with the labeled budget.

The distillation gain grows as supervision shrinks (figure~\ref{fig:fewshot}(c)), and it does so monotonically. Against the same backbone without RKD, REALM is ahead by $+0.066$ on the whole training fold ($[+0.051, +0.080]$, $d_z = 3.71$, $8/8$, $p < 0.001$), $+0.087$ at half of it ($[+0.065, +0.110]$, $d_z = 3.31$, $8/8$, $p < 0.001$), $+0.119$ at a fifth ($[+0.083, +0.155]$, $d_z = 2.79$, $8/8$, $p < 0.001$), $+0.121$ at a tenth ($[+0.071, +0.171]$, $d_z = 2.03$, $8/8$, $p < 0.001$) and $+0.122$ at a twentieth ($[+0.068, +0.176]$, $d_z = 1.87$, $8/8$, $p = 0.001$). The margin at the smallest budget is almost twice the margin at the largest, so distillation matters most in the regime a deployed decoder occupies. Its effect is to remove the need for data rather than to add a fixed increment on top of it: on half the training fold REALM reaches $0.578$, which is where the same backbone without RKD arrives only with all of it ($0.571$, a difference of $+0.007$ with the interval covering zero, $[-0.016, +0.030]$, $d_z = 0.25$, $5/8$, $p = 0.500$).

Against the teacher, the budget changes little. The bidirectional teacher stays ahead at every budget and by a similar margin: $-0.115$ on the whole training fold, as section~\ref{subsec:finetuned} reports ($[-0.170, -0.059]$, $d_z = -1.73$, $0/8$, $p = 0.002$), $-0.112$ at half of it ($[-0.178, -0.046]$, $d_z = -1.42$, $0/8$, $p = 0.005$), $-0.118$ at a fifth ($[-0.210, -0.026]$, $d_z = -1.07$, $1/8$, $p = 0.019$), $-0.121$ at a tenth ($[-0.210, -0.032]$, $d_z = -1.13$, $1/8$, $p = 0.015$) and $-0.143$ at a twentieth ($[-0.233, -0.053]$, $d_z = -1.33$, $0/8$, $p = 0.007$). The teacher's offline access, to the future of each segment and to the zero-phase chain, is worth a roughly constant $0.11$ to $0.14$ whatever the supervision; what distillation removes is the need for labels, not this price of causal operation.

\subsection{Model size and edge deployment}\label{result-model inference}

Parameter capacity affects both the accuracy and the inference time of a deep-learning decoder, so we introduce six REALM versions, three causal and three bidirectional: REALM-S ($2.11$\,M, $8$ layers), REALM ($4.91$\,M, $10$ layers) and REALM-L ($10.50$\,M, $22$ layers); and REALM-bi-S ($2.23$\,M, $3$ layers), REALM-bi ($5.54$\,M, $8$ layers) and REALM-bi-L ($11.54$\,M, $10$ layers). All six share the teacher's width, $d_{\text{model}} = 256$, because the representation alignment compares student and teacher states elementwise; capacity is set by depth and by the state-space expansion factor instead. A bidirectional layer runs the recurrence in both directions and costs about twice as much as a causal one, which is why the bidirectional family reaches a comparable parameter count with fewer layers ($3$, $8$ and $10$ against $8$, $10$ and $22$).

\begin{figure}[t]
  \centering
  \includegraphics[width=\textwidth]{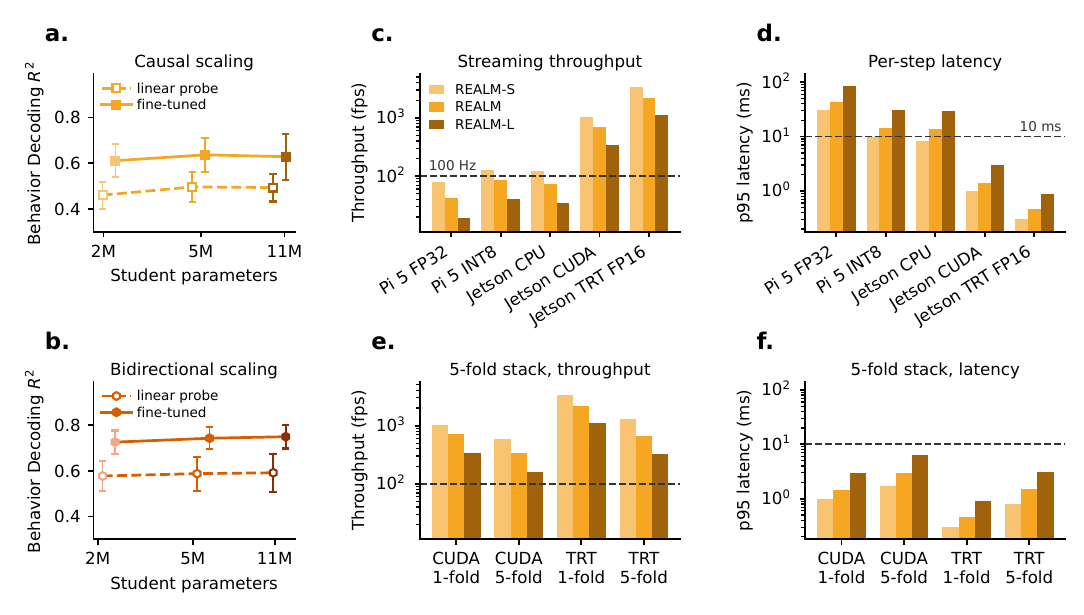}
    \caption{\textbf{Accuracy against model size, and the cost of streaming on edge hardware.} \textbf{(a)} $R^2$ against student capacity for the three causal sizes (REALM-S, REALM and REALM-L, light to dark), under linear-probe evaluation (dashed line, open markers) and after per-session fine-tuning (solid line, filled markers). \textbf{(b)} The same for the three bidirectional sizes (REALM-bi-S, REALM-bi and REALM-bi-L); the values of (a, b) are in Supplementary Table~S7; the causal sizes are scored on the causally processed held-out sessions, the bidirectional ones on the zero-phase chain. \textbf{(c)} Throughput in frames per second (fps) of the three causal variants on a Raspberry Pi 5 and a Jetson Orin Nano (FP32 and FP16, 32- and 16-bit floating point; INT8, 8-bit integer quantization; CUDA, the model replayed as CUDA Graphs; TRT, a TensorRT engine; Supplementary Table~S8); the dashed line marks the $100$\,Hz streaming requirement. \textbf{(d)} The 95th-percentile (p95) latency per step for the same configurations, with the dashed line at the $10$\,ms budget. \textbf{(e, f)} Throughput and latency on the Jetson of a single model (1-fold) and of the five-member ensemble (5-fold), on the same axis ranges as (c, d). \textit{Statistics.} Markers in (a, b) are the mean and error bars the s.d.\ over the eight held-out sessions; throughput is the reciprocal of the median step time and latency its 95th percentile, over at least $500$ timed single-step invocations after warm-up, with no inferential test. The Jetson was timed with its clocks locked at their maximum, the Raspberry Pi 5 in an earlier session with the CPU frequency governor unrecorded. Full specification in section~\ref{subsubsec:stats}.}
    \label{fig:realtime}
\end{figure}

Figure~\ref{fig:realtime}(a), (b) plot the linear-probe $R^2$ against parameter count for the three causal sizes and the three bidirectional ones, and both families show the same pattern. Within a family, capacity adds little: the causal curve rises from $0.460$ at REALM-S to $0.496$ at REALM and then levels off, $0.493$ at REALM-L, a last step that is not separable ($-0.003$, $[-0.018, +0.013]$, $d_z = -0.14$, $6/8$ sessions, $p = 0.704$) for more than twice the parameters; the bidirectional curve runs $0.577$ to $0.588$ to $0.592$ and its last step is not separable ($+0.004$, $[-0.010, +0.018]$, $d_z = 0.23$, $5/8$, $p = 0.532$). Between the families, the gap does not move: lifting the causal constraint, on the model and on the preprocessing, adds $+0.118$ at the small size ($[+0.069, +0.166]$, $d_z = 2.02$, $8/8$, $p < 0.001$), $+0.092$ at the middle one ($[+0.057, +0.127]$, $d_z = 2.17$, $8/8$, $p < 0.001$) and $+0.098$ at the large one ($[+0.053, +0.144]$, $d_z = 1.82$, $8/8$, $p = 0.001$). That is the useful comparison: the price of causal operation is roughly constant across a fivefold range of capacity, and it is larger than anything scaling recovers within either family, so a causal decoder cannot be made to read like a bidirectional one by making it bigger. When the encoder is fine-tuned instead of frozen, the curve is similarly flat: REALM-L and REALM are not separated ($-0.008$, $[-0.037, +0.020]$, $d_z = -0.24$, $3/8$, $p = 0.521$). Since the sizes decode nearly alike, the choice between them is a question about hardware, which the rest of this section addresses.

We then evaluate REALM on two edge-AI devices. The first is the Raspberry Pi 5 (4\,GB RAM; 4 cores $\times$ Cortex-A76; CPU frequency: 2.4\,GHz; operating system: 64-bit Linux, kernel 6.12.47), which is representative of an affordable wearable or battery-operated companion device. The second is the NVIDIA Jetson Orin Nano (Super developer kit; 6 cores $\times$ Cortex-A78AE; frequency: 1.728\,GHz; GPU: Ampere class, 1.02\,GHz; memory: 8\,GB LPDDR5). All Jetson timings are taken in its uncapped MAXN SUPER power mode with the CPU, GPU and memory clocks locked at their maximum (CPU $1.728$\,GHz, GPU $1.02$\,GHz). Only the three causal variants are benchmarked; the bidirectional models cannot be deployed for streaming decoding. Each Raspberry Pi and Jetson configuration is warmed up with 50 invocations and then timed over 500 invocations of the stateful single-step API.  We report the mean, median (p50) and 95th-percentile (p95) per-step latency, together with throughput in frames per second (fps). Because the decoder must keep up with a 100\,Hz iBCI stream, the per-step latency budget is 10\,ms. Summary data are shown in figure~\ref{fig:realtime}, and the detailed data for each configuration are listed in Supplementary Table~S8.

On the Raspberry Pi 5, we characterize a baseline single-thread configuration in 32-bit floating point (FP32) and then apply two cumulative optimizations: dynamic 8-bit integer (INT8) quantization of all linear layers and pinning the process to three of the four cores. REALM (4.9\,M) reaches 42\,fps in FP32 and 85\,fps in INT8, which is far above the 30\,Hz operating rate commonly used in closed-loop iBCI systems, whereas REALM-S attains 79\,fps (FP32) and 127\,fps (INT8); in INT8 it is the sole Pi 5 setup that achieves the full 100\,Hz online budget. Beyond the headline throughput, the optimization stack also dramatically tightens the latency distribution: the single-thread FP32 baseline produces a markedly right-skewed tail (REALM-S mean 15.30\,ms vs.\ p95 31.05\,ms, a p95-to-mean tail factor of $2.03\times$), whereas INT8 quantization with core pinning collapses it (REALM-S mean 8.33\,ms vs.\ p95 9.80\,ms, a tail factor of $1.18\times$). Closed-loop control quality is governed by worst-case step time rather than by the average, so this jitter reduction is as practically valuable as the throughput gain.

The Jetson Orin Nano is a more powerful platform with an integrated Ampere-class GPU. We benchmark three configurations on it (figure~\ref{fig:realtime}(c), (d)): PyTorch on the CPU in FP32, PyTorch with CUDA Graphs on the GPU, and a TensorRT engine in 16-bit floating point (FP16) exported from a stateless ONNX graph. Capturing the streaming step into a CUDA graph, so that the many small kernels of each step are replayed as a single launch, raises throughput $8.3$--$9.9\times$ over the Jetson CPU (1025\,fps for REALM-S, 703\,fps for REALM, 339\,fps for REALM-L), while exporting the same step to a TensorRT FP16 engine yields a further $3.1$--$3.3\times$ speedup (3342\,fps for REALM-S, 2183\,fps for REALM, 1121\,fps for REALM-L). End-to-end, the REALM-L p95 latency drops from 83.45\,ms on the Pi 5 to 0.90\,ms on the Jetson with TensorRT FP16; TensorRT FP16 places all three causal variants well below the $10$\,ms budget, and even the largest leaves an $11\times$ margin.

Two of these configurations change the arithmetic, not just the schedule, so the speedup is only meaningful if decoding accuracy survives it. We therefore measured accuracy at each precision on the same eight causally processed held-out sessions and three seeds, with the encoder frozen and the ridge penalty fixed at its FP32 selection, so that the number format is the only thing that varies. Half precision leaves accuracy nearly unchanged: $R^2$ moves by at most $0.0082$ across the three sizes ($0.4600$, $0.4958$ and $0.4931$ at FP32 against $0.4518$, $0.4957$ and $0.4924$ at FP16). Dynamic INT8 quantization of the linear layers costs more, up to $0.1120$ ($-0.0807$, $-0.1120$ and $-0.0781$), so its speedup comes at a real accuracy price. Neither format is guaranteed to be harmless: the same encoders in bfloat16, which keeps FP32's exponent range but carries three fewer mantissa bits than FP16, collapse to $0.19$--$0.25$. The state-space recurrence accumulates over $500$ steps, so mantissa precision is what it is sensitive to, and FP16 happens to have enough of it. 

The stacking ensemble of section~\ref{sec:stacking} runs its five base models inside a single CUDA graph (or TensorRT engine) and averages their predictions, which raises throughput by $1.4\times$ to $2.8\times$ over naive $5\times$ serial replay (figure~\ref{fig:realtime}(e), (f)). Even with ensembling, TensorRT FP16 streams all three models in real time on the Jetson, at $1278$, $665$ and $324$\,fps (figure~\ref{fig:realtime}(e)), that is, median step times of $0.78$, $1.50$ and $3.09$\,ms. None of the ensemble configurations exceeds the $10$\,ms latency budget: the slowest, REALM-L on PyTorch CUDA Graphs, has a p95 latency of $6.28$\,ms, and conversion to TensorRT FP16 brings it down to $3.09$\,ms. The $+0.017$ to $+0.038$ $R^{2}$ lift from stacking (Supplementary Table~S4) is therefore essentially free in deployment terms.

Together, these configurations span more than two orders of magnitude of throughput across the two devices, from $19$ to $3342$\,fps. REALM-S runs at $100$\,Hz on the Raspberry Pi~5 with INT8 quantization and core pinning, at a p95 latency of $9.80$\,ms. Alternatively, the TensorRT FP16 flow can run all causal models, including the five-member ensemble, in the Jetson's uncapped MAXN SUPER power mode with the clocks locked at their maximum. For low-latency real-time iBCI requirements, such as a $30$\,Hz decoding frequency, REALM can achieve hard real-time performance, with no misses. To our knowledge, this is the first demonstration of a purely LFP-based decoder operating causally end-to-end at $100$\,Hz or above on a portable, low-power device, supporting the broader claim that RKD can produce decoders that combine high accuracy with real-time clinical applicability.

\subsection{Extension: retrospective knowledge distillation also improves the bidirectional model}

We further extend the REALM pipeline to the bidirectional setting (REALM-bi) and evaluate it under the same protocol used for the causal student.

\begin{figure}[t]
  \centering
  \includegraphics[width=\textwidth]{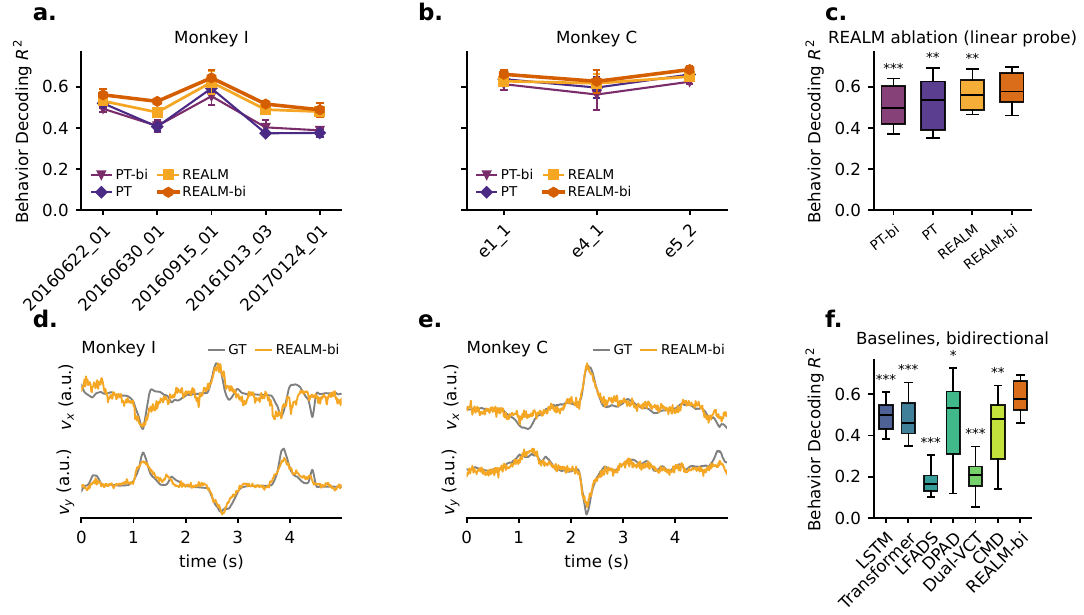}
    \caption{\textbf{The same recipe with a bidirectional student.} As in figure~\ref{fig:causal_result} for corpus, teacher, objective and linear-probe evaluation, with the student's causal constraint lifted; every model here, the causal student included, is scored on the zero-phase chain, as these are offline decoders. \textbf{(a, b)} Per-session $R^2$ on Makin (a, $N=5$) and Flint (b, $N=3$). \textbf{(c)} The bidirectional ablation: PT-bi, a bidirectional Mamba-2 matched to REALM-bi in size and given the teacher's masked pretraining but no distillation; PT, the teacher from which REALM-bi is distilled; REALM, the causal student, shown for the price of causality; REALM-bi, the distilled bidirectional student. \textbf{(d, e)} Decoded velocity ($v_x$, top; $v_y$, bottom) from REALM-bi against GT (gray) for one Makin (d) and one Flint (e) session. \textbf{(f)} The baselines of figure~\ref{fig:causal_result}(f) in their non-causal form, all scored on the full test fold; DPAD is read non-causally through the bidirectional layers of its reference implementation (Supplementary Table~S10). \textit{Statistics.} Per-axis $R^2$ on the eight held-out sessions and three seeds; markers and error bars in (a, b) are the mean and s.d.\ over the three seeds, and boxes in (c, f) are drawn as in figure~\ref{fig:causal_result}; stars mark two-sided paired $t$-tests against REALM-bi on the eight session means ($n = 8$; $^{*}p < 0.05$, $^{**}p < 0.01$, $^{***}p < 0.001$). Full specification in section~\ref{subsubsec:stats}.}
  \label{fig:bi_result}
\end{figure}

The ablation in figure~\ref{fig:bi_result}(c) shows that distillation helps the bidirectional student too, by a margin that access to future samples cannot explain. REALM-PT-bi, the same structure and parameter capacity pretrained without RKD, reaches $0.505$ (Supplementary Table~S9), so distillation adds $+0.083$ here ($[+0.060, +0.105]$, $d_z = 3.09$, $8/8$, $p < 0.001$) even though this student already sees the whole segment. The distilled student also overtakes the teacher it was distilled from, $0.588$ against $0.519$ ($+0.068$, $[+0.027, +0.109]$, $d_z = 1.40$, $8/8$, $p = 0.006$), which a pure transfer of the teacher's knowledge could not produce. The distillation term is therefore not merely a channel for the teacher's access to the future.

The per-session plots in figure~\ref{fig:bi_result}(a), (b) show that the ordering holds session by session in both subjects, and panel (f) places REALM-bi against the alternatives evaluated under the identical protocol and read bidirectionally (Supplementary Table~S10). It is ahead of all six on average, on every session against five of them and on seven of the eight against DPAD, and significantly so against each: $+0.095$ over an LSTM ($[+0.071, +0.118]$, $d_z = 3.40$, $8/8$, $p < 0.001$), $+0.110$ over a Transformer ($[+0.085, +0.135]$, $d_z = 3.71$, $8/8$, $p < 0.001$), $+0.413$ over LFADS~\cite{R2_pandarinath2018inferring} ($[+0.308, +0.518]$, $d_z = 3.29$, $8/8$, $p < 0.001$), $+0.124$ over DPAD~\cite{R34_sani2024dissociative} ($[+0.034, +0.215]$, $d_z = 1.15$, $7/8$, $p = 0.014$), $+0.380$ over Dual-VCT~\cite{R111_li2026dualvct} ($[+0.319, +0.441]$, $d_z = 5.17$, $8/8$, $p < 0.001$) and $+0.164$ over CrossModalDistill ($[+0.084, +0.245]$, $d_z = 1.70$, $8/8$, $p = 0.002$). 

Figure~\ref{fig:bi_result}(d), (e) show example traces. In both subjects and recording setups, REALM-bi follows the ground-truth velocity (gray) closely on both axes, including the fast directional reversals of the Makin session (d) and the slower, lower-amplitude modulations of the Flint session (e). The difference from the causal model in figure~\ref{fig:causal_result}(d), (e) is clearest on the rising edges of fast transients, where REALM-bi locks onto the ground truth sooner. This is where future context helps, and it is what the causal student must learn to reproduce without that context. REALM-bi is thus the offline upper bound of the REALM family: it keeps the pretraining and distillation gains while reading context in both directions, so it can be used directly as an offline decoder.

\section{Discussion}\label{sec:discussion}

REALM is an offline-to-online distillation framework for low-bandwidth, real-time decoding of local field potentials that addresses three challenges in translating iBCIs to clinical use: (i) High accuracy often relies on spiking activity and high-bandwidth recording, whose power cost and gradual unit loss limit what a wireless chronic implant can sustain. (ii) The most accurate neural decoders are non-causal and cannot be used directly for streaming closed-loop inference. (iii) Behavioral labels can also be difficult to obtain in clinical settings, and sometimes impossible to annotate by hand, limiting the calibration of supervised decoders. Taking its cue from offline-to-online architectures in speech recognition, REALM transfers the contextual structure of the neural signal from a non-causal teacher to a causal student: the RKD objective maximizes the cosine similarity between teacher and student representations at the encoder's final layer, so the student learns to extract and represent structure that the non-causal encoder recovers from future context. In continuous velocity decoding during non-human primate point-to-point reaching, the resulting causal student, run on held-out recordings that are also preprocessed causally, reaches an $R^2$ of $0.496$ under linear-probe evaluation and $0.636$ after supervised fine-tuning, compared with $0.309$ and $0.571$ for the same backbone without RKD. It uses LFPs alone as its neural signal and runs in real time on edge hardware at rates above $100$\,Hz. REALM has a higher mean accuracy than every causal baseline compared here and significantly outperforms each one, as shown by the ablations within the REALM family and the pairwise statistical comparisons. A gap to its bidirectional counterpart remains, but RKD makes this gap smaller than it is without distillation.

Existing iBCI decoders often face one or both of two limitations. Some rely on spiking activity, for which the number of recordable units may decline over chronic implantation, potentially reducing long-term decoding performance. The bandwidth required for spike recordings also poses a substantial challenge for wireless implants. Others are non-causal and require access to future samples at inference, making them unsuitable for real-time deployment. Meanwhile, existing causal models often achieve lower decoding accuracy. These limitations leave a need for an accurate causal decoder that operates on low-bandwidth signals with modest power requirements. REALM addresses this need: using low-rate neural signals, it distills a causal student from a non-causal teacher and achieves accurate decoding without access to future context at inference. On the same held-out sessions, splits, and evaluation protocol, REALM outperforms the strongest published baseline in each pipeline, CrossModalDistill in both. The differences are $+0.168$ under linear-probe evaluation ($[+0.098, +0.239]$, $d_z = 1.99$, $8/8$, $p < 0.001$ by a two-sided paired $t$-test) and $+0.035$ after fine-tuning ($[+0.021, +0.050]$, $d_z = 2.07$, $8/8$, $p < 0.001$), with the preprocessing of every held-out session run causally.

REALM is designed for chronic real-time iBCIs. Its primary applications include upper-limb prosthesis or neuroprosthesis control for people with severe motor impairment, as well as chronic LFP decoding for basic neuroscience. Success in these scenarios may translate into meaningful quality-of-life gains. On the NVIDIA Jetson Orin Nano, a low-power edge platform, REALM decodes in about $1.4$\,ms per inference ($703$\,fps at full precision), providing roughly a $7.0\times$ speed margin relative to the $100$\,Hz closed-loop control demands and a further speedup of about $3\times$ under TensorRT FP16. On the more constrained Raspberry Pi~5, REALM reaches $42$\,fps in the baseline FP32 configuration; the optimized configurations achieve higher throughput. Decoding and preprocessing are causal, so the pipeline runs on the signal as it arrives. Pretraining and distillation use neither kinematic labels nor a second neural signal modality. For each held-out session, only the linear read-out is fitted, in closed form, using that session's velocity labels, while the encoder is not. Using a single low-rate neural signal reduces the amount of data that a wireless implant must transmit and may lower its power requirement. For the patient, calibration therefore reduces to fitting the read-out rather than retraining the network. In the fine-tuned pipeline, the few-shot results (section~\ref{subsec:fewshot}) show how much labeled data are saved by distillation: with half of the training fold, REALM reaches the accuracy that the same backbone without RKD attains only with all of it. Distillation removes the need for data rather than adding a fixed increment on top of it, which matters most where session time is short.

Several limitations must be taken into account when interpreting these findings. (i) \emph{Subject and task generalization.} The experiments use rhesus macaque data from two datasets, and the task is confined to decoding two-dimensional cursor (Makin) and hand (Flint) velocity. Extending the method to other task types, such as discrete movements or three-dimensional reaching, and more importantly to human subjects, remains to be validated. (ii) \emph{No direct test of long-term stability.} LFP waveforms are known to remain stable over years, but the sessions tested here span weeks to months, and REALM has not been validated on degraded LFPs, as when electrode impedance drifts or channels fail. Ensuring that the distilled prior is robust to changes in LFP statistics is a precondition for any clinical use. (iii) \emph{The causal--bidirectional gap.} Even the largest causal student remains behind its bidirectional counterpart at matched capacity, and the gap after per-session fine-tuning is wider than under linear-probe evaluation (figure~\ref{fig:realtime}(a), (b); Supplementary Table~S7). Both appear to reflect an irreducible loss from removing future context, in the model and in the preprocessing, rather than a shortcoming of the distillation itself. (iv) \emph{Data scale.} The corpus is $5.9$\,h of M1 LFP from $34$ sessions and two subjects, far smaller than the spike-based corpora multi-session models are usually pretrained on; decoding also gains little from student capacity, which places the ceiling at the teacher or the data rather than at the student's architecture. (v) \emph{Energy not measured.} Latency and throughput were evaluated for edge deployment, but energy per inference was not. (vi) \emph{No content above $50$\,Hz.} The $100$\,Hz grid discards everything above $50$\,Hz, including the high-gamma power that band-power decoders use, so REALM cannot exploit it. No higher-bandwidth input or local-motor-potential-plus-band-power baseline was compared, so the conclusions apply to the low-frequency LFP setting studied here.

Several concrete steps follow from here. (i) The same retrospective-distillation workflow could be applied to other mesoscopic modalities such as electrocorticography and stereo-electroencephalography, where sizable public datasets already exist but the offline-to-online problem remains unresolved. (ii) Alternative objectives in unsupervised pretraining, such as contrastive learning or next-token prediction, may raise the representational power of the teacher and so of the student. A stronger feature extractor on the teacher side, such as the engineered spectral front ends built for offline LFP decoding~\cite{R111_li2026dualvct}, is a second route to the same end: RKD asks only that the teacher be effective, not that it be deployable, so an offline-only front end is an asset rather than an obstacle. (iii) A larger and more varied corpus may improve the teacher's representations, which the student then inherits. (iv) Prospective validation in long-term non-human primate experiments and, eventually, in humans with intracortical implants will need careful experimental planning to assess electrode degradation under realistic clinical conditions. (v) Pairing REALM with a fully implantable sensor and measuring energy per inference during closed-loop operation would help determine whether a multi-session pretrained, LFP-only encoder can serve as the decoder for a battery-free wireless iBCI.

\section{Conclusion}\label{sec:conclusion}
This study presents REALM (Retrospective Encoder Alignment for LFP Modeling), a framework that distills a non-causal LFP encoder into a causal model for real-time movement decoding. It shows that offline-to-online distillation, established in speech recognition, carries over to neural decoding: the contextual information learned by a bidirectional teacher can be transferred to a strictly causal student without behavioral labels during pretraining or distillation. The resulting decoder uses LFPs alone at every stage. It can be calibrated to a new session by fitting only a linear read-out rather than retraining the encoder, and it decodes one step in about $1.4$\,ms on a Jetson Orin Nano. On held-out sessions from two monkeys, REALM significantly outperforms every causal baseline evaluated under the same protocol, including CrossModalDistill, while using less than half as many parameters as CrossModalDistill's published student. The work has several limitations. Validation is restricted to laboratory recordings of two-dimensional reaching in non-human primates; generalization across subjects and long-term stability were not tested; and an accuracy gap relative to bidirectional decoding remains. Future work will focus on validation using human intracortical recordings, measurement of energy consumption on fully implantable hardware, and extension to other mesoscopic signals such as electrocorticography. Overall, REALM offers a practical route to accurate real-time decoding from a single low-bandwidth signal and provides a foundation for low-power wireless iBCIs for people with motor impairment.

\ack{We gratefully acknowledge the Ohio Supercomputer Center~\cite{OhioSupercomputerCenter1987} for providing the computational resources that made this work possible. We thank the authors and curators of the publicly available datasets used in this study, Makin et al.~\cite{R94_makin2018superior} and Flint et al.~\cite{R95_flint2012accurate}. Finally, we extend our heartfelt gratitude to the non-human primate subjects whose contributions were essential to this research: Monkey I (Makin) and Monkey C (Flint).}

\roles{
Conceptualization: PW, LD;\\
Methodology: PW, RM;\\
Software: PW;\\
Validation: PW, ZB;\\
Investigation: PW, ZB;\\
Visualization: PW, ZB, RM;\\
Writing -- original draft: PW;\\
Writing -- review \& editing: PW, ZB, RM, LD;\\
Supervision: LD.
}

\funding{This research received no specific grant from any funding agency in the public, commercial or not-for-profit sectors.}

\data{The Makin dataset is publicly available at \url{https://zenodo.org/record/583331}; the broadband recordings of the $30$ sessions used here are deposited as separate Zenodo records, one per session, whose DOIs are listed in the description of that record. The Flint dataset is publicly available at \url{https://crcns.org/data-sets/movements/dream}.  The entire codebase for the REALM experiments is publicly available at \url{https://github.com/percyance/REALM}.}

\suppdata{Supplementary material. Additional per-session results, representation-alignment analyses, streaming-inference measurements, complete training hyperparameters for every stage, dataset descriptions, and additional ablations on encoder backbone, the alignment term and layer-alignment depth.}

\section*{Ethics statement}
This study analyzed only publicly available data from previously published experiments~\cite{R94_makin2018superior, R93_odoherty2017nonhuman, R95_flint2012accurate}, and no new experiments with animals were performed. The animal procedures of those experiments were approved by the institutional animal care and use committees of the institutions where the data were recorded, as reported in the original publications.

\section*{Conflicts of interest}
The authors declare no conflicts of interest.

\clearpage
\bibliographystyle{iopart-num-long}
\bibliography{references}

\end{document}


\maketitle

\noindent\textbf{Description.} This document provides supplementary results and implementation details for the main manuscript. It contains per-session decoding results for the causal and bidirectional models of the main text, representation-alignment analyses, stacking and model-scaling results, streaming-inference latency on edge hardware, the hyperparameters of every training stage, baseline configurations, dataset and evaluation-split details, and additional ablations on the encoder backbone, the alignment depth, the causal deployment of the preprocessing chain and a time-ordered split.

\noindent\textbf{Rounding.} Throughout this document, differences and their statistics are computed from the unrounded session means, so a printed $\Delta R^2$ can differ by $0.001$ from the difference of the printed means.

\section{Detailed results}

\begin{table}[htbp]
\centering
\caption{\textbf{Linear-probe ablation, REALM internal.} Per-axis $R^2$ on the eight held-out sessions under causal deployment (Section~\ref{sec:ablation_causal_prep}; mean $\pm$ s.d.\ over three seeds) for the distillation ablation chain: RI, the architecture at initialization; PT-c, causal masked pretraining without retrospective knowledge distillation (RKD); PT, the bidirectional teacher, which draws on the whole $5$\,s segment, future samples included, and is scored on the zero-phase chain; REALM, that teacher distilled into the causal student. The lower block reports each baseline against REALM: the mean paired difference $\Delta R^2 = R^2_{\mathrm{REALM}}-R^2_{\mathrm{baseline}}$ over the eight session means, its $95\%$ confidence interval, Cohen's $d_z$, the number of sessions on which REALM is higher, and the two-sided paired-$t$ $p$-value.}
\label{tab:frozen_ablation}
\footnotesize
\setlength{\tabcolsep}{3pt}
\begin{tabular}{l cccc}
\toprule
Session & RI & PT-c & PT & \textbf{REALM} \\
\midrule
indy\_20160622\_01 & 0.255\std{0.024} & 0.313\std{0.016} & \textbf{0.519\std{0.024}} & 0.480\std{0.011} \\
indy\_20160630\_01 & 0.355\std{0.021} & 0.283\std{0.031} & \textbf{0.406\std{0.022}} & 0.385\std{0.015} \\
indy\_20160915\_01 & 0.286\std{0.011} & 0.398\std{0.035} & \textbf{0.592\std{0.029}} & 0.580\std{0.067} \\
indy\_20161013\_03 & 0.426\std{0.015} & 0.359\std{0.023} & 0.373\std{0.007} & \textbf{0.466\std{0.017}} \\
indy\_20170124\_01 & 0.411\std{0.001} & 0.354\std{0.024} & 0.375\std{0.021} & \textbf{0.458\std{0.002}} \\
Flint\_e1\_1 & 0.214\std{0.004} & 0.279\std{0.041} & \textbf{0.636\std{0.011}} & 0.543\std{0.014} \\
Flint\_e4\_1 & 0.247\std{0.033} & 0.224\std{0.052} & \textbf{0.595\std{0.051}} & 0.487\std{0.040} \\
Flint\_e5\_2 & 0.270\std{0.014} & 0.265\std{0.019} & \textbf{0.659\std{0.028}} & 0.568\std{0.008} \\
\midrule
Makin mean & 0.346\std{0.011} & 0.341\std{0.013} & 0.453\std{0.009} & \textbf{0.474\std{0.016}} \\
Flint mean & 0.244\std{0.011} & 0.256\std{0.030} & \textbf{0.630\std{0.007}} & 0.533\std{0.011} \\
\midrule
\textbf{Overall} & 0.308\std{0.004} & 0.309\std{0.007} & \textbf{0.519\std{0.003}} & 0.496\std{0.013} \\
\midrule
\textit{$\Delta R^2$ (REALM$-$)} & $+0.188$ & $+0.187$ & $-0.024$ & -- \\
\textit{95\% CI} & $[.081,.295]$ & $[.119,.254]$ & $[-.088,.041]$ & -- \\
\textit{Cohen's $d_z$} & $1.47$ & $2.30$ & $-0.31$ & -- \\
\textit{sessions} & $8/8$ & $8/8$ & $2/8$ & -- \\
\textit{$p$} & $0.004$ & $<0.001$ & $0.416$ & -- \\
\bottomrule
\end{tabular}
\end{table}

\begin{table}[htbp]
\centering
\caption{\textbf{Linear-probe baselines, external.} Per-axis $R^2$ on the eight held-out sessions (mean $\pm$ s.d.\ over three seeds) for REALM against two temporal backbones, a long short-term memory network (LSTM) and a Transformer, pretrained and read out exactly as REALM is but without RKD (Table~\ref{tab:backbone_ablation} gives both backbones distilled), and against CrossModalDistill (CMD) run causally under this protocol. Causal deployment (Section~\ref{sec:ablation_causal_prep}); Dual-VCT on the zero-phase chain. The lower block reports each baseline against REALM exactly as in Table~\ref{tab:frozen_ablation}. The other three published decoders, LFADS~\cite{R2_pandarinath2018inferring}, DPAD~\cite{R34_sani2024dissociative} and Dual-VCT~\cite{R111_li2026dualvct}, complete the comparison and, like CMD, are run under this protocol rather than quoted from their papers. Every column, DPAD included, is scored on the full test fold; Table~\ref{tab:artefact_sens} also gives DPAD's scores with the ten artifact segments dropped. DPAD alone uses labels before the read-out: it is fitted with the velocity of the $34$ non-held-out sessions as its behavior (Section~\ref{appendix-baselines}).}
\label{tab:frozen_baseline}
\footnotesize
\setlength{\tabcolsep}{2pt}
\begin{tabular}{l ccccccc}
\toprule
Session & LSTM & Transformer & LFADS & DPAD & Dual-VCT & CMD & \textbf{REALM} \\
\midrule
indy\_20160622\_01 & 0.449\std{0.031} & 0.439\std{0.020} & 0.181\std{0.010} & 0.454\std{0.017} & 0.237\std{0.015} & 0.400\std{0.006} & \textbf{0.480\std{0.011}} \\
indy\_20160630\_01 & 0.319\std{0.018} & 0.304\std{0.026} & 0.184\std{0.023} & 0.206\std{0.031} & 0.169\std{0.037} & 0.282\std{0.026} & \textbf{0.385\std{0.015}} \\
indy\_20160915\_01 & 0.463\std{0.038} & 0.486\std{0.048} & 0.222\std{0.028} & \textbf{0.604\std{0.047}} & 0.237\std{0.052} & 0.532\std{0.023} & 0.580\std{0.067} \\
indy\_20161013\_03 & 0.403\std{0.012} & 0.412\std{0.026} & 0.139\std{0.037} & 0.246\std{0.019} & 0.117\std{0.012} & 0.281\std{0.029} & \textbf{0.466\std{0.017}} \\
indy\_20170124\_01 & 0.380\std{0.016} & 0.396\std{0.023} & 0.181\std{0.052} & 0.205\std{0.011} & 0.191\std{0.018} & 0.244\std{0.034} & \textbf{0.458\std{0.002}} \\
Flint\_e1\_1 & 0.392\std{0.001} & 0.414\std{0.008} & 0.088\std{0.007} & 0.474\std{0.034} & 0.296\std{0.020} & 0.367\std{0.013} & \textbf{0.543\std{0.014}} \\
Flint\_e4\_1 & 0.304\std{0.056} & 0.330\std{0.068} & 0.022\std{0.037} & 0.276\std{0.296} & 0.024\std{0.284} & 0.224\std{0.080} & \textbf{0.487\std{0.040}} \\
Flint\_e5\_2 & 0.371\std{0.019} & 0.415\std{0.018} & -0.130\std{0.317} & 0.392\std{0.025} & 0.088\std{0.151} & 0.289\std{0.049} & \textbf{0.568\std{0.008}} \\
\midrule
Makin mean & 0.403\std{0.012} & 0.407\std{0.005} & 0.181\std{0.007} & 0.343\std{0.021} & 0.190\std{0.015} & 0.348\std{0.003} & \textbf{0.474\std{0.016}} \\
Flint mean & 0.356\std{0.013} & 0.386\std{0.026} & -0.007\std{0.105} & 0.381\std{0.094} & 0.136\std{0.102} & 0.294\std{0.039} & \textbf{0.533\std{0.011}} \\
\midrule
\textbf{Overall} & 0.385\std{0.008} & 0.399\std{0.009} & 0.111\std{0.043} & 0.357\std{0.036} & 0.170\std{0.043} & 0.328\std{0.016} & \textbf{0.496\std{0.013}} \\
\midrule
\textit{$\Delta R^2$ (REALM$-$)} & $+0.111$ & $+0.096$ & $+0.385$ & $+0.139$ & $+0.326$ & $+0.168$ & -- \\
\textit{95\% CI} & $[.060,.162]$ & $[.058,.134]$ & $[.256,.514]$ & $[.054,.224]$ & $[.241,.411]$ & $[.098,.239]$ & -- \\
\textit{Cohen's $d_z$} & $1.82$ & $2.13$ & $2.50$ & $1.37$ & $3.22$ & $1.99$ & -- \\
\textit{sessions} & $8/8$ & $8/8$ & $8/8$ & $7/8$ & $8/8$ & $8/8$ & -- \\
\textit{$p$} & $0.001$ & $<0.001$ & $<0.001$ & $0.006$ & $<0.001$ & $<0.001$ & -- \\
\bottomrule
\end{tabular}
\end{table}

\begin{table}[!htbp]
\centering
\caption{Representation alignment between model pairs, on the deepest encoder layer. Retrieval matches each teacher representation to the student representation of the same timestep among $15{,}500$ candidates, every $25$th timestep of $775$ non-overlapping $5$\,s segments drawn at random from the whole of the eight held-out sessions (one hundred per session, all $75$ of indy\_20160915\_01), so chance top-1 is $1/15{,}500$ and the chance median rank is $7750$. Representations are compared per timestep, as the distillation loss compares them. Teacher and student are the teacher and REALM of Table~\ref{tab:frozen_ablation}, each held-out session read by the REALM student distilled on its own corpus; REALM-PT-c is the same architecture and corpus without RKD and is the control the claim needs, REALM-RI is that architecture at initialization, and the last row replaces the student by Gaussian tensors. CKA is linear centered kernel alignment~\cite{R109_kornblith2019cka}, computed after normalizing each timestep's representation to zero mean and unit variance across its $256$ features, without a learned gain or bias. Arrows mark whether higher ($\uparrow$) or lower ($\downarrow$) is better.}
\label{tab:alignment}
\footnotesize
\setlength{\tabcolsep}{3pt}
\begin{tabular}{lcccc}
\toprule
Model pair & Top-1 $\uparrow$ & Top-5 $\uparrow$ & Median rank $\downarrow$ & CKA $\uparrow$ \\
\midrule
\textbf{\makecell[l]{Teacher \\ REALM}} & \textbf{0.9660} & \textbf{0.9845} & \textbf{1.0} & \textbf{0.754} \\
\midrule
\makecell[l]{Teacher \\ REALM-PT-c} & 0.0003 & 0.0010 & 5585.0 & 0.376 \\
\midrule
\makecell[l]{Teacher \\ REALM-RI} & 0.0002 & 0.0005 & 7376.0 & 0.341 \\
\midrule
\makecell[l]{REALM \\ REALM-RI} & 0.0003 & 0.0004 & 7068.0 & 0.416 \\
\midrule
\makecell[l]{Teacher \\ Gaussian tensors} & 0.0000 & 0.0003 & 7746.5 & 0.009 \\
\bottomrule
\end{tabular}
\end{table}

\begin{table}[!htbp]
\centering
\caption{Stacking REALM students, linear-probe evaluation on the eight held-out sessions. Column $k$ averages the predictions of $k$ base models, each distilled on a different four-fifths of the non-held-out sessions with the distillation seed held fixed, so the members differ in the data they saw rather than in an initialization. All members are scored on the same fold and the column is the mean over all $\binom{5}{k}$ subsets of that size; $k=1$ is a single model. Protocol as in Table~\ref{tab:frozen_baseline}: canonical $72/8/20$ split, the held-out set of CrossModalDistill~\cite{R24_erturk2025cross}, per-axis $R^2$, mean $\pm$ s.d.\ over the three evaluation splits (seeds $42$, $123$ and $456$). The held-out sessions are processed by the causal chain of Section~2.1 of the main text (causal deployment). Best per row in \textbf{bold}.}
\label{tab:stacking}
\footnotesize
\setlength{\tabcolsep}{3pt}
\begin{tabular}{lccccc}
\toprule
Session & $k=1$ & $k=2$ & $k=3$ & $k=4$ & $k=5$ \\
\midrule
indy\_20160622\_01 & 0.469\std{0.017} & 0.487\std{0.022} & 0.492\std{0.023} & 0.495\std{0.024} & \textbf{0.497\std{0.025}} \\
indy\_20160630\_01 & 0.379\std{0.018} & 0.399\std{0.018} & 0.405\std{0.019} & 0.409\std{0.020} & \textbf{0.411\std{0.021}} \\
indy\_20160915\_01 & 0.533\std{0.012} & 0.569\std{0.031} & 0.581\std{0.039} & 0.587\std{0.044} & \textbf{0.591\std{0.046}} \\
indy\_20161013\_03 & 0.481\std{0.009} & 0.502\std{0.010} & 0.509\std{0.011} & 0.512\std{0.011} & \textbf{0.514\std{0.012}} \\
indy\_20170124\_01 & 0.466\std{0.015} & 0.491\std{0.012} & 0.500\std{0.012} & 0.504\std{0.012} & \textbf{0.507\std{0.012}} \\
Flint\_e1\_1 & 0.513\std{0.014} & 0.522\std{0.015} & 0.526\std{0.015} & 0.527\std{0.015} & \textbf{0.528\std{0.015}} \\
Flint\_e4\_1 & 0.485\std{0.057} & 0.497\std{0.055} & 0.501\std{0.054} & 0.503\std{0.054} & \textbf{0.505\std{0.054}} \\
Flint\_e5\_2 & 0.550\std{0.007} & 0.560\std{0.007} & 0.564\std{0.007} & 0.566\std{0.007} & \textbf{0.567\std{0.006}} \\
\midrule
Makin mean & 0.465\std{0.003} & 0.489\std{0.008} & 0.497\std{0.010} & 0.502\std{0.011} & \textbf{0.504\std{0.012}} \\
Flint mean & 0.516\std{0.015} & 0.527\std{0.013} & 0.530\std{0.013} & 0.532\std{0.012} & \textbf{0.533\std{0.012}} \\
\midrule
\textbf{Overall} & 0.484\std{0.007} & 0.503\std{0.009} & 0.510\std{0.009} & 0.513\std{0.010} & \textbf{0.515\std{0.010}} \\
\bottomrule
\end{tabular}
\end{table}

\begin{table}[htbp]
\centering
\caption{\textbf{Fine-tuned ablation, REALM internal.} Per-axis $R^2$ on the eight held-out sessions (mean $\pm$ s.d.\ over three seeds) with the encoder unfrozen and adapted on each session, for the same ablation chain as Table~\ref{tab:frozen_ablation}. The lower block reports each baseline against REALM as in Table~\ref{tab:frozen_ablation}; PT is the bidirectional teacher, which sees the future of each segment and is an offline upper bound rather than a deployable model. Causal deployment (Section~\ref{sec:ablation_causal_prep}); PT on the zero-phase chain.}
\label{tab:finetuned_ablation}
\footnotesize
\setlength{\tabcolsep}{3pt}
\begin{tabular}{l cccc}
\toprule
Session & RI & PT-c & PT & \textbf{REALM} \\
\midrule
indy\_20160622\_01 & 0.541\std{0.028} & 0.662\std{0.008} & \textbf{0.765\std{0.015}} & 0.715\std{0.007} \\
indy\_20160630\_01 & 0.401\std{0.014} & 0.446\std{0.016} & \textbf{0.725\std{0.010}} & 0.509\std{0.026} \\
indy\_20160915\_01 & 0.357\std{0.019} & 0.527\std{0.028} & \textbf{0.710\std{0.034}} & 0.603\std{0.072} \\
indy\_20161013\_03 & 0.528\std{0.056} & 0.602\std{0.018} & \textbf{0.700\std{0.028}} & 0.671\std{0.029} \\
indy\_20170124\_01 & 0.592\std{0.005} & 0.657\std{0.033} & \textbf{0.744\std{0.020}} & 0.692\std{0.034} \\
Flint\_e1\_1 & 0.553\std{0.018} & 0.614\std{0.012} & \textbf{0.835\std{0.001}} & 0.679\std{0.026} \\
Flint\_e4\_1 & 0.473\std{0.074} & 0.501\std{0.079} & \textbf{0.749\std{0.071}} & 0.598\std{0.089} \\
Flint\_e5\_2 & 0.488\std{0.028} & 0.558\std{0.012} & \textbf{0.783\std{0.031}} & 0.625\std{0.026} \\
\midrule
Makin mean & 0.484\std{0.017} & 0.579\std{0.011} & \textbf{0.729\std{0.009}} & 0.638\std{0.012} \\
Flint mean & 0.505\std{0.027} & 0.558\std{0.026} & \textbf{0.789\std{0.021}} & 0.634\std{0.035} \\
\midrule
\textbf{Overall} & 0.492\std{0.012} & 0.571\std{0.017} & \textbf{0.751\std{0.013}} & 0.636\std{0.021} \\
\midrule
\textit{$\Delta R^2$ (REALM$-$)} & $+0.145$ & $+0.066$ & $-0.115$ & -- \\
\textit{95\% CI} & $[.106,.184]$ & $[.051,.080]$ & $[-.170,-.059]$ & -- \\
\textit{Cohen's $d_z$} & $3.10$ & $3.71$ & $-1.73$ & -- \\
\textit{sessions} & $8/8$ & $8/8$ & $0/8$ & -- \\
\textit{$p$} & $<0.001$ & $<0.001$ & $0.002$ & -- \\
\bottomrule
\end{tabular}
\end{table}

\begin{table}[htbp]
\centering
\caption{\textbf{Fine-tuned baselines, external.} Per-axis $R^2$ on the eight held-out sessions (mean $\pm$ s.d.\ over three seeds), the same models as in Table~\ref{tab:frozen_baseline}, with the encoder unfrozen and adapted on each held-out session. The lower block reports each baseline against REALM as in Table~\ref{tab:frozen_ablation}. As there, all four published decoders, CMD, LFADS~\cite{R2_pandarinath2018inferring}, DPAD~\cite{R34_sani2024dissociative} and Dual-VCT~\cite{R111_li2026dualvct}, are run under this protocol rather than quoted from their papers. Only DPAD is scored with the ten artifact segments of Table~\ref{tab:artefact_sens} dropped from the test fold, because the transients they carry pull its per-session fit down; dropping them raises its score by $+0.029$. Every other column is on the full fold, so the asymmetry is conservative for REALM's margin. DPAD is fitted afresh on each held-out session and early-stopped on its validation fold, from which any of those segments is dropped as well (Section~\ref{appendix-baselines}). Causal deployment (Section~\ref{sec:ablation_causal_prep}); Dual-VCT on the zero-phase chain.}
\label{tab:finetuned_baseline}
\footnotesize
\setlength{\tabcolsep}{2pt}
\begin{tabular}{l ccccccc}
\toprule
Session & LSTM & Transformer & LFADS & DPAD & Dual-VCT & CMD & \textbf{REALM} \\
\midrule
indy\_20160622\_01 & 0.658\std{0.008} & 0.701\std{0.003} & 0.640\std{0.019} & 0.489\std{0.024} & 0.474\std{0.021} & 0.710\std{0.012} & \textbf{0.715\std{0.007}} \\
indy\_20160630\_01 & 0.374\std{0.013} & 0.449\std{0.013} & 0.462\std{0.009} & 0.482\std{0.020} & 0.412\std{0.037} & 0.471\std{0.014} & \textbf{0.509\std{0.026}} \\
indy\_20160915\_01 & 0.565\std{0.022} & \textbf{0.622\std{0.039}} & 0.339\std{0.221} & 0.496\std{0.042} & 0.259\std{0.072} & 0.580\std{0.052} & 0.603\std{0.072} \\
indy\_20161013\_03 & 0.555\std{0.010} & 0.658\std{0.003} & 0.571\std{0.093} & 0.534\std{0.035} & 0.278\std{0.060} & 0.630\std{0.029} & \textbf{0.671\std{0.029}} \\
indy\_20170124\_01 & 0.588\std{0.007} & 0.669\std{0.010} & 0.631\std{0.017} & 0.545\std{0.083} & 0.362\std{0.065} & 0.659\std{0.002} & \textbf{0.692\std{0.034}} \\
Flint\_e1\_1 & 0.628\std{0.012} & 0.632\std{0.020} & 0.582\std{0.018} & 0.545\std{0.018} & 0.486\std{0.065} & 0.628\std{0.045} & \textbf{0.679\std{0.026}} \\
Flint\_e4\_1 & 0.516\std{0.043} & 0.561\std{0.069} & 0.410\std{0.076} & 0.385\std{0.063} & 0.328\std{0.110} & 0.535\std{0.091} & \textbf{0.598\std{0.089}} \\
Flint\_e5\_2 & 0.567\std{0.032} & 0.622\std{0.007} & 0.392\std{0.085} & 0.474\std{0.053} & 0.336\std{0.038} & 0.594\std{0.028} & \textbf{0.625\std{0.026}} \\
\midrule
Makin mean & 0.548\std{0.010} & 0.620\std{0.010} & 0.529\std{0.069} & 0.509\std{0.027} & 0.357\std{0.033} & 0.610\std{0.016} & \textbf{0.638\std{0.012}} \\
Flint mean & 0.570\std{0.020} & 0.605\std{0.017} & 0.461\std{0.041} & 0.468\std{0.020} & 0.383\std{0.057} & 0.586\std{0.010} & \textbf{0.634\std{0.035}} \\
\midrule
\textbf{Overall} & 0.556\std{0.013} & 0.614\std{0.009} & 0.503\std{0.045} & 0.494\std{0.016} & 0.367\std{0.038} & 0.601\std{0.013} & \textbf{0.636\std{0.021}} \\
\midrule
\textit{$\Delta R^2$ (REALM$-$)} & $+0.080$ & $+0.022$ & $+0.133$ & $+0.143$ & $+0.270$ & $+0.035$ & -- \\
\textit{95\% CI} & $[.051,.109]$ & $[.001,.043]$ & $[.063,.203]$ & $[.091,.194]$ & $[.191,.348]$ & $[.021,.050]$ & -- \\
\textit{Cohen's $d_z$} & $2.29$ & $0.89$ & $1.60$ & $2.31$ & $2.88$ & $2.07$ & -- \\
\textit{sessions} & $8/8$ & $7/8$ & $8/8$ & $8/8$ & $8/8$ & $8/8$ & -- \\
\textit{$p$} & $<0.001$ & $0.040$ & $0.003$ & $<0.001$ & $<0.001$ & $<0.001$ & -- \\
\bottomrule
\end{tabular}
\end{table}

\begin{table}[!htbp]
\centering
\caption{Scaling of the causal (upper block) and bidirectional (lower block) students, under linear-probe evaluation and after per-session fine-tuning, on the eight held-out sessions. Within each family the three sizes differ in depth and state-space expansion factor at $d_{\text{model}}=256$ ($8$, $10$ and $22$ layers causal; $3$, $8$ and $10$ bidirectional) and are distilled from the same teacher under the same objective. Under linear-probe evaluation, the curve is nearly flat past $4.9$\,M: REALM-L is ahead of REALM by $-0.003$ overall ($[-0.018, +0.013]$, $d_z = -0.14$, $6/8$, $p = 0.704$), a small but consistent step, so capacity has nearly saturated there and the ceiling sits at the teacher or the corpus rather than at the student's architecture. After fine-tuning, REALM-L is ahead by $-0.008$ ($[-0.037, +0.020]$, $d_z = -0.24$, $3/8$, $p = 0.521$), again small against the doubling of parameters. The causal sizes are scored on the held-out sessions processed by the causal chain of Section~2.1 of the main text, the bidirectional ones on the zero-phase chain. REALM's fine-tuned column is the REALM column of Table~\ref{tab:finetuned_baseline}, cell for cell. Canonical $72/8/20$ split, CrossModalDistill's held-out set, per-axis $R^2$, mean $\pm$ s.d.\ over three seeds ($s\in\{42,123,456\}$); the linear probe selects its ridge penalty per session on the validation fold, as everywhere else in this paper. Best per row within each block in \textbf{bold}.}
\label{tab:scaling}
\footnotesize
\setlength{\tabcolsep}{3pt}
\begin{tabular}{l ccc ccc}
\toprule
\multirow{2}{*}{Session} & \multicolumn{3}{c}{Linear probe} & \multicolumn{3}{c}{Fine-tuned} \\
\cmidrule(lr){2-4}\cmidrule(lr){5-7}
 & REALM-S & REALM & REALM-L & REALM-S & REALM & REALM-L \\
 & (2.1\,M) & (4.9\,M) & (10.5\,M) & (2.1\,M) & (4.9\,M) & (10.5\,M) \\
\midrule
indy\_20160622\_01 & 0.468\std{0.027} & 0.480\std{0.011} & \textbf{0.484\std{0.018}} & 0.704\std{0.005} & 0.715\std{0.007} & \textbf{0.728\std{0.009}} \\
indy\_20160630\_01 & 0.326\std{0.102} & 0.385\std{0.015} & \textbf{0.392\std{0.021}} & 0.498\std{0.017} & \textbf{0.509\std{0.026}} & 0.505\std{0.042} \\
indy\_20160915\_01 & 0.500\std{0.029} & \textbf{0.580\std{0.067}} & 0.543\std{0.031} & 0.588\std{0.047} & \textbf{0.603\std{0.072}} & 0.585\std{0.017} \\
indy\_20161013\_03 & 0.471\std{0.013} & 0.466\std{0.017} & \textbf{0.471\std{0.016}} & 0.662\std{0.012} & 0.671\std{0.029} & \textbf{0.691\std{0.025}} \\
indy\_20170124\_01 & 0.460\std{0.017} & 0.458\std{0.002} & \textbf{0.466\std{0.008}} & 0.680\std{0.008} & 0.692\std{0.034} & \textbf{0.711\std{0.009}} \\
Flint\_e1\_1 & 0.485\std{0.016} & 0.543\std{0.014} & \textbf{0.552\std{0.010}} & 0.626\std{0.046} & \textbf{0.679\std{0.026}} & 0.676\std{0.041} \\
Flint\_e4\_1 & 0.449\std{0.065} & \textbf{0.487\std{0.040}} & 0.457\std{0.068} & 0.545\std{0.044} & \textbf{0.598\std{0.089}} & 0.511\std{0.158} \\
Flint\_e5\_2 & 0.521\std{0.025} & 0.568\std{0.008} & \textbf{0.579\std{0.016}} & 0.583\std{0.027} & \textbf{0.625\std{0.026}} & 0.618\std{0.070} \\
\midrule
Makin mean & 0.445\std{0.028} & \textbf{0.474\std{0.016}} & 0.471\std{0.014} & 0.626\std{0.010} & 0.638\std{0.012} & \textbf{0.644\std{0.012}} \\
Flint mean & 0.485\std{0.013} & \textbf{0.533\std{0.011}} & 0.530\std{0.022} & 0.585\std{0.012} & \textbf{0.634\std{0.035}} & 0.602\std{0.049} \\
\textbf{Overall} & 0.460\std{0.018} & \textbf{0.496\std{0.013}} & 0.493\std{0.010} & 0.611\std{0.008} & \textbf{0.636\std{0.021}} & 0.628\std{0.023} \\
\midrule
 & REALM-bi-S & REALM-bi & REALM-bi-L & REALM-bi-S & REALM-bi & REALM-bi-L \\
 & (2.2\,M) & (5.5\,M) & (11.5\,M) & (2.2\,M) & (5.5\,M) & (11.5\,M) \\
\midrule
indy\_20160622\_01 & 0.552\std{0.027} & \textbf{0.560\std{0.029}} & 0.559\std{0.027} & 0.742\std{0.022} & 0.749\std{0.023} & \textbf{0.777\std{0.010}} \\
indy\_20160630\_01 & 0.518\std{0.012} & 0.528\std{0.013} & \textbf{0.530\std{0.015}} & 0.747\std{0.031} & 0.754\std{0.016} & \textbf{0.755\std{0.025}} \\
indy\_20160915\_01 & \textbf{0.650\std{0.035}} & 0.643\std{0.037} & 0.633\std{0.033} & \textbf{0.682\std{0.038}} & 0.669\std{0.066} & 0.679\std{0.040} \\
indy\_20161013\_03 & 0.510\std{0.019} & \textbf{0.515\std{0.014}} & 0.491\std{0.014} & 0.668\std{0.020} & 0.721\std{0.027} & \textbf{0.737\std{0.023}} \\
indy\_20170124\_01 & 0.494\std{0.025} & 0.487\std{0.034} & \textbf{0.499\std{0.027}} & 0.742\std{0.026} & \textbf{0.753\std{0.016}} & 0.737\std{0.030} \\
Flint\_e1\_1 & 0.635\std{0.016} & 0.660\std{0.022} & \textbf{0.690\std{0.006}} & 0.791\std{0.003} & 0.810\std{0.008} & \textbf{0.827\std{0.004}} \\
Flint\_e4\_1 & 0.607\std{0.056} & 0.626\std{0.055} & \textbf{0.632\std{0.061}} & 0.708\std{0.079} & 0.722\std{0.049} & \textbf{0.726\std{0.066}} \\
Flint\_e5\_2 & 0.653\std{0.013} & 0.683\std{0.017} & \textbf{0.699\std{0.030}} & 0.726\std{0.058} & \textbf{0.766\std{0.037}} & 0.759\std{0.057} \\
\midrule
Makin mean & 0.545\std{0.003} & \textbf{0.547\std{0.004}} & 0.542\std{0.003} & 0.716\std{0.010} & 0.729\std{0.016} & \textbf{0.737\std{0.005}} \\
Flint mean & 0.632\std{0.011} & 0.656\std{0.020} & \textbf{0.674\std{0.016}} & 0.742\std{0.028} & 0.766\std{0.016} & \textbf{0.771\std{0.028}} \\
\textbf{Overall} & 0.577\std{0.004} & 0.588\std{0.008} & \textbf{0.592\std{0.007}} & 0.726\std{0.006} & 0.743\std{0.015} & \textbf{0.750\std{0.012}} \\
\bottomrule
\end{tabular}
\end{table}

\begin{table}[htbp]
\centering
\caption{Streaming inference latency of the three causal REALM variants on a Raspberry Pi 5 (FP32 on one thread; INT8 with the process pinned to three cores) and a Jetson Orin Nano (PyTorch CPU FP32 on one thread with core pinning; PyTorch CUDA Graphs and TensorRT (TRT) FP16 on the integrated Ampere GPU). Throughput in frames per second (fps) is the reciprocal of the median (p50) step time, fps $= 1000/\mathrm{p50}$ with p50 in ms and not rounded; p95 is the 95th-percentile step time. FP32 and FP16 denote 32- and 16-bit floating point, and INT8 dynamic 8-bit integer quantization of all linear layers. The Jetson rows were measured in the uncapped MAXN SUPER power mode with the CPU, GPU and memory clocks locked at their maximum (CPU $1.728$\,GHz, GPU $1.02$\,GHz), with PyTorch~2.5 and TensorRT~10.3; each configuration was warmed up with $50$ calls and timed over $500$. Every row is the wall-clock time of one decoder step called from Python and synchronized; on the GPU, the step is captured once in a CUDA graph, a PyTorch graph or one TensorRT engine execution, and replayed. For reference, the GPU compute time of the same TensorRT engines reported by \texttt{trtexec} is $0.26$, $0.41$ and $0.85$\,ms. The Raspberry Pi 5 rows are from an earlier measurement, also warmed up with $50$ calls and timed over $500$, in which the CPU frequency governor was not recorded. Every row times the whole decoder step (front end, Mamba-2 stack and read-out head), whereas the streaming cost in Table~\ref{tab:backbone_ablation} times the temporal stack alone. Best result per model in \textbf{bold}.}
\label{tab:inference}
\footnotesize
\setlength{\tabcolsep}{3pt}
\begin{tabular}{llccccc}
\toprule
Model & Config & Params & Mean (ms) & p50 (ms) & p95 (ms) & fps \\
\midrule
\multirow{5}{*}{REALM-S}
 & Pi 5 FP32, 1-thread & 2.11\,M & 15.30 & 12.72 & 31.05 & 79 \\
 & Pi 5 INT8, 3 pinned cores & 2.11\,M & 8.33 & 7.87 & 9.80 & 127 \\
 & Jetson CPU FP32, 1-thread, pinned & 2.11\,M & 8.14 & 8.14 & 8.27 & 123 \\
 & Jetson CUDA Graphs & 2.11\,M & 0.98 & 0.98 & 0.99 & 1025 \\
 & Jetson TRT FP16 & 2.11\,M & \textbf{0.30} & \textbf{0.30} & \textbf{0.30} & \textbf{3342} \\
\midrule
\multirow{5}{*}{REALM}
 & Pi 5 FP32, 1-thread & 4.91\,M & 25.45 & 23.75 & 43.45 & 42 \\
 & Pi 5 INT8, 3 pinned cores & 4.91\,M & 12.03 & 11.70 & 14.28 & 85 \\
 & Jetson CPU FP32, 1-thread, pinned & 4.91\,M & 13.89 & 13.89 & 14.05 & 72 \\
 & Jetson CUDA Graphs & 4.91\,M & 1.42 & 1.42 & 1.43 & 703 \\
 & Jetson TRT FP16 & 4.91\,M & \textbf{0.46} & \textbf{0.46} & \textbf{0.46} & \textbf{2183} \\
\midrule
\multirow{5}{*}{REALM-L}
 & Pi 5 FP32, 1-thread & 10.50\,M & 56.01 & 53.64 & 83.45 & 19 \\
 & Pi 5 INT8, 3 pinned cores & 10.50\,M & 25.53 & 25.19 & 30.22 & 40 \\
 & Jetson CPU FP32, 1-thread, pinned & 10.50\,M & 29.19 & 29.20 & 29.44 & 34 \\
 & Jetson CUDA Graphs & 10.50\,M & 2.95 & 2.95 & 2.96 & 339 \\
 & Jetson TRT FP16 & 10.50\,M & \textbf{0.89} & \textbf{0.89} & \textbf{0.90} & \textbf{1121} \\
\bottomrule
\end{tabular}
\end{table}

\begin{table}[htbp]
\centering
\footnotesize
\setlength{\tabcolsep}{3pt}
\caption{\textbf{Bidirectional linear-probe ablation, REALM internal.} The distillation chain of Table~\ref{tab:frozen_ablation} run with a bidirectional student: PT-bi, the REALM-bi architecture given masked pretraining alone, without RKD (the bidirectional counterpart of PT-c, not the teacher); PT, the bidirectional teacher; REALM-bi, that teacher distilled into a bidirectional student. REALM (causal) is carried across from Table~\ref{tab:frozen_ablation} so the cost of causality can be read off the same rows. Per-axis $R^2$ on the eight held-out sessions, mean $\pm$ s.d.\ over three seeds. The lower block reports each baseline against REALM-bi exactly as in Table~\ref{tab:frozen_ablation}.}
\label{tab:bi_ablation}
\begin{tabular}{l cccc}
\toprule
Session & PT-bi & PT & REALM (causal) & \textbf{REALM-bi} \\
\midrule
indy\_20160622\_01 & 0.493\std{0.016} & 0.519\std{0.024} & 0.530\std{0.015} & \textbf{0.560\std{0.029}} \\
indy\_20160630\_01 & 0.409\std{0.029} & 0.406\std{0.022} & 0.475\std{0.012} & \textbf{0.528\std{0.013}} \\
indy\_20160915\_01 & 0.553\std{0.044} & 0.592\std{0.029} & 0.621\std{0.057} & \textbf{0.643\std{0.037}} \\
indy\_20161013\_03 & 0.402\std{0.036} & 0.373\std{0.007} & 0.488\std{0.011} & \textbf{0.515\std{0.014}} \\
indy\_20170124\_01 & 0.387\std{0.011} & 0.375\std{0.021} & 0.478\std{0.005} & \textbf{0.487\std{0.034}} \\
Flint\_e1\_1 & 0.612\std{0.027} & 0.636\std{0.011} & 0.626\std{0.016} & \textbf{0.660\std{0.022}} \\
Flint\_e4\_1 & 0.562\std{0.076} & 0.595\std{0.051} & 0.616\std{0.046} & \textbf{0.626\std{0.055}} \\
Flint\_e5\_2 & 0.624\std{0.013} & 0.659\std{0.028} & 0.649\std{0.012} & \textbf{0.683\std{0.017}} \\
\midrule
Makin mean & 0.449\std{0.015} & 0.453\std{0.009} & 0.519\std{0.011} & \textbf{0.547\std{0.004}} \\
Flint mean & 0.599\std{0.030} & 0.630\std{0.007} & 0.630\std{0.006} & \textbf{0.656\std{0.020}} \\
\midrule
\textbf{Overall} & 0.505\std{0.015} & 0.519\std{0.003} & 0.561\std{0.008} & \textbf{0.588\std{0.008}} \\
\midrule
\textit{$\Delta R^2$ (REALM-bi$-$)} & $+0.083$ & $+0.068$ & $+0.027$ & -- \\
\textit{95\% CI} & $[.060,.105]$ & $[.027,.109]$ & $[.015,.039]$ & -- \\
\textit{Cohen's $d_z$} & $3.09$ & $1.40$ & $1.89$ & -- \\
\textit{sessions} & $8/8$ & $8/8$ & $8/8$ & -- \\
\textit{$p$} & $<0.001$ & $0.006$ & $0.001$ & -- \\
\bottomrule
\end{tabular}
\end{table}

\begin{table}[htbp]
\centering
\footnotesize
\setlength{\tabcolsep}{2pt}
\caption{\textbf{Bidirectional linear-probe baselines, external.} REALM-bi against a bidirectional LSTM and a bidirectional Transformer, trained by bidirectional masked pretraining alone, without RKD, and against CMD in its bidirectional form, so that neither side of the comparison is denied future context. Layout, read-out and statistics exactly as in Table~\ref{tab:frozen_baseline}. The other three published decoders, LFADS~\cite{R2_pandarinath2018inferring}, DPAD~\cite{R34_sani2024dissociative} and Dual-VCT~\cite{R111_li2026dualvct}, complete the comparison and, like CMD, are run under this protocol rather than quoted from their papers; DPAD, the only model fitted with labels before the read-out (the velocity of the $34$ non-held-out sessions), is read non-causally through the bidirectional recurrent layers of its reference implementation (Section~\ref{appendix-baselines}). Every column, DPAD included, is scored on the full test fold; Table~\ref{tab:artefact_sens} also gives DPAD's scores with the ten artifact segments dropped.}
\label{tab:bi_baseline}
\begin{tabular}{l ccccccc}
\toprule
Session & LSTM & Transformer & LFADS & DPAD & Dual-VCT & CMD (bi) & \textbf{REALM-bi} \\
\midrule
indy\_20160622\_01 & 0.504\std{0.023} & 0.424\std{0.034} & 0.157\std{0.023} & 0.546\std{0.032} & 0.220\std{0.057} & 0.465\std{0.024} & \textbf{0.560\std{0.029}} \\
indy\_20160630\_01 & 0.440\std{0.032} & 0.419\std{0.008} & 0.303\std{0.028} & 0.354\std{0.035} & 0.237\std{0.034} & 0.321\std{0.018} & \textbf{0.528\std{0.013}} \\
indy\_20160915\_01 & 0.551\std{0.017} & 0.488\std{0.029} & 0.181\std{0.037} & \textbf{0.678\std{0.051}} & 0.278\std{0.066} & 0.586\std{0.036} & 0.643\std{0.037} \\
indy\_20161013\_03 & 0.423\std{0.016} & 0.381\std{0.025} & 0.189\std{0.021} & 0.309\std{0.035} & 0.115\std{0.053} & 0.216\std{0.003} & \textbf{0.515\std{0.014}} \\
indy\_20170124\_01 & 0.394\std{0.013} & 0.392\std{0.037} & 0.208\std{0.023} & 0.272\std{0.023} & 0.177\std{0.057} & 0.198\std{0.052} & \textbf{0.487\std{0.034}} \\
Flint\_e1\_1 & 0.591\std{0.031} & 0.584\std{0.047} & 0.110\std{0.015} & 0.643\std{0.032} & 0.270\std{0.115} & 0.606\std{0.049} & \textbf{0.660\std{0.022}} \\
Flint\_e4\_1 & 0.499\std{0.053} & 0.528\std{0.055} & 0.121\std{0.020} & 0.464\std{0.294} & 0.208\std{0.052} & 0.450\std{0.086} & \textbf{0.626\std{0.055}} \\
Flint\_e5\_2 & 0.542\std{0.057} & 0.609\std{0.048} & 0.128\std{0.019} & 0.439\std{0.278} & 0.157\std{0.040} & 0.546\std{0.019} & \textbf{0.683\std{0.017}} \\
\midrule
Makin mean & 0.462\std{0.012} & 0.421\std{0.008} & 0.208\std{0.019} & 0.432\std{0.023} & 0.205\std{0.037} & 0.357\std{0.006} & \textbf{0.547\std{0.004}} \\
Flint mean & 0.544\std{0.033} & 0.574\std{0.019} & 0.120\std{0.003} & 0.515\std{0.109} & 0.212\std{0.068} & 0.534\std{0.035} & \textbf{0.656\std{0.020}} \\
\midrule
\textbf{Overall} & 0.493\std{0.012} & 0.478\std{0.008} & 0.175\std{0.012} & 0.463\std{0.048} & 0.208\std{0.048} & 0.424\std{0.011} & \textbf{0.588\std{0.008}} \\
\midrule
\textit{$\Delta R^2$ (REALM-bi$-$)} & $+0.095$ & $+0.110$ & $+0.413$ & $+0.124$ & $+0.380$ & $+0.164$ & -- \\
\textit{95\% CI} & $[.071,.118]$ & $[.085,.135]$ & $[.308,.518]$ & $[.034,.215]$ & $[.319,.441]$ & $[.084,.245]$ & -- \\
\textit{Cohen's $d_z$} & $3.40$ & $3.71$ & $3.29$ & $1.15$ & $5.17$ & $1.70$ & -- \\
\textit{sessions} & $8/8$ & $8/8$ & $8/8$ & $7/8$ & $8/8$ & $8/8$ & -- \\
\textit{$p$} & $<0.001$ & $<0.001$ & $<0.001$ & $0.014$ & $<0.001$ & $0.002$ & -- \\
\bottomrule
\end{tabular}
\end{table}

\clearpage
\section{Training details and hyperparameters}\label{appendix-training}

Masked pretraining runs for up to 300 epochs with early stopping (patience 30) on the primary motor cortex (M1) channels of Makin and Flint, with a segment stride of half the segment length for augmentation. The pretrained bidirectional teacher is then distilled separately on the Makin and on the Flint non-held-out sessions, for up to 300 epochs with patience 30, giving one student per corpus and seed; each held-out session is decoded by the student of its own corpus. Each student's neural tokenizer is initialized as a copy of the teacher's, and its Mamba-2 layers at random (Table~\ref{tab:hp_distill_unsup}). The distillation objective carries no velocity labels: the alignment and reconstruction terms are the only training signal. Each distilled student is then evaluated in two regimes. The linear-probe regime keeps the encoder exactly as distilled and fits a closed-form read-out on each held-out session, so no gradient reaches the encoder at any point after distillation. The fine-tuned regime adapts the whole encoder on each held-out session with velocity labels, which is the only stage in either regime where labels reach the encoder; in the linear-probe regime they enter only the closed-form read-out. Hyperparameters for all stages are summarized in Tables~\ref{tab:hp-pretrain}--\ref{tab:hp_finetune}. Masked pretraining of the bidirectional teacher runs on $16$ NVIDIA A100 GPUs and completes in $43$ minutes of wall-clock time; every later stage --- distillation, the linear probe and per-session fine-tuning --- runs on a single A100, where distilling one REALM student takes $67$--$82$\,min on Makin and $29$--$34$\,min on Flint; Tables~\ref{tab:hp-pretrain}--\ref{tab:hp_finetune} give the wall-clock time of each stage. CrossModalDistill's student, at its published size of ten Transformer layers of width $256$, carries $10.59$\,M parameters against REALM's $4.91$\,M (Table~\ref{tab:baseline_config}), the comparison behind the $2\times$ reduction in parameter count quoted in the abstract.

\begin{table}[htbp]
\centering
\footnotesize
\setlength{\tabcolsep}{3pt}
\caption{Hyperparameters for masked pretraining. The values are those of the bidirectional teacher (REALM-PT). REALM-PT-c and REALM-PT-bi use the same settings except that each has the encoder of the student it controls for, REALM's causal Mamba-2 ($4.91$\,M parameters) and REALM-bi's ($5.54$\,M), against the teacher's $11.50$\,M, and is trained on a single GPU with batch $48$ and without drop path; REALM-PT-c is pretrained with seed $42$ and REALM-PT-bi with seeds $42$, $123$ and $456$.}
\label{tab:hp-pretrain}
\begin{tabular}{lc}
\toprule
\textbf{Hyperparameters} & \textbf{Values} \\
\midrule
Batch size & 24 per GPU (384 across 16 GPUs) \\
Peak learning rate & $1.53\times10^{-3}$ \\
Learning rate scheduler & Linear warmup, then exponential decay (0.995 per epoch) \\
Warmup epochs & 30 (starting at $0.3\times$ the peak) \\
Optimizer & AdamW \\
Weight decay & $10^{-5}$ \\
Total epochs & 300 \\
Gradient clipping & 1.0 \\
Mask ratio & 0.6 \\
Mask type & Block (10--50 steps) \\
Segment length / stride & 500 / 250 \\
Channel dropout & 0.15 \\
Amplitude jitter & Uniform(0.85, 1.15) \\
Gaussian noise $\sigma$ & 0.05 \\
Drop path & $0 \to 0.1$ \\
Dropout & 0.1 \\
Encoder & BiMamba-2, 10 layers, $d_{\text{model}}=256$, depthwise convolution width 4 \\
\quad Expansion factor / heads $H$ & 2 / 8 per direction \\
\quad State dimension $N$ / head dimension $P$ & 64 / 64 \\
Predictor depth / expansion factor & 1 / 1 \\
Early stopping patience & 30 \\
Session embedding & None \\
Corpus & Makin and Flint, M1 channels only \\
Held-out set & The 8 sessions held out by CrossModalDistill (5 Makin, 3 Flint) \\
Seed & 42 \\
Hardware / wall-clock time & 16 NVIDIA A100 GPUs / 43\,min \\
\bottomrule
\end{tabular}
\end{table}

\begin{table}[htbp]
\centering
\footnotesize
\setlength{\tabcolsep}{3pt}
\caption{Hyperparameters for retrospective distillation, the configuration behind every linear-probe and fine-tuned REALM result in this work. Apart from the tokenizer initialization, the alignment term is the only training signal from the teacher. Every student shares the teacher's $d_{\text{model}}=256$, state dimension $N=64$, head dimension $P=64$ and depthwise convolution width 4 (Table~\ref{tab:hp-pretrain}); a bidirectional student has $H$ heads per direction.}
\label{tab:hp_distill_unsup}
\begin{tabular}{lc}
\toprule
\textbf{Hyperparameters} & \textbf{Values} \\
\midrule
Batch size & 32 \\
Peak learning rate & $5\times10^{-4}$ \\
Minimum learning rate & $5\times10^{-6}$ \\
Learning rate scheduler & Cosine annealing (no warmup) \\
Optimizer & AdamW \\
Weight decay & $10^{-5}$ \\
Total epochs & 300 \\
Early stopping patience & 30 \\
Gradient clipping & 1.0 \\
Train / val split & 80\% / 20\% of the non-held-out sessions, split by session \\
Segment length & 500 samples (5\,s at 100\,Hz) \\
\midrule
$\lambda_{\text{repr}}$ (alignment term) & 1.0 \\
$\lambda_{\text{ae}}$ (reconstruction term) & 1.0 \\
Reconstruction-loss timesteps & 0.3 of each segment, drawn at random \\
Alignment layers & 1 (final layer only) \\
Reconstruction head & Two linear layers with a GELU between them \\
\midrule
Student (causal) & REALM-S / REALM / REALM-L (2.11 / 4.91 / 10.50\,M) \\
\quad Layers (expansion factor, heads $H$) & 8 (1, 4) / 10 (2, 8) / 22 (2, 8) \\
Student (bidirectional) & REALM-bi-S / REALM-bi / REALM-bi-L (2.23 / 5.54 / 11.54\,M) \\
\quad Layers (expansion factor, heads $H$) & 3 (1, 4) / 8 (1, 4) / 10 (2, 8) \\
Student initialization & Teacher's neural tokenizer copied; Mamba-2 layers random \\
Teacher & REALM-PT (11.5\,M, bidirectional; Table~\ref{tab:hp-pretrain}) \\
Datasets & Makin and Flint (distilled separately) \\
Seeds & $42, 123, 456$ (one student per seed and corpus) \\
Hardware / wall-clock time & 1 NVIDIA A100; REALM: 67--82\,min (Makin), 29--34\,min (Flint) \\
\bottomrule
\end{tabular}
\end{table}

\begin{table}[htbp]
\centering
\footnotesize
\setlength{\tabcolsep}{3pt}
\caption{Hyperparameters for five-fold stacking (Table~\ref{tab:stacking}). The five base models are distilled with the configuration of Table~\ref{tab:hp_distill_unsup} and differ only in which fold of the non-held-out sessions is withheld from each; the distillation seed is identical across members. Members are read out and scored on identical segments within an evaluation split, so the ensemble averages predictions for the same time points.}
\label{tab:hp_stacking}
\begin{tabular}{lc}
\toprule
\textbf{Hyperparameters} & \textbf{Values} \\
\midrule
Base models & 5 \\
Fold partition & 5 disjoint folds of the non-held-out sessions (fold seed 42) \\
Withheld per member & 1 fold; each model is distilled on the other $4/5$ \\
Distillation seed & 42 (identical for all five members) \\
Distillation configuration & As in Table~\ref{tab:hp_distill_unsup} \\
Datasets & Makin and Flint (stacked separately) \\
Hardware / wall-clock time & 1 NVIDIA A100 per member; distillation 50--74\,min (Makin), 16--34\,min (Flint) \\
\midrule
Read-out & Frozen encoder, ridge read-out refit per member \\
Evaluation splits & 3 (canonical split seeds $42, 123, 456$) \\
Combination & Unweighted mean of the members' velocity predictions \\
Reported as & Mean over all $\binom{5}{k}$ subsets of size $k$ \\
\bottomrule
\end{tabular}
\end{table}

\begin{table}[htbp]
\centering
\footnotesize
\setlength{\tabcolsep}{3pt}
\caption{Hyperparameters for the linear probe. The encoder is frozen exactly as distilled --- no per-session adaptation is performed --- and velocity is decoded by ridge regression. The read-out is solved in closed form, so it has no epochs, learning rate, initialization or stopping rule, which is what makes it a probe of the representation rather than a second model. Its one hyperparameter, the ridge penalty, is not tuned by hand either: it is chosen per session and seed on the validation fold from $\{1, 10, 10^2, 10^3, 10^4, 10^5, 10^6\}$ and then refit on the training and validation folds, by the same rule for every model in this paper, and the test fold is never used to choose it. A radial basis function kernel approximated by $4000$ random Fourier features was also measured across all $130$ frozen models in this work and changes the read-out by $-0.003$ $R^2$ on average, so the linear form is reported: the ranking does not depend on the kernel, and dropping the kernel leaves one fewer choice to defend.}
\label{tab:hp_frozen_probe}
\begin{tabular}{lc}
\toprule
\textbf{Hyperparameters} & \textbf{Values} \\
\midrule
Encoder & Frozen; no per-session masked-autoencoding adaptation \\
Encoder mode & Causal for the causal models; bidirectional for REALM-PT, PT-bi, \\
 & REALM-bi (all three sizes) and the baselines of Table~\ref{tab:bi_baseline} \\
Read-out & Ridge regression, closed form \\
Ridge penalty $\alpha$ & Variable (per session and seed; see caption) \\
Temporal context & Past-only $[t-10,\,t]$: $11$ frames, $110$\,ms \\
\midrule
Train / val / test split & Canonical $72/8/20$ (train+val fit, test scored) \\
Held-out set & The 8 sessions held out by CrossModalDistill (5 Makin, 3 Flint) \\
Metric & Per-axis $R^2$, averaged over $v_x$ and $v_y$ \\
Hardware / wall-clock time & 1 NVIDIA A100; REALM: 1--2\,min per corpus and seed \\
Seeds & $42, 123, 456$ \\
\bottomrule
\end{tabular}
\end{table}

\begin{table}[htbp]
\centering
\footnotesize
\setlength{\tabcolsep}{3pt}
\caption{Hyperparameters for per-session supervised fine-tuning: the distilled encoder is adapted on each held-out session with velocity labels, with nothing frozen except the neural tokenizer.}
\label{tab:hp_finetune}
\begin{tabular}{lc}
\toprule
\textbf{Hyperparameters} & \textbf{Values} \\
\midrule
Batch size & 32 \\
Peak learning rate & $5\times10^{-4}$ \\
Minimum learning rate & $5\times10^{-6}$ \\
Learning rate scheduler & Cosine annealing (no warmup) \\
Optimizer & AdamW \\
Weight decay & $10^{-5}$ \\
Total epochs & 100 \\
Early stopping patience & 20 \\
Gradient clipping & 1.0 \\
Train / val / test split & Canonical $72/8/20$ (train fit, validation early stopping, test scored) \\
Frozen modules & Neural tokenizer (Dual-VCT: none) \\
Hardware / wall-clock time & 1 NVIDIA A100; REALM: 4--6\,min (Makin), about 2\,min (Flint) per corpus and seed \\
Encoder mode & Causal for the causal models; bidirectional for REALM-PT \\
 & and REALM-bi (all three sizes) \\
Held-out set & The 8 sessions held out by CrossModalDistill (5 Makin, 3 Flint) \\
Seeds & $42, 123, 456$ \\
\bottomrule
\end{tabular}
\end{table}

\clearpage
\subsection{Baseline configurations}\label{appendix-baselines}

Every baseline is run under this paper's protocol rather than quoted from its paper, so that held-out sessions, split, read-out and metric are identical across models, and each is run in a causal and a non-causal form. As in REALM, no label of any session is used before the read-out, except by DPAD, whose latent states are defined by behavior. Table~\ref{tab:baseline_config} lists the configuration of each baseline.

\begin{table}[!htbp]
\centering
\caption{\textbf{Baseline configurations.} Every model is trained on the $34$ non-held-out sessions (CMD's spike teacher on $226$ sessions of spikes, none of them among the $8$ held-out sessions), evaluated on the $8$ held-out sessions under the canonical $72/8/20$ split, and evaluated either by the same linear probe (ridge regression) (Table~\ref{tab:hp_frozen_probe}) or after the fine-tuning of the last column. Parameters are those of the encoder in the causal and the non-causal form. RKD, retrospective knowledge distillation; ELBO, evidence lower bound; VMD, variational mode decomposition; BN, batch normalization.}
\label{tab:baseline_config}
\footnotesize
\renewcommand{\arraystretch}{1.0}\setlength{\tabcolsep}{4pt}
\begin{tabularx}{\linewidth}{@{}l>{\raggedright\arraybackslash\hsize=1.52\hsize}X c>{\raggedright\arraybackslash\hsize=0.64\hsize}X>{\raggedright\arraybackslash\hsize=0.84\hsize}X@{}}
\toprule
Model & Training before the read-out & Params (M) & Causal form & Fine-tuning \\
\midrule
REALM & Masked pretraining of a bidirectional teacher, then RKD into the student; no labels (Tables~\ref{tab:hp-pretrain}, \ref{tab:hp_distill_unsup}) & 4.91 / 5.54 & Mamba-2 run forward & Table~\ref{tab:hp_finetune} \\
LSTM & REALM's masked pretraining alone, without RKD & 6.14 / 5.29 & Unidirectional layers & Table~\ref{tab:hp_finetune} \\
Transformer & REALM's masked pretraining alone, without RKD & 4.95 / 5.74 & Attention masked to the past & Table~\ref{tab:hp_finetune} \\
LFADS & Its own ELBO in its published configuration, with Gaussian observations and a learning rate of $1.53\times10^{-3}$ & 0.39 / 0.39 & Initial condition a learned constant; controller encoder run forward & Whole model with a linear head; learning rate $5\times10^{-4}$, batch $32$, up to $150$ epochs, patience $20$ \\
DPAD & Its own fit, one model per corpus ($n_x = n_1 = 128$), with the velocity of these sessions as its behavior & 0.06 / 0.12 & Its own forward recurrence; non-causal form: the bidirectional layers of its implementation & Refitted per session on the training fold, early-stopped on the validation fold; artifact segments of Table~\ref{tab:artefact_sens} dropped \\
Dual-VCT & Its own VMD front end ($30$ planes: local motor potential and five band powers up to $300$\,Hz, recomputed from the Makin broadband and the $2$\,kHz Flint recordings rather than from the $100$\,Hz LFP); stage~1: masked reconstruction of the LFP with REALM's masking and schedule; no labels & 0.99 / 0.99 & Convolutions padded on the past side, attention masked to the past, pooled steps held & Table~\ref{tab:hp_finetune}; BN statistics re-estimated with dropout off before every read-out and validation score \\
CrossModalDistill & LFP student (ten Transformer layers of width $256$; published tokenizer and objective, $\lambda = 5$) distilled from our masked-autoencoding spike teacher ($226$ sessions, none of them among the held-out sessions); no labels; held-out sessions read with the mean of the $34$ space-embedding tables & 10.59 / 10.59 & Causal input convolution, attention masked to the past & The paper's procedure (new space-embedding table, all weights updated); AdamW $10^{-4}$, weight decay $0.01$, batch $8$, ${\le}\,100$ epochs, patience $20$ \\
\bottomrule
\end{tabularx}
\end{table}

\emph{Dual-VCT's first stage.} Dual-VCT~\cite{R111_li2026dualvct} was published with a supervised first stage that regresses velocity. To keep every model label-free before the read-out, the comparisons of the main text replace it by masked reconstruction of the LFP with REALM's masking and schedule (Table~\ref{tab:baseline_config}). This lowers Dual-VCT markedly: its input is a set of VMD planes of local-motor-potential and band-power features, from which the raw LFP at the masked samples is hard to reconstruct, so the self-supervised objective gives its encoder little to learn from. We therefore also ran the same network with its original supervised first stage, trained on the velocity of the $34$ non-held-out sessions, under the same split, read-out and fine-tuning. It then reaches $0.465$ under linear-probe evaluation and $0.639$ after fine-tuning, against $0.170$ and $0.367$ with the label-free first stage, and causally deployed REALM is not separated from it: $+0.031$ ($[-0.135, +0.197]$, $d_z = 0.16$, $3/8$, $p = 0.671$) under linear-probe evaluation and $-0.002$ ($[-0.064, +0.060]$, $d_z = -0.03$, $3/8$, $p = 0.933$) after fine-tuning. The supervised Dual-VCT, however, uses three things that REALM does not: the velocity labels of the $34$ non-held-out sessions, band powers up to $300$\,Hz computed from the broadband recordings, and a non-causal front end on the zero-phase chain.

\begin{table}[htbp]
\centering
\caption{\textbf{Artifact sensitivity for DPAD.} Ten held-out segments across four sessions carry LFP transients above Tukey's far-out fence ($23.1\sigma$ on the $\log_{10}$ segment peak, computed over all $1441$ held-out segments). DPAD~\cite{R34_sani2024dissociative} is scored in every regime both on the full test fold and with those segments dropped. Only the fine-tuned DPAD row enters the main text and Tables~\ref{tab:frozen_baseline}, \ref{tab:finetuned_baseline} and \ref{tab:bi_baseline} from the right-hand column, because those segments collapse that per-session fit; every other row is reported from the full test fold, as is every other model, REALM included. Training and fitting are the same in both columns and only the test fold shrinks; fine-tuned DPAD, which early-stops on the validation fold, also has the listed segment dropped from that fold in the two sessions whose validation fold holds one, in both columns alike. Dropping the segments raises fine-tuned DPAD by $+0.029$, so doing so there and nowhere else is conservative for REALM's margin. The causal rows are scored on the causally processed held-out sessions (Section~2.1 of the main text), the bidirectional row on the zero-phase chain.}
\label{tab:artefact_sens}
\footnotesize
\begin{tabular}{lccc}
\toprule
Regime & Full test fold & Artifact segments dropped & $\Delta R^2$ \\
\midrule
Linear probe, causal & 0.357 & 0.383 & $+0.026$ \\
Fine-tuned, causal & 0.464 & 0.494 & $+0.029$ \\
Linear probe, bidirectional & 0.463 & 0.507 & $+0.044$ \\
\bottomrule
\end{tabular}
\end{table}

\FloatBarrier
\section{Dataset and preprocessing details}\label{appendix-dataset}

All datasets used in this study are publicly available, as summarized in Table~\ref{tab:datasets}. The pretraining corpus comprises the $34$ non-held-out sessions of the Makin~\cite{R94_makin2018superior} and Flint~\cite{R95_flint2012accurate} datasets, $5.9$\,h of multi-channel M1 LFP from two macaques, with continuous velocity targets: cursor velocity for Makin and hand velocity for Flint. Both are brought to a common format of $96$ channels at $100$\,Hz by the chain of Section~2.1 of the main text: a notch at $60$\,Hz and its harmonics, band-pass filtering ($0.05$--$50$\,Hz), common average referencing, resampling to $100$\,Hz and per-channel $z$-scoring. For every causal result the held-out sessions are processed instead by the causal counterpart of this chain, described in Section~2.1 of the main text and compared in Section~\ref{sec:ablation_causal_prep}. The same two datasets supply the eight held-out sessions on which every result is scored, so pretraining, distillation and evaluation all draw on one coherent corpus of self-paced and center-out reaching; no held-out session is used for pretraining or distillation.

\begin{table}[htbp]
\caption{Datasets used in this work. In the Sessions column the first number is the count of sessions used for masked pretraining and retrospective distillation, and the second the count held out for evaluation; the held-out sessions enter neither stage and are used only for the per-session read-out or fine-tuning and for scoring. The total duration covers both. RT and CO denote random-target reaching and center-out reaching, respectively.}
\label{tab:datasets}
\centering
\footnotesize
\setlength{\tabcolsep}{3pt}
\begin{tabular}{lcccccc}
\toprule
{Dataset} & {Regions} & Experimental & {Subjects} & Sessions & Total & Total \\
 & & tasks & & (train--held-out) & duration (h) & channels \\
\midrule

Makin et al.~\cite{R94_makin2018superior} & M1 & RT & 1 & 25--5 & 5.89 & 96 \\
\midrule
Flint et al.~\cite{R95_flint2012accurate} & M1 & CO & 1 & 9--3 & 2.03 & 95 \\
\bottomrule
\end{tabular}
\end{table}

\begin{itemize}

\item \textbf{Makin} \citep{R93_odoherty2017nonhuman, R94_makin2018superior}: A self-paced random-target reaching dataset recorded from one monkey (Monkey~I) using a 96-channel Utah array in M1, released only as broadband at $24{,}414$\,Hz, so the LFP has to be extracted first. Decimation to the $763$\,Hz working rate is staged, $4\times4\times2$, because a single decimation by $32$ would put the anti-aliasing filter's transition band against its own Nyquist frequency; from $763$\,Hz the recording enters the shared chain. Broadband recordings are released for 30 of the dataset's 47 sessions, all of them from Monkey~I, and we use all 30, each with continuous cursor-velocity targets. We hold out 5 sessions (indy\_20160622\_01, indy\_20160630\_01, indy\_20160915\_01, indy\_20161013\_03, indy\_20170124\_01) for evaluation, matching the held-out set of CrossModalDistill~\cite{R24_erturk2025cross}; the remaining 25 sessions are used for masked pretraining and distillation.

\item \textbf{Flint} \citep{R95_flint2012accurate}: A center-out reaching dataset recorded from one monkey (Monkey~C) at $2$\,kHz across 5 experimental days (12 sessions total), with continuous hand-velocity targets. The released recording has 95 channels, and common average referencing is computed over these 95. The encoder takes a fixed $96$-channel input, so the missing channel is zero-padded and a channel mask multiplies it out before the tokenizer sees it; the tensor shape quoted in Section~2.1 of the main text is therefore the same for every corpus while the electrode count is not. We hold out 3 sessions (Flint\_e1\_1, Flint\_e4\_1, Flint\_e5\_2) for evaluation; the remaining 9 sessions are used for masked pretraining and distillation.

\end{itemize}


\subsection{The canonical evaluation split}\label{appendix-split}
Every model in this paper --- REALM, its ablations and every reproduced baseline --- is scored on the same three partitions of the held-out sessions, one per seed, generated once and written to disk for all of them to read; the single exception, described in Section~\ref{appendix-baselines}, is that fine-tuned DPAD has any of the ten artifact segments of Table~\ref{tab:artefact_sens} that fall in its test or validation fold dropped from that fold. This is what makes the paired statistics of Section~3.3 of the main text possible: two methods that each drew their own permutation of the same session would be scored on different segments, and the per-session paired differences the tests rest on would then be undefined. Within a held-out session, its $n$ non-overlapping $5$\,s segments are shuffled by a seeded random permutation, whose seed combines the run seed with a hash of the session name, and the permutation is cut at $72\%$ and $80\%$ into training, validation and test folds, so the test fold is the last $20\%$ of the permutation and is never read during fitting. Folding the session hash into the seed decorrelates the draw across sessions, and folding in the run seed means each of the three seeds $\{42, 123, 456\}$ yields a different partition, so a three-seed average spans three independent splits as well as three initializations.

\section{Additional ablations}

\subsection{The encoder backbone architecture}
\label{sec:ablation_backbone}

We use Mamba-2 as the encoder backbone. This choice is motivated by a downstream requirement of REALM: the bidirectional teacher must be distilled into a strictly \emph{causal} student. Mamba-2's selective state-space formulation provides a causal/bidirectional duality: the same architecture can be instantiated as a causal scan for the student or a bidirectional scan for the teacher, ensuring that the two share matched inductive biases and that distillation does not need to bridge a fundamental architectural gap. Beyond this structural advantage, Mamba-2 also offers linear-time complexity in sequence length, which is well suited to the long LFP windows used in behavior decoding.

We tested this choice. Table~\ref{tab:backbone_ablation} runs three backbones --- an LSTM, a Transformer with rotary position embeddings~\cite{R86_su2024roformer} and Mamba-2 --- through the identical distillation pipeline at matched capacity, and evaluates each in both regimes, on the causally processed held-out sessions.

Accuracy separates only the LSTM, and only under linear-probe evaluation. With the encoder frozen, the overall $R^2$ is $0.496$ for Mamba-2, $0.490$ for the Transformer and $0.476$ for the LSTM; after per-session fine-tuning it is $0.636$, $0.643$ and $0.625$. The Transformer is not separated from Mamba-2 in either regime; the LSTM is behind under linear-probe evaluation, by $+0.020$ ($[+0.001, +0.039]$, $d_z = 0.87$, $8/8$ sessions, $p = 0.044$), and is not separated once fine-tuned ($+0.011$, $[-0.009, +0.031]$, $d_z = 0.47$, $4/8$, $p = 0.228$). What REALM reports therefore rests on the distillation stage rather than on the backbone: the same architecture without RKD reaches $0.309$ against $0.496$ with it.

What does separate the backbones is the cost of producing one output when one new sample arrives, which is what a streaming decoder is asked to do. Mamba-2 carries an explicit state, so a step of its temporal stack costs $1.37$\,ms on the Jetson regardless of how much context it has already seen. Attention has no $O(1)$ form: the Transformer must recompute over a buffer, which costs $2.52$\,ms at $100$ samples and rises to $15.68$\,ms at $1000$. At the $500$-sample context used here that is a $5.6\times$ gap, and the Transformer crosses the $10$\,ms budget of a $100$\,Hz loop before $L=1000$ while the cost of Mamba-2 does not change. The LSTM has a stepping form too and is the cheapest of the three, but it is also the only backbone whose linear-probe accuracy is separated from, and below, Mamba-2's.

Mamba-2 is, in addition, the one backbone here that instantiates as both halves of the method, through the causal/bidirectional duality described at the start of this section. That duality and the streaming cost are why we adopt it; we do not claim that it decodes better.

\begin{table}[!htbp]
\centering
\footnotesize
\setlength{\tabcolsep}{3pt}
\caption{\textbf{Backbone at matched capacity.} Three backbones are distilled from the same teacher with the same objective ($\lambda_{\text{repr}}=1.0$, $\lambda_{\text{ae}}=1.0$, final-layer alignment, $300$ epochs, patience $30$, batch $32$) and then evaluated in both regimes on the canonical $72/8/20$ split of the eight held-out sessions: through the same closed-form ridge read-out with the encoder frozen, and after per-session supervised fine-tuning with nothing frozen except the neural tokenizer. The backbone block, with its depth set to match capacity, is the only setting that differs across the columns. Student capacity is matched to within $2\%$: $4.995$, $4.991$ and $4.909$\,M parameters for LSTM, Transformer and REALM's Mamba-2. Per-axis $R^2$, averaged over three seeds; the statistics rows compare each backbone against REALM as in Table~\ref{tab:frozen_ablation}. The lower block gives the cost of producing one output when one new sample arrives, measured on a Jetson Orin Nano in its uncapped MAXN SUPER power mode with the clocks locked (CPU $1.728$\,GHz, GPU $1.02$\,GHz), in the same session and with the same software (PyTorch~2.5) as the PyTorch rows of Table~\ref{tab:inference}, each timed call captured into a CUDA graph, as the median over $500$ invocations after $50$ warm-ups. A backbone with an $O(1)$ stepping form pays the same cost regardless of the context it has seen, so its row is flat; attention has no such form, so its row is the cost of recomputing a buffer of that length. These rows time the temporal stack alone, whereas Table~\ref{tab:inference} times the whole decoder step with its front end and read-out head, which adds about $0.06$\,ms for REALM. }
\label{tab:backbone_ablation}
\begin{tabular}{lccc}
\toprule
Backbone & LSTM & Transformer & \textbf{Mamba-2} \\
\midrule
\multicolumn{4}{l}{\emph{Linear-probe evaluation}} \\
\quad Makin mean & $0.463$ & $\mathbf{0.476}$ & $0.474$ \\
\quad Flint mean & $0.498$ & $0.513$ & $\mathbf{0.533}$ \\
\quad Overall & $0.476$ & $0.490$ & $\mathbf{0.496}$ \\
\quad \textit{$\Delta R^2$ (REALM$-$)} & $+0.020$ & $+0.006$ & -- \\
\quad \textit{95\% CI} & $[.001,.039]$ & $[-.006,.017]$ & -- \\
\quad \textit{Cohen's $d_z$} & $0.87$ & $0.42$ & -- \\
\quad \textit{sessions} & $8/8$ & $5/8$ & -- \\
\quad \textit{$p$} & $0.044$ & $0.269$ & -- \\
\midrule
\multicolumn{4}{l}{\emph{Per-session fine-tuning}} \\
\quad Makin mean & $0.636$ & $\mathbf{0.653}$ & $0.638$ \\
\quad Flint mean & $0.608$ & $0.625$ & $\mathbf{0.634}$ \\
\quad Overall & $0.625$ & $\mathbf{0.643}$ & $0.636$ \\
\quad \textit{$\Delta R^2$ (REALM$-$)} & $+0.011$ & $-0.006$ & -- \\
\quad \textit{95\% CI} & $[-.009,.031]$ & $[-.020,.008]$ & -- \\
\quad \textit{Cohen's $d_z$} & $0.47$ & $-0.36$ & -- \\
\quad \textit{sessions} & $4/8$ & $2/8$ & -- \\
\quad \textit{$p$} & $0.228$ & $0.345$ & -- \\
\midrule
\multicolumn{4}{l}{\emph{Streaming cost per output (ms), Jetson Orin Nano, CUDA Graphs}} \\
\quad \textit{$O(1)$ stepping form} & yes & no & yes \\
\quad context $L=100$ & $\mathbf{0.71}$ & $2.52$ & $1.37$ \\
\quad context $L=250$ & $\mathbf{0.71}$ & $4.09$ & $1.37$ \\
\quad context $L=500$ & $\mathbf{0.71}$ & $7.70$ & $1.37$ \\
\quad context $L=1000$ & $\mathbf{0.71}$ & $15.68$ & $1.37$ \\
\bottomrule
\end{tabular}%
\end{table}

\newpage

\subsection{Alignment depth}

The alignment term can act on any number of the last encoder layers; REALM aligns the final layer only (Table~\ref{tab:hp_distill_unsup}), and here that number is varied. We vary it under linear-probe evaluation, with everything else held at the configuration of Table~\ref{tab:hp_distill_unsup}, so that apart from the tokenizer initialization the alignment term is the only training signal from the teacher. The ``None'' variant sets $\lambda_{\text{repr}}=0$, removing the alignment loss but keeping the student's initialization, in which the neural tokenizer (per-channel temporal convolution, efficient channel attention and value embedding) is copied from the teacher and the Mamba-2 layers are random; the four variants therefore differ in alignment depth alone.

Table~\ref{tab:ablation_layer} shows that the depth barely matters; what matters is whether the term is present. Aligning the last one, two or four layers spans $0.479$ to $0.496$, and none of the three differs from the others at $n=8$: last-1 is ahead of last-2 by $+0.017$ ($[-0.001, +0.034]$, $d_z = 0.80$, $7/8$, $p = 0.059$) and of last-4 by $+0.011$ ($[-0.004, +0.026]$, $d_z = 0.61$, $6/8$, $p = 0.130$), and last-2 against last-4 is $-0.006$ ($[-0.031, +0.019]$, $d_z = -0.20$, $3/8$, $p = 0.598$). Removing the term costs $0.192$ ($[+0.087, +0.297]$, $d_z = 1.53$, $8/8$, $p = 0.003$) and lands at $0.304$, essentially the $0.309$ of the same backbone without RKD at all: with no alignment, the distillation stage buys nothing. We align the final layer only.

\begin{table}[!htbp]
\centering
\footnotesize
\setlength{\tabcolsep}{3pt}
\caption{Alignment depth under linear-probe evaluation, on the causally processed held-out sessions. Mean $\pm$ s.d.\ over three seeds ($s \in \{42, 123, 456\}$), eight held-out sessions, per-axis $R^2$. Best per column in \textbf{bold}.}
\label{tab:ablation_layer}
\begin{tabular}{lccc}
\toprule
Aligned layers & Makin & Flint & Overall \\
\midrule
None ($\lambda_{\text{repr}}=0$) & 0.349\std{0.015} & 0.229\std{0.009} & 0.304\std{0.008} \\
last-1 (adopted) & 0.474\std{0.016} & \textbf{0.533\std{0.011}} & \textbf{0.496\std{0.013}} \\
last-2 & 0.453\std{0.016} & 0.522\std{0.025} & 0.479\std{0.010} \\
last-4 & \textbf{0.474\std{0.014}} & 0.503\std{0.035} & 0.485\std{0.019} \\
\bottomrule
\end{tabular}
\end{table}

\subsection{Causal deployment of the preprocessing chain}\label{sec:ablation_causal_prep}

Three steps of the preprocessing of Section~2.1 of the main text, which is the published pipeline of CrossModalDistill, read samples later than the one being processed: the zero-phase notch and band-pass filters, the Fourier-domain resampling of the whole recording, and the per-channel standardization with the statistics of the whole session. The pretraining and distillation corpus is prepared offline, so these steps are harmless there; a held-out session, however, has to be processed as it is recorded. Every causal result in the main text and in this document is therefore reported under causal deployment: the models are trained on the zero-phase corpus as described, and each held-out session is processed by the causal counterpart of the chain given in Section~2.1 of the main text, with the same cut-offs and referencing, before any read-out, fine-tuning or scoring. This section compares causal deployment with two alternatives. In the first, the held-out sessions are processed by the zero-phase chain as well, which is the protocol of the submitted manuscript. In the second, dual-chain distillation, the teacher reads the zero-phase corpus while the student reads the same recordings band-limited by causal minimum-phase filters, so that the student is distilled on causally filtered input; it is then deployed in the same way.

Table~\ref{tab:causal_prep} shows that causal deployment lowers absolute accuracy for every model, but leaves the benefit of RKD intact. REALM falls by $0.065$ under linear-probe evaluation ($[-0.096, -0.033]$, $d_z = -1.73$, $0/8$, $p = 0.002$) and by $0.049$ after fine-tuning ($[-0.078, -0.021]$, $d_z = -1.45$, $0/8$, $p = 0.005$), and REALM-PT-c by $0.050$ and $0.058$. The benefit of RKD remains on every session in both evaluations: $+0.187$ ($[+0.119, +0.254]$, $d_z = 2.30$, $8/8$, $p < 0.001$) under linear-probe evaluation and $+0.066$ ($[+0.051, +0.080]$, $d_z = 3.71$, $8/8$, $p < 0.001$) after fine-tuning, against $+0.202$ and $+0.057$ with the zero-phase chain. Every baseline is deployed in the same way (Tables~\ref{tab:frozen_baseline} and \ref{tab:finetuned_baseline}), so the comparisons of the main text are made under one protocol. Distilling the student on causally filtered input does not help: it is $0.010$ below causal deployment under linear-probe evaluation ($[-0.018, -0.002]$, $d_z = -1.00$, $2/8$, $p = 0.026$) and level with it after fine-tuning ($-0.006$, $[-0.015, +0.004]$, $d_z = -0.50$, $2/8$, $p = 0.200$). The student is therefore distilled once, on the zero-phase corpus, and only the held-out sessions are processed causally.

\begin{table}[!htbp]
\centering
\footnotesize
\setlength{\tabcolsep}{3pt}
\caption{Causal deployment against zero-phase evaluation and dual-chain distillation. Per-axis $R^2$ on the eight held-out sessions, averaged over three seeds ($s \in \{42, 123, 456\}$), canonical $72/8/20$ split. REALM-PT-c is the same causal backbone pretrained by CMAE without RKD; it has no distillation stage, so its dual-chain entry is its causal deployment. $\Delta R^2$ is REALM minus REALM-PT-c, with the number of sessions on which REALM is ahead; $p$ is a two-sided paired $t$-test over the eight session means. Causal deployment is the protocol of every causal result in the main text.}
\label{tab:causal_prep}
\begin{tabular}{lllccc}
\toprule
Evaluation & Student distilled on & Held-out sessions & REALM & REALM-PT-c & $\Delta R^2$ ($p$) \\
\midrule
\multirow{3}{*}{Linear probe} & zero-phase & zero-phase & 0.561 & 0.359 & $+0.202$ ($8/8$, $p < 0.001$) \\
 & zero-phase & causal (main results) & 0.496 & 0.309 & $+0.187$ ($8/8$, $p < 0.001$) \\
 & minimum-phase (dual-chain) & causal & 0.486 & 0.309 & $+0.177$ ($8/8$, $p < 0.001$) \\
\midrule
\multirow{3}{*}{Fine-tuned} & zero-phase & zero-phase & 0.686 & 0.629 & $+0.057$ ($8/8$, $p < 0.001$) \\
 & zero-phase & causal (main results) & 0.636 & 0.571 & $+0.066$ ($8/8$, $p < 0.001$) \\
 & minimum-phase (dual-chain) & causal & 0.631 & 0.571 & $+0.060$ ($8/8$, $p < 0.001$) \\
\bottomrule
\end{tabular}
\end{table}

\subsection{History length at decoding}
\label{sec:ablation_history}

The $5$\,s segment length fixes how much history the causal student can draw on at decoding. To measure its influence, REALM was re-evaluated with the recurrent states and the running channel-attention statistics reset every $1$, $2$, $2.5$, $5$ or $10$\,s at test time. Training is unchanged: the linear-probe read-out and the fine-tuned model are fitted on $5$\,s segments decoded from a reset state, with the hyperparameters and seeds of the main text, and only the reset interval used to decode the test data varies. So that no test interval reads signal from a training segment, each held-out session is split $72/8/20$ at random over $10$\,s units (pairs of consecutive $5$\,s segments), and every reset interval divides the $10$\,s unit. The Monkey~C split is therefore the one of the main text, while the Monkey~I split is redrawn over $10$\,s units. The ten-frame temporal context of the linear probe is also built within each reset interval, so nothing before a reset is read. Table~\ref{tab:history} gives the result. Accuracy rises monotonically with the history available. Resetting every $1$\,s lowers $R^2$ by $0.059$ under linear-probe evaluation and $0.055$ after fine-tuning, and resetting every $2.5$\,s by $0.013$ and $0.014$, on all eight sessions. Doubling the interval from $5$ to $10$\,s raises $R^2$ by only $0.006$ under both read-outs, although on all eight sessions. Most of the benefit of history is therefore reached within the $5$\,s segments, and REALM continues to make use of history beyond the $5$\,s length it was trained on.

\begin{table}[htbp]
\centering
\caption{\textbf{History length at decoding.} Per-axis $R^2$ of REALM under causal deployment when the recurrent states and running channel-attention statistics are reset every $1$ to $10$\,s at test time, averaged over the eight held-out sessions (each session averaged over three seeds), with the Monkey~I / Monkey~C means in parentheses. The split is a random $72/8/20$ split over $10$\,s units (Section~\ref{sec:ablation_history}); the $5$\,s row is the main-text setting, and its Monkey~C values equal the main text, whose split is the same. Each shorter or longer interval is compared with the $5$\,s row: paired difference, number of sessions on which it is higher and two-sided paired-$t$ $p$-value.}
\label{tab:history}
\footnotesize
\setlength{\tabcolsep}{3pt}
\resizebox{\textwidth}{!}{%
\begin{tabular}{l cc cc}
\toprule
 & \multicolumn{2}{c}{Linear probe} & \multicolumn{2}{c}{Fine-tuned} \\
\cmidrule(lr){2-3}\cmidrule(lr){4-5}
Reset interval (s) & $R^2$ & vs.\ $5$\,s & $R^2$ & vs.\ $5$\,s \\
\midrule
1 & 0.436 (0.400 / 0.495) & $-0.059$, $0/8$, $p < 0.001$ & 0.590 (0.598 / 0.578) & $-0.055$, $0/8$, $p < 0.001$ \\
2 & 0.470 (0.440 / 0.521) & $-0.024$, $0/8$, $p = 0.001$ & 0.626 (0.634 / 0.614) & $-0.019$, $0/8$, $p = 0.002$ \\
2.5 & 0.482 (0.456 / 0.524) & $-0.013$, $0/8$, $p < 0.001$ & 0.631 (0.639 / 0.619) & $-0.014$, $0/8$, $p < 0.001$ \\
5 (main text) & 0.494 (0.471 / 0.533) & -- & 0.646 (0.653 / 0.634) & -- \\
10 & 0.500 (0.477 / 0.539) & $+0.006$, $8/8$, $p = 0.001$ & 0.652 (0.658 / 0.641) & $+0.006$, $8/8$, $p = 0.003$ \\
\bottomrule
\end{tabular}}
\end{table}

\subsection{Time-ordered split}
\label{sec:ablation_chrono}

The canonical $72/8/20$ split draws the $5$\,s segments of a held-out session randomly, so a test segment usually has training segments on either side of it. Neighbor segments share slow signal drifts and similar behavior, so a read-out fitted on them could score higher on the test fold than on a stretch of recording it has never seen. To measure this effect, REALM, REALM-RI, REALM-PT-c and the LSTM and Transformer backbones of Tables~\ref{tab:frozen_ablation} and \ref{tab:frozen_baseline} were re-evaluated with a time-ordered split of each held-out session. The first $72\%$ of the $5$\,s segments train the read-out or fine-tuning, the next $8\%$ is for validation and the last $20\%$ for testing, each block contiguous in time. Everything else is as in the main text: causal deployment, $5$\,s segments decoded from a reset state, and the same read-outs, hyperparameters and seeds. Table~\ref{tab:chrono} gives the result. Testing on the end of the session lowers every model by $0.02$ to $0.05$, as expected when the decoder has to extrapolate forward in time, but it changes no comparison: REALM remains ahead of every other model on all eight sessions under both read-outs. The benefit of RKD over REALM-PT-c is $+0.183$ under linear-probe evaluation and $+0.061$ after fine-tuning ($+0.187$ and $+0.066$ with the canonical split), and the margin over the Transformer is $+0.097$ and $+0.026$ ($+0.096$ and $+0.022$).

\begin{table}[htbp]
\centering
\caption{\textbf{Canonical against time-ordered split.} Per-axis $R^2$ under causal deployment, averaged over the eight held-out sessions (each session averaged over three seeds), with the Monkey~I / Monkey~C means in parentheses. Canonical: the random $72/8/20$ split of the main text. Time-ordered: the first $72\%$ of each session's segments for training, the next $8\%$ for validation and the last $20\%$ for testing (Section~\ref{sec:ablation_chrono}). The lower block gives REALM's paired difference from each model, the number of sessions on which REALM is higher and the two-sided paired-$t$ $p$-value, as in Table~\ref{tab:frozen_ablation}.}
\label{tab:chrono}
\footnotesize
\setlength{\tabcolsep}{3pt}
\resizebox{\textwidth}{!}{%
\begin{tabular}{l cc cc}
\toprule
 & \multicolumn{2}{c}{Linear probe} & \multicolumn{2}{c}{Fine-tuned} \\
\cmidrule(lr){2-3}\cmidrule(lr){4-5}
 & Canonical & Time-ordered & Canonical & Time-ordered \\
\midrule
REALM-RI & 0.308 (0.346 / 0.244) & 0.279 (0.333 / 0.191) & 0.492 (0.484 / 0.505) & 0.469 (0.492 / 0.430) \\
REALM-PT-c & 0.309 (0.341 / 0.256) & 0.286 (0.317 / 0.234) & 0.571 (0.579 / 0.558) & 0.542 (0.564 / 0.505) \\
LSTM & 0.385 (0.403 / 0.356) & 0.355 (0.387 / 0.303) & 0.556 (0.548 / 0.570) & 0.508 (0.504 / 0.513) \\
Transformer & 0.399 (0.407 / 0.386) & 0.372 (0.393 / 0.338) & 0.614 (0.620 / 0.605) & 0.577 (0.595 / 0.546) \\
REALM & 0.496 (0.474 / 0.533) & 0.469 (0.460 / 0.484) & 0.636 (0.638 / 0.634) & 0.603 (0.624 / 0.569) \\
\midrule
REALM $-$ REALM-RI & $+0.188$, $8/8$, $p = 0.004$ & $+0.189$, $8/8$, $p = 0.004$ & $+0.145$, $8/8$, $p < 0.001$ & $+0.134$, $8/8$, $p < 0.001$ \\
REALM $-$ REALM-PT-c & $+0.187$, $8/8$, $p < 0.001$ & $+0.183$, $8/8$, $p < 0.001$ & $+0.066$, $8/8$, $p < 0.001$ & $+0.061$, $8/8$, $p < 0.001$ \\
REALM $-$ LSTM & $+0.111$, $8/8$, $p = 0.001$ & $+0.113$, $8/8$, $p = 0.001$ & $+0.080$, $8/8$, $p < 0.001$ & $+0.096$, $8/8$, $p = 0.002$ \\
REALM $-$ Transformer & $+0.096$, $8/8$, $p < 0.001$ & $+0.097$, $8/8$, $p < 0.001$ & $+0.022$, $7/8$, $p = 0.040$ & $+0.026$, $8/8$, $p = 0.018$ \\
\bottomrule
\end{tabular}}
\end{table}

\clearpage
\bibliographystyle{iopart-num-long}
\bibliography{references}